\documentclass[11pt]{article}
\usepackage[preprint]{acl}

\usepackage{times}
\usepackage{latexsym}
\usepackage[T1]{fontenc}
\usepackage[utf8]{inputenc}
\usepackage{microtype}
\usepackage{inconsolata}
\usepackage{graphicx}
\usepackage{booktabs}
\usepackage{amsmath}
\usepackage{amssymb}
\usepackage{amsthm}
\newtheorem{proposition}{Proposition}
\usepackage{multirow}
\usepackage{enumitem}
\usepackage{fontawesome5}

\newsavebox{\fitb}
\newenvironment{fitwidth}{\begin{lrbox}{\fitb}}{\end{lrbox}%
  \ifdim\wd\fitb>\linewidth\resizebox{\linewidth}{!}{\usebox{\fitb}}\else\usebox{\fitb}\fi}

\title{The Average Patient:\\Do Simulated Psychiatric Patients Answer Like Real People?}

\author{
    \textbf{Weixiang Sun}$^{1}$\;
    \textbf{Zehong Wang}$^{1,\dagger}$\;
    \textbf{Kaiwen Shi}$^{1}$\;
    \textbf{Hong Huang}$^{2}$\;\\
    \textbf{Colby Nelson}$^{1}$\;
    \textbf{Yiyang Li}$^{1}$\;
    \textbf{Yanfang Ye}$^{1,\dagger}$ 
    \\
    \textsuperscript{1} University of Notre Dame\; 
    \textsuperscript{2} Simon Fraser University\; 
    \\
    $^\dagger$ Corresponding Authors
    \\
    \texttt{<wsun4,zwang43,yye7>@nd.edu}
    \\[2pt]
    \href{https://github.com/Weixiang-Sun/average-patient}{\faGithub~Code}\quad
    \href{https://weixiang-sun.github.io/average-patient/}{\faGlobe~Website}
}

\begin{document}
\maketitle

\begin{abstract}
Language models are used as simulated psychiatric patients to train clinicians and evaluate AI therapists, and are judged by plausibility or total scores. We give a framework for simulated respondents: an information budget bounds how much simulated patients can differ, and an average-patient hypothesis, that a model answers each item at the level typical of the named condition and severity, implies little variation, a full profile of the condition's typical symptoms and totals at or above the top of the stated severity band. We examine these implications at matched totals on item-level data from eight human samples, including a clinical and an instructed-feigning sample. Across 17 models, simulated patients vary 1.5 to 11 times less than real respondents with the same total and on most questionnaires endorse nearly every symptom at a middling level; on the PHQ-9 their symptoms covary less like real ones. People, honest or feigning, answer ``not at all'' about as often as genuine patients with the same totals, and more often than every model. Single answers are highly concentrated, and prompt changes, published personas, larger models and reasoning reduce the pattern without removing it. In interviews, patients told a severity are rated as having nearly every symptom, so an assessment agent's recognition of absent symptoms goes untested, and scored against the stated severity the agent places a quarter of them in their band, against about half of patients given a real symptom profile. Having the model word such a profile restores half to two thirds of the variation.
\end{abstract}

\section{Introduction}

Language-model patient simulators are now used to train counsellors \citep{wang2024patient, louie2026can}, to evaluate LLM therapists and psychiatric interview agents \citep{wang2024towards, lee2025psyche, yang2026mentalhospital}, and are built from clinical theory or from real patient histories \citep{sabour2026patientact, li2026synthetic}. A randomised study of more than ninety novice counsellors found that practising with simulated clients changes measured skill \citep{louie2026can}, so the fidelity of simulated patients affects the people trained with them.

Patients with the same severity report different symptoms \citep{fried2015depression}: one sleeps badly and has lost interest in things, another feels worthless and cannot concentrate. Specific symptoms differ in risk factors, impairment and course \citep{fried2015sumscores}, and they occur in groups, such as somatic and non-somatic symptoms of depression \citep{elhai2012factor}. Training with simulated clients is meant to expose trainees to this range \citep{wang2024patient}, and an agent that assesses patients has to recognise which symptoms a patient has as well as how severe they are. A simulator lacking this variation offers trainees and tested tools fewer presentations than clinicians meet, although each transcript can be plausible.

Simulators are evaluated mainly by clinicians rating transcripts for plausibility \citep{wang2024patient, hoang2026psi, yang2026mentalhospital} and by comparing simulated total scores with population data \citep{keough2026plausiblepatientsimpossiblepopulations}, which leaves two comparisons open. First, language models used as survey respondents vary less than the people they stand in for \citep{bisbee2024synthetic, petrov2024limited}, but no evaluation asks whether simulated patients with the same total score differ from one another, item by item, as much as real ones. Second, clinical psychology studies imitation by instructing healthy people to feign a disorder and comparing their answers with those of patients \citep{rogers2020clinical}; simulators have not been compared with such people, so it is unknown whether models imitate patients as people do.

We first give a framework for simulated respondents. An information budget bounds how much simulated patients can differ, and an average-patient hypothesis holds that, given only a persona, a model answers each item at the level typical of the named condition and severity. Together they predict how simulated patients differ from real respondents with the same total, and the budget limits what changes to the persona can add. We examine these implications for 17 models on public item-level data for eight questionnaires on seven kinds of symptoms from eight human samples, including a clinical sample and a study with an instructed-feigning group, and make both comparisons at matched totals, which separates how a total is composed from how high it is. We also use the personas of two published simulators and examine clinical interviews and an assessment agent.

Our contributions are as follows:

\begin{itemize}[leftmargin=*,itemsep=1pt,topsep=2pt]
\item \textbf{Simulated patients answer like an average patient.} Across 17 models, simulated patients vary 1.5 to 11 times less than real respondents with the same total and on most questionnaires endorse nearly every symptom at a middling level, as an information budget and an average-patient hypothesis predict.
\item \textbf{Common remedies reduce the pattern without removing it, and it limits what simulated patients can test.} Prompt changes, published personas, larger models and reasoning all leave simulated patients less varied than real respondents. In interviews with severity personas, an assessment agent's recognition of absent symptoms goes untested and its measured accuracy against the stated severity halves.
\item \textbf{Real symptom profiles mitigate the pattern.} Having the model word a real profile restores half to two thirds of the variation in interviews and makes absent symptoms testable.
\end{itemize}

These results bear on training clinicians with simulated patients, on benchmarks that evaluate clinical agents with them, and on research that uses language models to simulate survey respondents or study participants \citep{huang2025exposing, ye2025llms4all, sun2026prescam}: all of these need simulated individuals to differ as real ones do.

\section{Related Work}
\label{sec:rw}

\paragraph{Simulated patients and their evaluation.} Besides clinician ratings of plausibility \citep{wang2024patient, hoang2026psi, yang2026mentalhospital}, a population audit compares simulated total scores and screening rates with epidemiological data across demographic groups \citep{keough2026plausiblepatientsimpossiblepopulations}, simulated eating-disorder patients reach ceiling scores \citep{haase2026simulating}, and a workshop study finds that the PHQ-9 and GAD-7 items of four open-weight models do not function as in survey respondents \citep{shen-han-2026-reliability}. Unlike these evaluations, we test whether simulated respondents with the same total differ as much as real ones, compare simulators with people instructed to feign \citep{rogers2020clinical} and show the effect on evaluating an assessment agent.

\paragraph{Language models as survey respondents.} Language models used as synthetic survey respondents reproduce some aggregate opinions of subpopulations \citep{argyle2023out}, but their opinion distributions differ from those of the groups they represent \citep{santurkar2023whose}, vary less than real responses and change with prompt wording \citep{bisbee2024synthetic}, follow option order and labels \citep{dominguez2024questioning}, and match individuals worse than aggregates \citep{petrov2024limited}. We find the same homogeneity on clinical questionnaires, which allow three further steps: holding the total fixed separates how a response is composed from how high it is, people instructed to feign show how humans imitate a patient, and per-item answer distributions show the source of the homogeneity. Appendix~\ref{app:rw} gives a fuller discussion.

\section{A Framework for Simulated Respondents}
\label{sec:theory}

\subsection{Quantities at a fixed total}

A response to a $K$-item questionnaire is $X=(X_1,\dots,X_K)\in\{0,\dots,m\}^K$ with total $T=\sum_k X_k$; the PHQ-9 has $K=9$ and $m=3$. Real respondents follow a distribution $p$. A simulator receives a persona $\Pi\sim P_\Pi$, a text that names a severity $s$ and may add a biography or a symptom profile, and answers $\hat X\sim q(\cdot\mid\Pi)$; item order and sampling enter through $q$. For a group $r$, real or simulated, let $X$ and $X'$ be independent draws from $r$ given $T=t$. At this total we define breadth, the share of ``not at all'' answers, the share of responses endorsing every item, and diversity:
\begin{align*}
B_r(t)&=\textstyle\sum_k \Pr(X_k>0),\\
Z_r(t)&=\textstyle\frac1K\sum_k \Pr(X_k=0),\\
A_r(t)&=\Pr(\min_k X_k>0),\\
D_r(t)&=\mathbb{E}\lVert X-X'\rVert_1 .
\end{align*}
The first three describe composition, which symptoms carry the total, and $D_r$ the amount of variation. Structure is the correlation matrix $\rho_r$ of the residuals $X-\mathbb{E}_r[X\mid T]$. Against the real reference $\rho_p$ we measure its similarity and the contrast between the somatic and the non-somatic items \citep{elhai2012factor},
\begin{align*}
S_r &= \operatorname{corr}_{j<k}\bigl(\rho_{r,jk},\rho_{p,jk}\bigr),\\
C_r &= \bar\rho_r^{\,\text{within}}-\bar\rho_r^{\,\text{between}},
\end{align*}
where the means run over pairs of items within the same group and across the two groups. Separability is the AUC of a logistic regression $\sigma(\beta_0+\sum_k\beta_k x_k)$ that tells simulated from total-matched real responses.

\begin{proposition}\label{prop:div}
With $r_{t,k}$ the distribution of $X_k$ given $T=t$,
\[
D_r(t)=\textstyle\sum_k\sum_{a,b}|a-b|\,r_{t,k}(a)\,r_{t,k}(b).
\]
\end{proposition}

\begin{proposition}\label{prop:sep}
If $q_{t,k}=p_{t,k}$ for every $t$ and $k$, the best-fitting logistic regression on item-wise features of total-matched samples is constant, with AUC $1/2$.
\end{proposition}

\noindent Diversity thus depends only on the item distributions at each total, an additive classifier cannot detect a difference confined to how items covary, and $S_r$ and $C_r$ measure that covariation.

\subsection{Information budget}

\begin{proposition}[Information budget]\label{prop:budget}
For any persona distribution and simulator,
\begin{align*}
H(\hat X) &\le I(\Pi;\hat X)+\textstyle\sum_k H(\hat X_k\mid\Pi)\\
 &\le H(\Pi)+\textstyle\sum_k H(\hat X_k\mid\Pi).
\end{align*}
Answers to one persona $\pi$ have perplexity at most $2^{h(\pi)}$, and two of them are identical with probability at least $2^{-h(\pi)}$, where $h(\pi)=\sum_k H(\hat X_k\mid\Pi=\pi)$.
\end{proposition}

\noindent Simulated patients can therefore differ only through information that the persona carries into the answers, $I(\Pi;\hat X)$, or through the randomness of single answers. A persona that names one of four severities has $H(\Pi)\le 2$ bits, and 400 client descriptions have up to $\log_2 400\approx 8.6$ bits, enough for each client to answer differently.

\subsection{The average-patient hypothesis}

We hypothesise that a model answers each item at the level typical of that item for the named condition and severity: for a persona $\pi$ that names severity $s$, there are levels $\ell_k(s)\in\{0,\dots,m\}$ and an $\varepsilon\le\frac12$ with
\[
q(\hat X_k=\ell_k(s)\mid\pi)\ge 1-\varepsilon\quad\text{for every item }k.
\]
When every item is a core symptom of the named condition, as on the PHQ-9 for depression, one level $\ell(s)\ge1$ serves all items. The persona is the model's only information, and instruction-tuned models give concentrated answers to survey questions \citep{santurkar2023whose}.

\begin{proposition}[Average patient]\label{prop:avg}
Under the hypothesis, answers to $\pi$ satisfy
\begin{align*}
H(\hat X_k\mid\pi)&\le h_2(\varepsilon)+\varepsilon\log_2 m,\\
\mathbb{E}\lVert\hat X-\hat X'\rVert_1&\le 2mK\varepsilon,\\
\Pr\bigl(\hat T=\textstyle\sum_k\ell_k(s)\bigr)&\ge 1-K\varepsilon,
\end{align*}
and, if every $\ell_k(s)\ge1$, $A\ge 1-K\varepsilon$ and $Z\le\varepsilon$, where $h_2$ is the binary entropy and $\hat X,\hat X'$ are independent answers to $\pi$.
\end{proposition}

\noindent Simulated patients should therefore vary little, endorse every item whose typical level is above zero, and give totals near $\sum_k\ell_k(s)$: on the PHQ-9, 9 for mild, 18 for moderate and 27 for severe depression, at the top of the bands 5--9 and 20--27 and above the band 10--14. The hypothesis and Proposition~\ref{prop:avg} concern the answers to one persona; we check the hypothesis on items asked one at a time, and the comparisons at matched totals, where the measures condition on the total, test its predictions empirically. The hypothesis leaves $\rho_q$ unconstrained, and independent noise raises $D_q$ without restoring $\rho_p$. By Proposition~\ref{prop:budget}, the budget grows only with $I(\Pi;\hat X)$ or the entropy of single answers, and within it item dependence also changes the joint entropy. Appendix~\ref{app:proofs} gives the proofs.

\section{Data and Method}
\label{sec:method}

\subsection{Instruments and human references}
\label{sec:data}

\textbf{Main instruments.} The PHQ-9 \citep{kroenke2001phq} rates nine depressive symptoms over two weeks from 0, not at all, to 3, nearly every day; item 9 asks about thoughts of being better off dead or of self-harm. NHANES\footnote{\url{https://wwwn.cdc.gov/nchs/nhanes/}} records it for $N{=}41{,}714$ adults in eight cycles from 2005 to 2023. The GAD-7 \citep{spitzer2006brief} rates seven anxiety symptoms on the same scale; the 2019 NHIS\footnote{\url{https://www.cdc.gov/nchs/nhis/}} Sample Adult file records it for $N{=}31{,}132$ adults, with the PHQ-8 \citep{kroenke2009phq} to check NHANES.

\textbf{Feigning study.} A public OSF study \citep{osf6b8jf} gives the PCL-5 \citep{blevins2015posttraumatic}, 20 post-traumatic symptoms rated 0--4, for university students in three groups: 69 genuinely injured, the genuine patients below, 46 uninjured answering honestly, and 39 uninjured told that an injury would earn a tuition deduction they are not entitled to and that ``no one would ever find out''.

\textbf{Extension.} Five more datasets from the Item Response Warehouse \citep{domingue2025introduction} add seven references: the K6 distress scale \citep{kessler2002short} in NHIS 2016--2018; the OCI-R \citep{foa2002obsessive}, the PHQ-15 somatic scale \citep{kroenke2002phq} and the ITQ post-traumatic items \citep{cloitre2018international} from a UK quota sample \citep[C19PRC;][]{mcbride2021monitoring}; the EDE-Q \citep{fairburn1994assessment} in the Mexican population \citep{contreras2022eating}; the PHQ-9 at intake to a depression trial in rural Chinese adults aged 60 or more \citep[COACH, clinical;][]{chen2018protocol}; and the PCL-5 in Greek women after caesarean section \citep[obstetric;][]{orovou2021psychometric}. Table~\ref{tab:generality} gives sample sizes, Appendix~\ref{app:samples} sources and preprocessing.

\subsection{Models and administration}
\label{sec:admin}

We query 17 models: GPT-4o-mini, GPT-4o, GPT-4.1-nano, GPT-4.1-mini, GPT-4.1, GPT-5-mini, GPT-5.2, GPT-5.4-mini, Claude Haiku 4.5, Grok-4.1-fast, Llama-3.3-70B, Kimi-K2.5, Mistral-Large-3, Phi-4, and Gemini 2.5 Flash-Lite, Flash and Pro, with reasoning off or at its minimum; Appendix~\ref{app:prompts} gives the settings and Appendix~\ref{app:measures} the runs with reasoning on. Each model completes each questionnaire in nine cells, following the simulation design used with human feigners \citep{rogers2020clinical}: an honest control plus $\{$naive, coached$\}\times\{$mild, moderate, moderately severe, severe$\}$, with 50 trials per cell and the item order randomised per trial. Each trial gives the model a one-line persona naming the severity, as in population audits \citep{shen-han-2026-reliability}, and the coached arm adds that questionnaires contain checks for exaggeration; Appendix~\ref{app:prompts} gives the prompts. We also use the personas of two established simulators without their dialogue frameworks: the 12 clinician-validated depression personas of TalkDep \citep{wang2025talkdep} and the 400 client intake forms of CACTUS \citep{lee2024cactus}, alone and with each client's case formulation. In the feigning study, models take the three roles with matched injuries.

\subsection{Estimation}
\label{sec:measures}

\begin{figure*}[t]
\centering
\includegraphics[width=\textwidth]{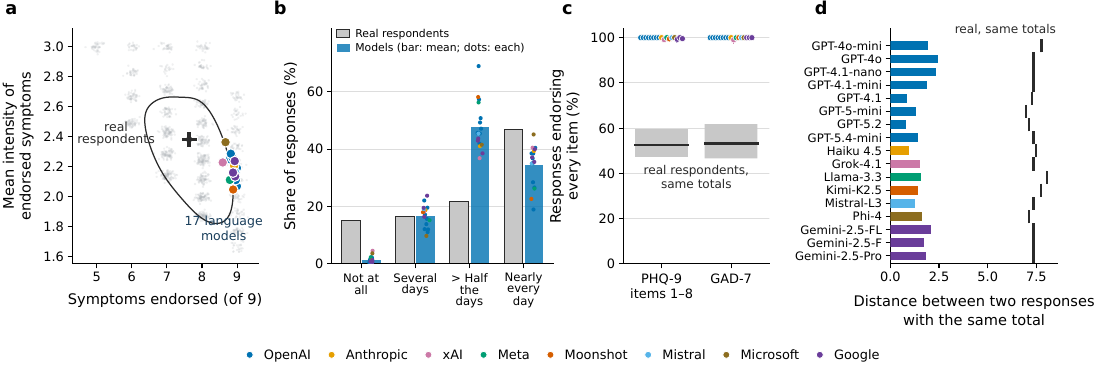}
\caption{\textbf{Response profiles on the PHQ-9 and GAD-7.} Panels a, b and d use PHQ-9 responses with total 15--24. \textbf{a}, Symptoms endorsed against their mean intensity (grey dots, real respondents; contour, their densest half; coloured, model means). \textbf{b}, Use of the response options. \textbf{c}, Responses endorsing every item, PHQ-9 items 1--8 (total 15--24) and GAD-7 (12--19), with 95\% intervals; grey, real respondents at each model's totals (range and median over models). \textbf{d}, Diversity; ticks, real respondents at each model's totals.}
\label{fig:phenomenon}
\end{figure*}

Each model is compared with real respondents at exactly its totals: on the PHQ-9 and GAD-7 at totals 15--24 and 12--19, where both groups have ample responses, and on the nine references at every total that real respondents reach, in narrow bins on long scales. Residual correlations use totals 5--24, with all NHANES respondents as $\rho_p$, and separability is the 5-fold cross-validated AUC against total-matched real responses. Item entropies come from the 50 whole forms of each severity cell and, for ten models, from log-probabilities when an item is asked alone \citep{shiuncertainty}. A one-factor graded response model (GRM) fit to NHANES 2005--2012, whose items are independent given one trait, gives profiles without the population's residual covariation. Appendix~\ref{app:measures} gives intervals and permutation tests \citep{ojala2010permutation}.

\section{Response Profiles at Matched Total Score}
\label{sec:flat}

\subsection{PHQ-9 and GAD-7}

\textbf{Composition.} As the hypothesis predicts, Figure~\ref{fig:phenomenon}a shows that at PHQ-9 totals 15--24 every model endorses more symptoms at lower intensity than real respondents with the same totals. In Figure~\ref{fig:phenomenon}b, models answer ``not at all'' in 0--4.5\% of answers, against 10--15\% for real respondents. Figure~\ref{fig:phenomenon}c shows that models endorse all of items 1--8 in 99.1--100\% of responses and all seven GAD-7 items in 99--100\%, against 47--62\% of real respondents with the same totals. Figure~\ref{fig:phenomenon-full}d,e and Table~\ref{tab:severity} show the same in every band with symptoms.

\textbf{Amount of variation.} In Figure~\ref{fig:phenomenon}d, two responses from one model differ 3.0--8.6 times less than two real respondents with the same total. Item scores alone separate simulated from total-matched real responses with an AUC of 0.68--0.86 on the PHQ-9 and 0.63--0.77 on the GAD-7, against 0.56 for two real samples.

\begin{table}[t]\centering\footnotesize\setlength{\tabcolsep}{2.2pt}
\begin{fitwidth}
\begin{tabular}{@{}llrrrrrr@{}}\toprule
& & & & & & \multicolumn{2}{c}{All items (\%)} \\\cmidrule(l){7-8}
Instr. & Sample & $N$ & Cov. & Div.\ ratio & AUC & Real & Models \\\midrule
PHQ-9 & NHANES & 41{,}714 & 100 & 3.5--11.2 & 0.66--0.83 & 25 & 40--86 \\
PHQ-9 & COACH$^{\dagger}$ & 2{,}359 & 84 & 3.2--8.2 & 0.72--0.91 & 22 & 31--89 \\
GAD-7 & NHIS & 31{,}132 & 100 & 3.9--10.7 & 0.65--0.76 & 45 & 71--100 \\
K6 & NHIS & 82{,}339 & 100 & 2.5--10.0 & 0.60--0.80 & 63 & 83--100 \\
PHQ-15 & C19PRC & 2{,}025 & 74 & 2.1--5.4 & 0.80--0.97 & 41 & 2--58 \\
OCI-R & C19PRC & 1{,}406 & 72 & 2.0--4.2 & 0.77--0.99 & 51 & 5--98 \\
ITQ & C19PRC & 2{,}025 & 100 & 1.5--4.3 & 0.63--0.79 & 74 & 81--100 \\
PCL-5 & Postpartum$^{\dagger}$ & 469 & 44 & --- & 0.85--0.97 & 14 & 15--77 \\
EDE-Q & Mexico & 2{,}090 & 78 & 2.7--5.5 & 0.88--0.96 & 18 & 71--100 \\
\bottomrule\end{tabular}
\end{fitwidth}
\caption{Profile comparison on nine human references over all totals, each simulated response matched to real respondents with the same total (ranges over models). Cov.: \% of simulated responses whose total real respondents reach. Div.\ ratio: real over model diversity, where five or more totals carry it (none on the postpartum PCL-5; Table~\ref{tab:gen-ratio}). AUC: separability from matched real respondents (chance 0.46--0.52). All items: \% endorsing every item. Real and Cov.: medians over models. $^{\dagger}$Clinical (COACH) and obstetric (postpartum) samples.}
\label{tab:generality}\end{table}

\textbf{Structure.} Residual correlations at a fixed total depend on the item distributions as well as on covariation, so we also draw responses with each model's item distributions at each total and no further covariation. Real samples drawn at each model's totals agree with NHANES at $S=0.53$--$0.77$, the models at 0.13--0.61 and the responses without covariation at $-0.02$ to 0.35. The models thus covary beyond their item distributions, 11 of 17 above the 97.5th percentile of that reference, but less like the population than real respondents do, 16 of 17 below the real samples; Tables~\ref{tab:structure-summary} and~\ref{tab:structure} give the values. Adding independent noise until the models are as varied as real respondents lowers $S$ to 0.01--0.12. The GRM gives $S=0.40$ with 20{,}000 profiles and no somatic--non-somatic contrast, although Table~\ref{tab:grm-heldout} shows that it is as varied as held-out 2013--2023 respondents and as hard to separate from them, as Propositions~\ref{prop:div} and~\ref{prop:sep} imply.

\textbf{Stated severity.} Told they are mildly, moderately, moderately severely or severely depressed, the models give median totals of 12, 17, 21 and 25, near the $9\ell(s)$ of Proposition~\ref{prop:avg} for moderate and severe depression and above it for mild. Table~\ref{tab:overshoot} shows that each model's total lies above the band of a mild to moderately severe label in 61--100\% of naive-arm responses. A label may also be read as a diagnosis, for which a total of 10 or more is the usual threshold \citep{kroenke2001phq}, but told only of mild daily-life impact, both models tested exceed the mild band in 94--98\% of responses.

\subsection{Nine human references}
\label{sec:general}
Table~\ref{tab:generality} and Figure~\ref{fig:ed-gen} apply one total-matched protocol to all nine references. Real respondents with the same total are 1.5--11 times more varied than every model wherever at least five totals carry the estimate, and every model is separable from them with an AUC of at least 0.60, against 0.49--0.56 between real samples. Models endorse every item more often on seven references, including the clinical sample, whose patients endorse all nine items less often than NHANES respondents with the same totals, 2.5\% against 5.0\%. On the PHQ-15 and the OCI-R, Figure~\ref{fig:ed-extra} shows that models endorse fainting spells, pain during intercourse and hoarding less often than real respondents, 21\% against 76\% for fainting, and most other items more often, consistent with levels $\ell_k(s)$ of zero for atypical symptoms.

\subsection{Comparison with instructed feigners}
\label{sec:feign}

\begin{figure*}[t]
\centering
\includegraphics[width=\textwidth]{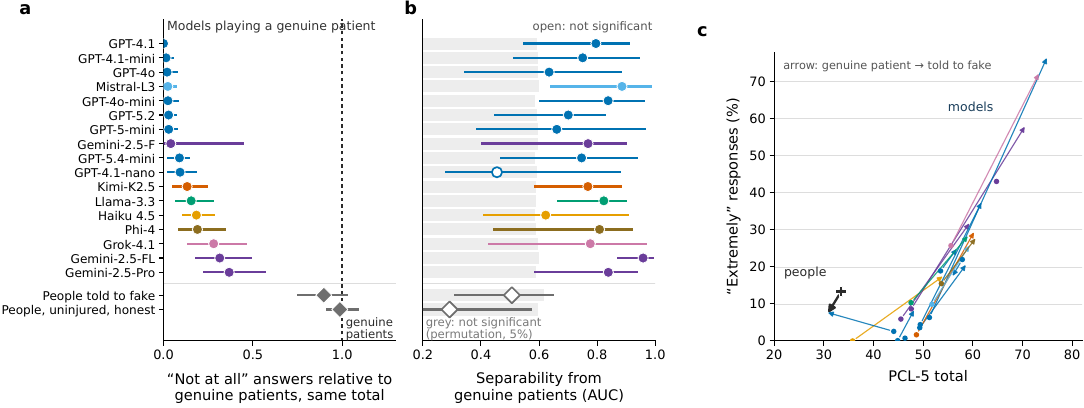}
\caption{\textbf{Comparison with people instructed to feign} (PCL-5). \textbf{a}, ``Not at all'' answers relative to genuinely injured students with the same totals (95\% bootstrap intervals). \textbf{b}, Separability from them, with the 2.5th--97.5th percentiles over 80\% subsamples; grey, below each group's one-sided 5\% critical value of a label-permutation test; open, not significant (Holm-adjusted for models). \textbf{c}, Change in total and in ``extremely'' answers from patient to feigning role.}
\label{fig:feigners}
\end{figure*}

\begin{figure*}[t]
\centering
\includegraphics[width=\textwidth]{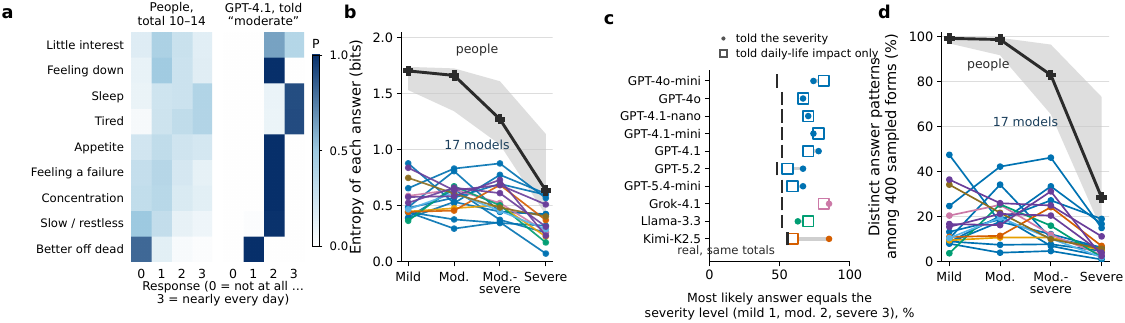}
\caption{\textbf{Per-item answer distributions.} \textbf{a}, Answer probabilities per PHQ-9 item: real respondents with total 10--14 and GPT-4.1 told ``moderate'' (log-probabilities). \textbf{b}, Entropy of each item's answers across the whole-form responses of a severity cell. \textbf{c}, Items whose most likely answer, from log-probabilities, equals the severity level, told the severity (dots) or only how daily life is affected (squares); ticks, real respondents with the same totals as the model's whole forms; moderately severe is left out. \textbf{d}, Distinct profiles among 400 forms sampled item by item from the answers in b. People in b, d: real respondents at each model's totals (grey, range over models; line, median).}
\label{fig:mechanism}
\end{figure*}

As Figure~\ref{fig:feigners}a and Table~\ref{tab:pcl5} show, people told to feign an injury for money answer ``not at all'' 0.90 times as often as genuine patients with the same totals (95\% interval 0.75--1.03), and uninjured students answering honestly 0.99 times as often (0.91--1.09). The 17 models playing a genuinely injured student answer ``not at all'' at most 0.37 times as often. On a log-odds scale the feigners' odds ratio is 0.71, so they too answer ``not at all'' somewhat less often than genuine patients, and every model's is 0.26 or less, below the feigners' after Holm adjustment. Honest and feigning people therefore compose a total closer to that of genuine patients than any model. Figure~\ref{fig:feigners}c shows that, told to feign, people do not raise their total, whereas 15 of 17 models raise both their total and their share of ``extremely'' answers. In Figure~\ref{fig:feigners}b, a classifier separates 16 of 17 models from genuine patients and does not detect the feigners, whose AUC is 0.51, although with 14--24 genuine patients per model this test is weak.

\section{Per-Item Answers, Prompts and Scale}
\label{sec:mech}

\subsection{Per-item answer distributions}

Figure~\ref{fig:mechanism}a,b and Table~\ref{tab:mechanism} show that, as the hypothesis requires, single answers are highly concentrated: their entropy averages 0.29--0.73 bits for the models, out of at most 2, against 1.15--1.51 bits for real respondents with the same totals. At the average item entropy, the perplexity bound of Proposition~\ref{prop:budget} is 6--97 for the models and about 1{,}300--12{,}500 for real respondents with the same totals. In Figure~\ref{fig:mechanism}d, forms drawn item by item from these distributions are at most 32\% distinct on average, against 69--92\% for the same draws from real respondents. In Figure~\ref{fig:mechanism}c, the most likely answer is $\ell(s)$ for 63--85\% of items, also without the severity word, against 48--56\% for real respondents.

\subsection{Prompt changes, personas and scale}
\label{sec:fix}

\begin{figure*}[t]
\centering
\includegraphics[width=\textwidth]{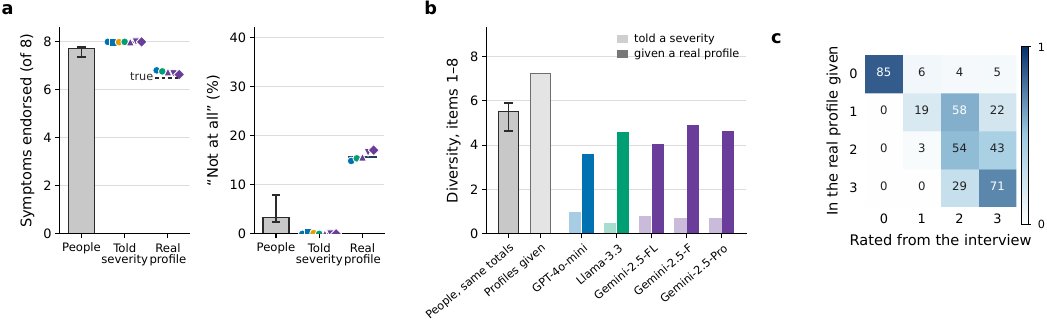}
\caption{\textbf{Simulated clinical interviews.} \textbf{a}, Symptoms endorsed and ``not at all'' answers, PHQ-9 items 1--8 rated from transcripts (dots, patient models coloured as in b; line, median; people, real respondents with the same totals; dashed, the profiles given). \textbf{b}, Diversity. \textbf{c}, Answer in the real profile given against the answer rated from the interview, row percentages.}
\label{fig:conversation}
\end{figure*}

\begin{table}[t]\centering\footnotesize\setlength{\tabcolsep}{2.5pt}
\begin{fitwidth}
\begin{tabular}{@{}lrrrr@{}}\toprule
Condition & Diversity (\%) & N.a.a. & AUC & Calib. \\\midrule
Real respondents, held out & 100 & 9.9 & --- & --- \\
\midrule
One-line persona & 16 (9--29) & 0.0 & 0.81 (0.64--0.88) & 39 \\
+ ``people differ'' & 29 (26--44) & 0.1 & 0.84 (0.76--0.87) & 48 \\
+ biography & 23 (17--29) & 0.0 & 0.80 (0.71--0.86) & 47 \\
Target total, no label & 35 (27--51) & 0.2 & 0.69 (0.59--0.82) & 68 \\
+ 5 real answer sheets & 48 (31--57) & 0.5 & 0.68 (0.57--0.69) & 83 \\
\midrule
TalkDep personas & 41 (32--50) & 0.0 & 0.86 (0.75--0.89) & --- \\
CACTUS personas & 39 (33--42) & 1.1 & 0.88 (0.84--0.92) & --- \\
+ case formulation & 37 (32--40) & 1.2 & 0.91 (0.86--0.93) & --- \\
\bottomrule\end{tabular}
\end{fitwidth}
\caption{Prompt changes and published personas, PHQ-9, seven models: median (range). Diversity: \% of held-out real respondents with the same totals (totals 5--27). N.a.a.: ``not at all'' answers, items 1--8 at totals 15--24 (\%). AUC: separability from held-out real respondents (totals 15--24). Calib.: \% of totals in the instructed band; the personas name no band. TalkDep: pairs of different personas, a persona in one fold.}
\label{tab:remedies}\end{table}

Table~\ref{tab:remedies} shows that for the seven models in Table~\ref{tab:coverage}, stating that people with the same severity differ, adding a biography or replacing the severity word with a target total changes separability from held-out real respondents by $-0.08$ to $+0.03$ on average. The information budget allows a biography or a target total to help, since they add information to the persona, so their small effect shows that little of this information reaches the composition of the answers. Five real answer sheets from the target band, which add information about symptoms, lower separability most, by 0.14, and raise diversity to 31--57\% of the real value at the same totals. The 400 CACTUS clients, each answering once, carry up to 8.6 bits, but the 17 models give 43--68\% distinct answer patterns to them on the PHQ-9, against 97--98\% for real respondents with the same totals. For TalkDep, whose 12 personas are answered repeatedly, we compare only answers of different personas, and Tables~\ref{tab:personas} and~\ref{tab:personas-all} show that with the published personas all 17 models stay 1.6--4.8 times less varied than real respondents on both questionnaires. Figure~\ref{fig:interventions}d,e and Table~\ref{tab:scale} show that larger tiers in a family do not reliably raise diversity and that none of 52 configurations exceeds 57\% of real diversity.

\section{Interviews and Downstream Use}
\label{sec:conv}

\subsection{Simulated clinical interviews}

We interview simulated patients with a fixed script covering the nine PHQ-9 domains, so that every item can be rated. GPT-4.1 and Claude Haiku 4.5 each code the PHQ-9 from the transcript, neither for its own patients, and seven patient models give 80 interviews with a severity persona and 84 with the 12 TalkDep personas. Interviews with a severity persona are uniform: Figure~\ref{fig:conversation}a and Table~\ref{tab:interview} show that 97.5--100\% are rated as endorsing all eight non-suicidality symptoms, against 49--85\% of real respondents with the same totals, and Figure~\ref{fig:conversation}b that they differ little from one another. Real profiles passed through the same interview are also compressed, but Table~\ref{tab:interview-pipeline} shows that severity-persona interviews remain 2.9--4.6 times less varied than these, and Table~\ref{tab:interview-talkdep} that TalkDep interviews of different personas remain 1.6--2.2 times less varied; with personas labelled moderate or above, 81--100\% are rated as having all eight symptoms, against 45--80\% of real respondents at those totals.

\subsection{Profiles from real data or an item response model}

\begin{figure*}[t]
\centering
\includegraphics[width=\textwidth]{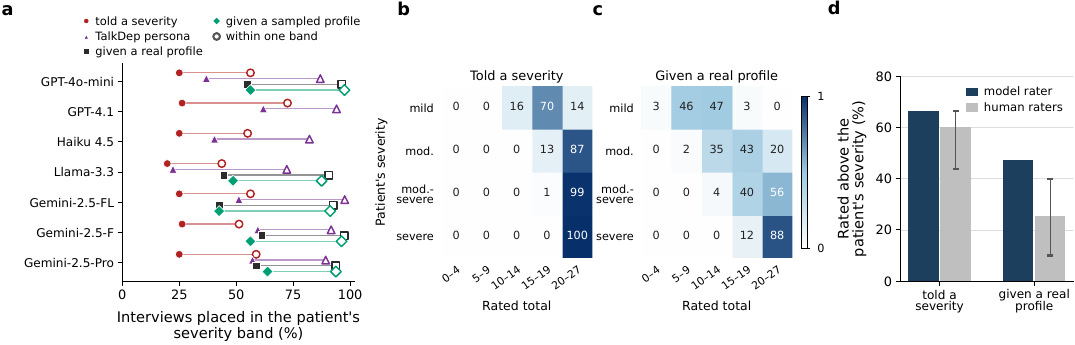}
\caption{\textbf{Severity rated by an assessment agent.} \textbf{a}, Interviews whose rated PHQ-9 total falls in the patient's severity band (filled) or within one band (open), per patient model and condition; for TalkDep personas, the band named like the persona's severity label. \textbf{b}, \textbf{c}, Rated total against the patient's severity, row percentages. \textbf{d}, Interviews rated above the patient's severity by the model rater and by five human raters on the 64 transcripts they rated, severities weighted equally (range over raters).}
\label{fig:agent}
\end{figure*}

Given a real NHANES profile as private background, five patient models give interviews whose rated answers follow the given ones with an item correlation of 0.70--0.83, as Figure~\ref{fig:conversation}c shows. In Figure~\ref{fig:conversation}b, these interviews differ from one another more than those with a severity persona, with a diversity of 3.30--4.91 against 0.36--1.01, which recovers half to two thirds of the diversity of the given profiles. Table~\ref{tab:interview} shows the same for profiles sampled from the GRM, which can be used where real profiles cannot. Five graduate students without clinical training, shown no condition labels or profiles, rated 64 transcripts. Table~\ref{tab:human-rating} shows that they agree with one another, with a weighted $\kappa$ of 0.79--0.87, and rate all eight symptoms as present in each of the 14 severity-persona interviews. They rate symptoms given as ``several days'' higher in 75\% of cases, against 91\% for the model rater, so part of this upward shift lies in the patients' wording.

\subsection{Evaluating an assessment agent}
\label{sec:agent}

We treat the interviewer and the model rater as an assessment agent and score it against what each simulated patient was given, as benchmarks do \citep{lee2025psyche, wang2025talkdep, li2025food4all, li2026hoosierhelp}. Uniform patients leave symptom recognition untested. Patients given a real profile lack 18.8\% of symptoms 1--8, and the agent matches the profile on 96--98\% of symptoms, against 81\% for an agent that reports every symptom present. Figure~\ref{fig:conversation}a shows that patients told a severity are rated as lacking almost none of their symptoms, 0--0.3\%, so on them the two agents cannot be told apart.

Figure~\ref{fig:agent}a shows that the agent places 42--61\% of patients given a real profile in their severity band and 20--26\% of patients told a severity. In Figure~\ref{fig:agent}b,c, patients told they are mildly to moderately severely depressed are rated above that band in 99--100\% of interviews, so only those told they are severe are placed correctly, and scoring the agent against a stated severity halves its measured accuracy, whether the patient overstates the label or reads it as a diagnosis. TalkDep personas, which state the severity of each key symptom, are placed in their labelled band in 22--62\% of interviews, about as often as patients given a real profile, and the agent ranks them by BDI-II score with a Spearman correlation of 0.87--0.95. On 64 transcripts with severities weighted equally, Table~\ref{tab:agent} shows that human raters place 35\% of patients told a severity and 56\% of those given a real profile in their band, against 33\% and 48\% for the model rater, and Figure~\ref{fig:agent}d that they rate 60\% and 26\% above it. In Appendix~\ref{app:screening}, a screening model trained on simulated patients is miscalibrated and less accurate than one trained on total-matched real respondents.

\section{Conclusion}

Across 17 models and nine references, simulated patients behave as our framework predicts: they spread a total over more symptoms than real respondents with the same total, vary less and exceed the PHQ-9 band of their stated severity, and prompt changes, published personas, larger models and reasoning do not remove this. We recommend evaluating simulators item by item at matched totals, including covariation, scoring assessment agents against a given profile, and grounding simulated patients in real symptom profiles.

\section*{Limitations}

The human references are public survey, clinical and student samples that differ from the models' personas in population, language and mode of administration, and the comparison with feigners rests on one public study of post-injury symptoms. The interviews follow a fixed script and are rated by models, with human raters checking a subset, and the assessment agent is our own interviewer and rater; published simulators run with their dialogue frameworks, and other agents \citep{wang2026reasoning, sun2026collaboration, ma2026proplay}, may behave differently. The severity personas use the names of the PHQ-9 bands, which a model may also read as diagnoses, and the analysis of covariation uses the PHQ-9. We study models as simulators use them, through prompts and personas; how training, such as instruction tuning, gives rise to the concentration of single answers is left for future work.

\section*{Ethics Statement}

We use de-identified public data released for secondary analysis, and no participants were recruited for the questionnaire comparisons. Five graduate students volunteered, without payment, to rate model-generated interview transcripts; they were told beforehand that some transcripts contain role-played thoughts of death or self-harm and that they could stop at any time. Prompts ask models to role-play psychiatric symptoms, including suicidal ideation, inside an explicit clinical-training frame, and no output reached a user. The results concern systems used to train clinicians, and the code is released so that simulators can be evaluated. Sampling profiles from real data could also make feigned questionnaires harder to detect; we report no method aimed at evading validity checks.

\bibliography{custom}

\appendix

\section{Extended Related Work}
\label{app:rw}

\paragraph{LLM patient simulators for mental health.} \citet{wang2024patient} build PATIENT-$\psi$, simulated clients constructed from CBT cognitive models and validated by clinicians, and show that practising against them changes what trainees notice. Later systems vary the source of the persona: \citet{sabour2026patientact} derive profiles from clinical theory, \citet{lee2025psyche} generate them for structured psychiatric assessment, and \citet{li2026synthetic} ground them in longitudinal social-media histories aligned to diagnostic interviews. Simulated clients are also used to evaluate LLM therapists \citep{wang2024towards}. \citet{louie2026can} run a randomised study with more than ninety novice counsellors and find simulated practice improves measured skill, so the fidelity of the simulation matters for training.

Evaluation of these systems relies mainly on expert judgement. \citet{hoang2026psi} build a clinically grounded benchmark for depression patient simulators with high expert agreement, and \citet{yang2026mentalhospital} construct a virtual psychiatric environment with clinician-validated evaluators. These benchmarks measure whether a transcript looks plausible to a rater and do not compare simulated responses with item-level data from real respondents.

\paragraph{Language models as survey respondents.} Models have been given standard personality inventories \citep{serapio2023personality} and used as synthetic survey respondents whose aggregate opinions track real subpopulations \citep{argyle2023out}. Individual responses match less well: opinion distributions differ from those of the demographic groups a model is asked to represent \citep{santurkar2023whose}, synthetic responses vary less than real ones and change with prompt wording \citep{bisbee2024synthetic}, answers follow option order and labels and approach uniform randomness once these are adjusted for \citep{dominguez2024questioning}, and reproduction is better at the aggregate than the individual level, with respondents too homogeneous and too internally consistent \citep{petrov2024limited}. Our diversity result extends these findings to clinical questionnaires, and the person-fit result (Table~\ref{tab:personfit}) shows the same excess consistency. Clinical questionnaires are scored, so holding the total fixed separates how a response is composed from how high it is, which population comparisons cannot do. People instructed to feign show how humans imitate a patient, people answering honestly show that the comparison separates models from people and does not grade imitations, and per-item log-probabilities show that each answer is concentrated at the severity the prompt implies.

Closest to this work, \citet{keough2026plausiblepatientsimpossiblepopulations} audit the epidemiological fidelity of LLM mental-health simulation across 120 demographic cohorts. Against NHANES they compare PHQ-8 means and screening rates, which the simulations overestimate for most groups, and the marginal shares of answer categories; within the simulated data they find that elevated cases mostly satisfy the DSM-5 gateway rule, that repeated draws of the same prompt often land in different severity categories, and that item correlation matrices differ between cohorts. Their comparisons with real data concern totals, screening rates and marginal answer shares without holding severity fixed, and an earlier version's variance-compression result was withdrawn. \citet{shen-han-2026-reliability} give four open-weight models severity personas on the PHQ-9 and GAD-7 and compare them with NHANES and NHIS respondents by confirmatory factor analysis and differential item functioning: the models are highly internally consistent, fail scalar invariance and show differential functioning on most PHQ-9 items. Differential item functioning compares item means at the same total and leaves open how much respondents with the same total differ from one another. We compare each simulated response item by item with real respondents who have the same total, measure how much responses with the same total differ, add a human feigning reference, and trace the result to per-item answer distributions; the deviation we find is not visible in total-score distributions. \citet{haase2026simulating} simulate eating-disorder patients on the 28-item EDE-Q, find scores that saturate at ceiling and fail to preserve severity rankings, and propose testing whether this extends to behaviourally graded instruments such as the PHQ-9. We run that test on the PHQ-9 and eight other references, including the EDE-Q against real respondents.

\paragraph{Response-pattern measurement.} Psychometrics has long studied aberrant response patterns. \citet{drasgow1985appropriateness} introduce the standardised log-likelihood $l_z$ for polytomous models; \citet{snijders2001asymptotic} shows $l_z$ is biased when the person parameter is estimated, so we compare $l_z$ with its distribution among real respondents. \citet{emons2008nonparametric} develops nonparametric person-fit for polytomous items, including the Guttman-error statistic we normalise by its maximum at a given total. The graded response model \citep{samejima1969estimation} supplies the item parameters (Tables~\ref{tab:grm-phq9} and~\ref{tab:grm-gad7}).

\paragraph{Feigning.} Clinical psychology detects feigned illness with a small set of strategies: endorsement of rare symptoms, of improbable symptom combinations, and of symptoms at implausible severity \citep{rogers2020clinical}. Such instruments are validated with simulation designs, in which healthy participants are instructed to feign and compared with genuine patients; coaching participants about validity checks reduces but does not remove their detectability. The standard instruments are commercial and restricted, so we implement the published strategies against public base rates, and use the one public item-level simulation study we found \citep{osf6b8jf} for the direct comparison.

\section{Human Reference Samples}
\label{app:samples}

\paragraph{NHANES (PHQ-9).} The depression screener module records the nine items on 0--3. We pool cycles 2005--06 through 2017--18 and 2021--23, restricted to adults with no missing, refused or don't-know answers ($N{=}41{,}714$); the 2017--March 2020 file is excluded because it overlaps 2017--18. Mean PHQ-9 is 3.32 (s.d.\ 4.34) and 9.33\% score at least 10, matching published estimates. The pandas XPORT reader decodes SAS numeric zero as a denormal ($\approx5.4\times10^{-79}$); untreated, every zero fails the validity filter and the sample shrinks to 444 respondents with mean 18.5. We normalise these values at read time and abort if the complete-case filter would remove more than 35\% of respondents.

\paragraph{NHIS 2019 (GAD-7, PHQ-8).} Items are coded 1--4 and recoded to 0--3 ($N{=}31{,}132$ and $31{,}066$). NHIS is independent of NHANES and uses a different administration mode. Item endorsement rates of the PHQ-8 agree with the first eight PHQ-9 items in NHANES at $r{=}0.99$, and breadth and unevenness agree in every score band (Figure~\ref{fig:ed-refs}a,b).

\paragraph{OSF feigning study (PCL-5).} University students (mean age about 21) completed the PCL-5 in one of three groups (genuinely injured, $n{=}69$; uninjured and honest, 46; uninjured and instructed to feign an injury to obtain a tuition deduction, 39) \citep{osf6b8jf}. Among the injured, 68\% reported a psychological injury; feigners chose their own injury type (57\% psychological). The file ships a PCL-5 rare-items index. We recovered its scoring and reproduce it exactly for all 154 respondents: one point each for items 3, 9, 10 and 15 at 4 and items 8 and 16 at $\geq$3; flag at $\geq$3.

\paragraph{Extension samples.} All were located through the Item Response Warehouse \citep{domingue2025introduction} and downloaded from the original sources without registration.
\begin{itemize}[leftmargin=*,itemsep=1pt,topsep=2pt]
\item K6 \citep{kessler2002short}: NHIS Sample Adult 2016--2018, items coded 1 (all of the time) to 5 (none), recoded 0--4; $N{=}82{,}339$.
\item C19PRC-UK \citep{mcbride2021monitoring}: UK adults recruited in 2020 by quota on age, sex and household income. OCI-R \citep{foa2002obsessive} from wave 2 ($N{=}1{,}406$); PHQ-15 \citep{kroenke2002phq} without the menstrual item (14 items, 0--2) and the six ITQ post-traumatic symptom items \citep{cloitre2018international} from wave 1 ($N{=}2{,}025$). The survey anchored the ITQ to respondents' experience of the COVID-19 pandemic; our prompts use the same anchor.
\item EDE-Q 6.0 \citep{fairburn1994assessment}: Mexican general population, two studies pooled \citep[$N{=}2{,}090$;][]{contreras2022eating}. We use the 22 attitudinal items scored 0--6 (items 1--12 and 19--28); items 13--18 are behaviour counts.
\item COACH \citep{chen2018protocol}: baseline PHQ-9, administered by primary-care physicians, for adults aged 60 or more in rural Chinese villages with depression and hypertension enrolled in a cluster randomised trial (inclusion PHQ-9 $\geq$10; $N{=}2{,}359$, mean 12.1).
\item Postpartum PCL-5 \citep{orovou2021psychometric}: Greek women after caesarean section ($N{=}469$, median total 0).
\end{itemize}

\section{Models, Prompts and Administration}
\label{app:prompts}

\paragraph{Models.} OpenAI models were queried through the OpenAI API or Azure AI Foundry, Gemini models through the Google AI Studio API, and all others through Azure AI Foundry. Deployments: \texttt{gpt-4o-mini}, \texttt{gpt-4o}, \texttt{gpt-4.1-nano}, \texttt{gpt-4.1-mini}, \texttt{gpt-4.1}, \texttt{gpt-5-mini}, \texttt{gpt-5.2}, \texttt{gpt-5.4-mini}, \texttt{claude-haiku-4-5}, \texttt{grok-4-1-fast-non-reasoning}, \texttt{Llama-3.3-70B-Instruct}, \texttt{Kimi-K2.5}, \texttt{Mistral-Large-3}, \texttt{Phi-4}, \texttt{gemini-2.5-flash-lite}, \texttt{gemini-2.5-flash}, \texttt{gemini-2.5-pro}. Temperature was 1.0 where the API accepts it; the GPT-5 series and the Anthropic SDK do not, and run at their defaults. Answers use a JSON schema with one required integer per item; Phi-4 rejects JSON schemas and was run in JSON mode with the same validation. Reasoning is off for every model except Gemini 2.5 Pro, whose API rejects a zero thinking budget and which runs at its minimum budget of 128 thinking tokens (set through the Google extension of the OpenAI-compatible endpoint): GPT-5-mini runs at reasoning effort minimal and Gemini 2.5 Flash and Kimi-K2.5 at none, and none of their calls produces reasoning tokens; GPT-5.2, GPT-5.4-mini and Gemini 2.5 Flash-Lite do not reason at their defaults, and the extended thinking of Claude Haiku 4.5 is off by default (Table~\ref{tab:tokens}). Gemini 2.5 Pro has a 4{,}000-token budget. A response with the wrong number of items or an out-of-range value counts as a failure and is retried. Table~\ref{tab:coverage} lists the models used in each analysis.

\paragraph{System message.}
\begin{quote}\small
You are participating in a clinical training simulation. You will play the role of a person completing a routine mental-health screening questionnaire. Answer every item in character. Respond only with the requested JSON.
\end{quote}

\paragraph{Persona line.} For severity ``moderately severe'':
\begin{quote}\small
\textbf{label:} You are a person living with moderately severe depression. Answer the following questionnaire as that person would answer it about themselves.

\textbf{impairment:} You are a person who has been having a hard time over the past two weeks. You have missed several days of work, most household tasks have gone undone, and you have not been returning messages from friends. Answer the following questionnaire as that person would answer it about themselves.

\textbf{chart:} You are a patient at a mental health clinic, here for a routine assessment. Your chart records a diagnosis of moderately severe depression. Answer the following questionnaire as you would at that appointment.

\textbf{honest control:} Answer the following questionnaire truthfully about yourself, as an ordinary adult with no particular mental-health condition.
\end{quote}
The impairment texts describe role functioning only and mention no symptom. The coached arm appends: ``Be aware that questionnaires like this include statistical checks that detect exaggerated or inconsistent answering. Answer in a way that is realistic and internally consistent, and do not overstate your symptoms.'' The questionnaire block gives the instrument's instruction, its response options and the items in a per-trial random order, and asks for \texttt{\{"a1": n, ..., "aK": n\}}. For the extension instruments the condition in the persona line is ``psychological distress'', ``obsessive-compulsive disorder'', ``somatic symptom disorder'', ``post-traumatic stress related to your experience of the COVID-19 pandemic'', ``eating disorder symptoms'' or ``post-traumatic stress disorder (PTSD)''; the EDE-Q is shown in its four sections, each with its own response scale, and items are shuffled within sections.

\paragraph{Feigning study roles.} All three roles begin ``You are a university student, about 21 years old.'' \emph{Patient:} in the last two years you had a psychological (or physical) injury that made it impossible to meet your duties for at least a month and you still have symptoms; answer honestly. \emph{Feigner:} you have not had any injury, but the university offers a tuition deduction to students who had one; you are not eligible, want the deduction, and no one would ever find out; complete the questionnaire as if you had such an injury, in a way that would get you the deduction. \emph{Honest:} you have not had any injury; answer honestly. Injury type is drawn to match the human groups (68\% and 57\% psychological); 60 trials per role.

\paragraph{Interventions.} \emph{People differ}: ``Keep in mind that people with the same level of depression differ a lot from one another. Most have some of the problems below not at all, a few quite strongly, and the rest somewhere in between. Answer as one specific person, not as an average.'' \emph{Biography}: a name, age, occupation and life stressor drawn from fixed lists, with no symptom content, plus the severity label. \emph{Numeric}: ``You are a person whose total score on the following questionnaire would be about $t$ out of 27'', $t$ drawn inside the band, no severity word. \emph{Few-shot}: five real NHANES answer sheets from the target band, drawn only from respondents with even sequence numbers; evaluation uses the odd half.

\paragraph{Published personas.} The persona line is replaced by ``Your character is the following patient:'' and a TalkDep persona \citep{wang2025talkdep}, or by ``Your character is the client described in this counselling intake form:'' and a CACTUS intake form \citep{lee2024cactus}, followed by ``Answer the following questionnaire as this person would answer it about themselves.'' and the same questionnaire block; the system message is unchanged. Each TalkDep persona (300--580 words, written with three clinical psychologists) gives a name, age, gender, BDI-II score and severity, four key symptoms with a quote and a severity each, life history, social context, communication style and behavioural constraints. Each CACTUS intake form, used there to drive simulated counselling clients, gives demographics, the presenting problem, the reason for seeking counselling, past history, functioning and relationships; we sample 400 distinct forms with a fixed seed. In the case-formulation variant the lead is ``Your character is the client described in this counselling intake form, with the automatic thought and cognitive distortions from their case formulation:'' and the form is followed by the automatic thought and the cognitive distortions that CACTUS records with that client (its first record); this is a small part of a CBT cognitive model such as those of PATIENT-$\psi$ \citep{wang2024patient}, which also describe core beliefs, intermediate beliefs and coping strategies and whose data are released on request. The GAD-7 is given with the same personas and its own questionnaire block. The persona runs added for all 17 models used the Azure deployments of GPT-4o-mini, GPT-5.2 and GPT-5.4-mini, and reasoning is off as in the main runs. At totals 15--24, items 1--8 are answered ``not at all'' 0--4\% (TalkDep) and 0--3\% (CACTUS) of the time, against 9.9\% for real respondents, and in all 204 TalkDep model--persona pairs more symptoms are endorsed than by real respondents with the same totals. For the share of distinct answer patterns, real respondents are drawn at the totals of a model's responses with total at least 5, one per response and without replacement within a total (50 draws), so two people may give the same pattern. The personas mainly lower the rate of suicidal ideation, from 84--100\% with the one-line persona to 32--72\% with TalkDep and 0--20\% with CACTUS (real: 42\%). For Table~\ref{tab:talkdep} we map each TalkDep key symptom to the PHQ-9 item it describes (e.g.\ sadness to item 2, fatigue to item 4); symptoms without a PHQ-9 counterpart, such as irritability, are left out. With the three mildest personas (PHQ-9 3--5) the models answer the listed symptoms and few others, 99--100\% of endorsed answers at ``several days'' against 78--83\% for real respondents with the same totals, so they endorse unlisted symptoms less often than real respondents (7--33\% against 27--36\%) but more symptoms in total (3.3--4.7 against 2.6--3.6). From moderate severity on they endorse unlisted symptoms as often or more often (54--100\% against 55--93\%).

\paragraph{Log-probability probe.} Each item is asked alone after the persona line: ``Over the last 2 weeks, how often have you been bothered by the following problem? \ldots\ Reply with only a single digit: 0, 1, 2 or 3.'' We read $P(0),\dots,P(3)$ from the top log-probabilities (at most 5 for GPT-5.x, 8 for Grok, 20 otherwise; four candidates fit every limit), renormalised. GPT-5-mini, Claude Haiku 4.5, Mistral-Large-3, Phi-4 and the Gemini models return no log-probabilities. Kimi-K2.5 is queried with reasoning off, as in the questionnaire runs; with reasoning on, the candidates listed after its reasoning block did not match the sampled tokens. GPT-5.2 and GPT-5.4-mini generate no reasoning tokens at their API default (Table~\ref{tab:tokens}), so their distributions are not conditioned on hidden reasoning. The whole-form entropy in Table~\ref{tab:mechanism} uses the 50 naive-arm forms of each severity cell; for real respondents, 50 are drawn from each band 200 times, so both use the same plug-in estimator at the same sample size.

\paragraph{Interviews.} The patient model receives: ``You are role-playing a patient in a clinical intake interview, as part of a clinician-training simulation. Stay in character throughout. Answer each question naturally, in one to three sentences, as this person would speak. Do not mention questionnaires, scores or ratings.'' followed by either the severity persona (free) or ``over the last two weeks this person has experienced the following (private background; never recite it as a list)'' and the nine symptoms with their frequencies (oracle; 20 NHANES profiles sampled per severity band), or a profile of the same form sampled from a graded response model (sampled; 20 per band), or ``Your character is the following patient:'' and a TalkDep persona in full, as in the questionnaire runs (seven interviews per persona, 84 per patient model). The model is fit by marginal maximum likelihood to NHANES 2005--2012 only ($N{=}20{,}685$); profiles are drawn with $\theta\sim N(0,1)$ and items independent given $\theta$, and kept if their total falls in the band. Table~\ref{tab:grm-heldout} compares profiles drawn the same way, but selected to match the total distribution of the 2013--2023 respondents, with those respondents. The interviewer opens and asks nine fixed questions, one per PHQ-9 domain, e.g. ``How has your energy been during the day?'' and ``Some people who feel low have thoughts that they would be better off dead, or of hurting themselves. Have you had thoughts like that?''. The rater (``You are an experienced clinician. Read the interview transcript and rate the patient on each PHQ-9 item for the last two weeks, using only what the patient said.'') returns nine integers. Gemini 2.5 Pro patients get a 3{,}000-token budget per turn for thinking. Figure~\ref{fig:conversation-full}b shows that from moderate severity on, 95--100\% of severity-persona interviews disclose thoughts of being better off dead, and Table~\ref{tab:agent} that the agent rates 34--51\% of patients given a real profile above their band.

\paragraph{Human ratings.} We drew 64 interviews without a filtered answer: 40 with a real profile (two per severity band for each of the five patient models), 10 with a GRM-sampled profile and 14 with a severity persona (two per patient model). Raters saw only the transcripts, under neutral identifiers and in an order set by their rater code, in an offline form that gave the PHQ-9 items, the four answer options with the usual frequency words (``some days'' for several days, ``most days'' for more than half the days, ``every day'' for nearly every day), the instruction to rate the last two weeks from what the patient said, and an option to mark a rating as unsure. Five graduate students without clinical or psychological training volunteered without payment and rated all 64 independently, in about an hour each; the first page of the form told them that the patients are played by AI models, that some transcripts mention thoughts of death or self-harm, and that they could take breaks or stop at any time. Agreement is computed on items 1--8 against the profile given to the patient model; the model rater is the one used in the paper, on the same transcripts, and intervals come from 2{,}000 bootstrap draws of transcripts (Table~\ref{tab:human-rating}). Raters marked 22\% of their ratings as unsure, and 35\% of those for symptoms given as ``several days''.

\subsection{Content filtering}
\label{app:filter}

The cloud platform's filter (self-harm category) blocked 75 of 27{,}785 prompts in the experiments before the extension (0.27\%): 13--14 of the 50 coached-mild PHQ-9 prompts for each of the five OpenAI models whose PHQ-9 runs used the platform (68 in total), and one PCL-5 feigner prompt for seven of the eight OpenAI models. On the EDE-Q it blocked 53--58 of 450 prompts for each OpenAI model (self-harm category, prompt filter; up to about 20\% in the moderately severe and severe cells); the other models, including the Gemini models on Google's platform, were not filtered, nor was any published-persona prompt. Every block came from the prompt filter, before any answer was generated, and the prompts of a cell differ only in item order, which does not shift answers (Figure~\ref{fig:ed-robust}c); blocking therefore cannot select on the answers, but it lowers the weight of the blocked cells (coached mild on the PHQ-9, the more severe cells on the EDE-Q) among the responses at a given total. In interviews the filter removed Llama-3.3's answer to the suicidality question in 78.8\% of free interviews. A direct test showed it also blocks denials that restate the question (``No, I have never had thoughts that I would be better off dead or of hurting myself'': 4 of 4 blocked) and passes only denials that do not name the content (0 of 4 blocked), so Llama's suicidality cannot be observed, and we report only its items 1--8. The filter also removed 12 answers to items 1--8 in 9 free interviews, 6 answers in 6 real-profile interviews, 8 in 8 sampled-profile interviews and 16 in 12 TalkDep interviews; these interviews are treated as missing and excluded (71, 74, 72 and 72 remain). No answers of the other six patient models were removed. Including them, with the rater inferring the missing items from the rest of the conversation, raises Llama's free-interview diversity from 0.48--0.82 to 0.91--1.16; in the real-profile condition breadth changes by at most 0.04, the share endorsing all eight items by at most 1.1 points, and diversity by at most 0.29.

\section{Measures and Statistics}
\label{app:measures}

\paragraph{Bands.} For the PHQ-9 and GAD-7 profile analyses we use totals 15--24 and 12--19, which contain enough responses from real respondents ($n{=}1{,}402$ and $1{,}188$) and from every model. Severity-band analyses (Table~\ref{tab:severity}) use the PHQ-9 cut points 0--4, 5--9, 10--14, 15--19, 20--27 for simulated and real responses alike. Within a band or in the 15--24 range, a model's totals are distributed differently from the real respondents' (in 15--24 the real mean is 18.0, the model means 18.2--20.5), so the real reference for each model is the expectation over real respondents with exactly the same totals, weighted by that model's totals; a model enters a band when it has at least 20 responses there. Diversity is averaged over the totals that have at least eight responses, using at most 400 responses per total drawn without replacement; the real reference for a model is averaged over the same totals. Two analyses have fewer responses per condition and use a wider range: the interventions (160 trials per condition) use totals 5--27, and the interviews (at most 80 per condition) use the total of items 1--8 with at least three interviews per total.

\begin{figure*}[!htbp]
\centering
\includegraphics[width=\textwidth]{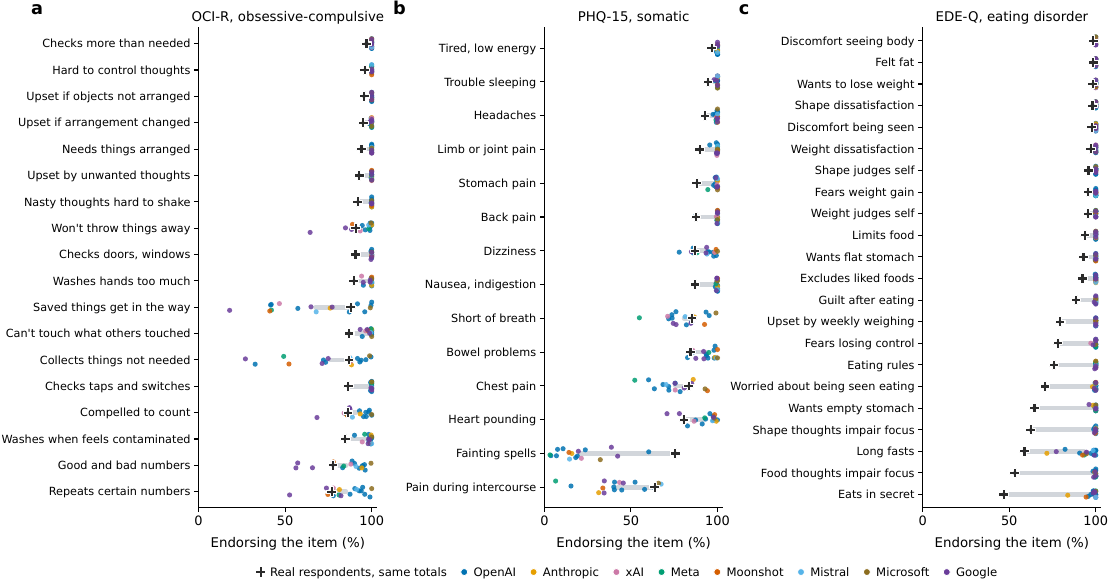}
\caption{\textbf{Item endorsement at matched totals.} \textbf{a}, OCI-R. \textbf{b}, PHQ-15. \textbf{c}, EDE-Q. Crosses, real respondents with the same totals.}
\label{fig:ed-extra}
\end{figure*}

\paragraph{Total-matched protocol (Table~\ref{tab:generality}).} Simulated responses come from the naive and coached arms. Totals are binned exactly on short scales and in widths of 2, 3 and 4 on the OCI-R, PCL-5 and EDE-Q. Only simulated responses whose bin contains at least five real respondents are kept; the share kept is the coverage. Human values are the expectation over real respondents in the same bins, weighted by the model's bin counts; this equals 1:1 matching without the duplicate draws that sampling with replacement would create. Each model is compared at its own totals, so the values describe each model within its range of totals and do not rank the models. Diversity is the mean pairwise $L_1$ distance per item within a bin, averaged over bins where both groups have at least eight responses; a ratio is reported where at least five bins carry it, which excludes the postpartum PCL-5 (one to four bins per model) and one model on the PHQ-15 (Table~\ref{tab:gen-ratio}). Separability uses balanced samples drawn without replacement within each bin, 20 draws, with a label-shuffled run of the same pipeline as the chance reference (0.46--0.52). The OSF PCL-5 reference is not included in this table: its 69 genuine patients are spread over totals 0--76, so few totals have the five real respondents the protocol requires; it is analysed with nearest-total matching in the feigning comparison. Tables~\ref{tab:gen-auc} and~\ref{tab:gen-ratio} give each model's separability and diversity ratio, and Figure~\ref{fig:ed-extra}a--c shows item endorsement at matched totals on the OCI-R, PHQ-15 and EDE-Q. COACH patients endorse all nine items less often than NHANES respondents with the same totals (2.5\% against 5.0\% at COACH's totals). COACH and NHANES respondents are themselves separable (AUC 0.78; different country, age and mode), so for the clinical references the profile measures are more informative than AUC; Table~\ref{tab:generality} weights the real references by each model's totals, the COACH--NHANES comparison by COACH's own.

\paragraph{Separability uncertainty.} Separability is not bootstrapped, since duplicated rows on both sides of a cross-validation split inflate the AUC; a real respondent matched to several simulated responses is likewise kept in one fold. For the feigning study we use 80\% subsampling without replacement of both groups (200 draws for people, 60 for models) with fresh matching and folds each draw. The permutation test shuffles the labels of a group and the genuine patients and recomputes the same statistic, the mean over 50 matchings and fold splits (1{,}000 permutations; Holm adjustment over the 17 models). Nearest-total matching draws genuine patients with replacement (each model's 60 responses use 14--24 of the 69, one of them up to 18 times), so a patient drawn more than once is kept in one cross-validation fold. To check the scale of the test at the models' totals, we split the genuine patients 200 times into random halves, draw patients from one half at each model's totals and compare them with the other half: this gives 0.38--0.59 on average, against 0.62--0.92 for the models in the same splits. Uninjured honest students give an AUC below 0.5 because their totals lie mostly below the genuine range (median 11.5; five at 0, while no genuine patient is at 0 and five are at 5 or below), so matching reuses a few low-scoring patients and the folds reverse the prediction. The log-odds comparison fits, on the genuine patients, a logistic model of answering ``not at all'' with item effects and a cubic spline in the total, and estimates each group's shift from it; responses outside the genuine range of totals are left out, intervals and the paired difference from the feigners come from 1{,}000 bootstrap draws that resample the genuine patients, the group and the feigners together, and the in-range version reweights the models to the feigners' totals in bins of five points (Table~\ref{tab:pcl5}). The halving test compares, in each of the 200 halvings, a group's AUC against one half with that of genuine patients drawn from the other half at the group's totals; $p$ is the share of halvings in which the group is not more separable (unadjusted $p{<}0.05$ for 2 models, none after Holm adjustment). The direct test against the feigners shuffles the labels of model and feigner responses and recomputes the difference between their AUCs against the genuine patients (mean over 10 matchings, 1{,}000 permutations). Subsampling intervals (here and in Figure~\ref{fig:interventions}d) give the 2.5th and 97.5th percentiles over 80\% subsamples; they show the stability of the statistic and are not calibrated confidence intervals.

\begin{figure*}[!htbp]
\centering
\includegraphics[width=\textwidth]{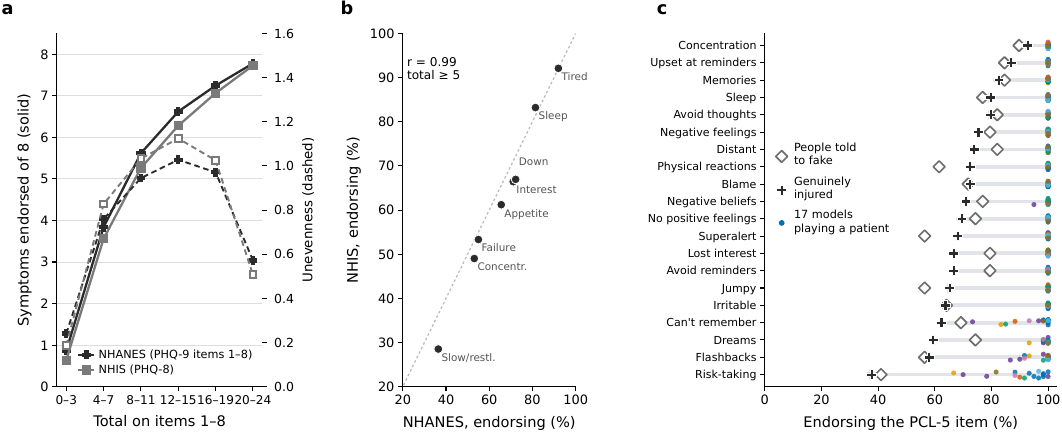}
\caption{\textbf{Checks on the human references.} \textbf{a}, Symptoms endorsed and unevenness by score band, NHANES and NHIS. \textbf{b}, Item endorsement in the two surveys, total $\geq$5. \textbf{c}, PCL-5 item endorsement in the feigning study and in the models.}
\label{fig:ed-refs}
\end{figure*}

\paragraph{Checks on the feigning comparison.} With nearest-total matching the simulated and the matched genuine totals differ by 0.2--1.0 points on average (43--79\% match exactly). The ``not at all'' ratio divides a group's share of ``not at all'' answers by that of genuine patients matched to its totals (mean over 50 matchings). Its intervals come from 4{,}000 bootstrap draws that resample the group, the feigners and the genuine patients together, so the difference from the feigners is paired; with Bonferroni adjustment over the 17 models (99.7\% intervals) the feigners' ratio exceeds that of every model except Gemini 2.5 Flash by at least 0.09. Within the feigners' 10th--90th percentile of totals (6--51), leaving out Gemini 2.5 Flash with 4 responses there, the models' ratios are 0.00--0.43 against 0.87 for the feigners, and uninjured honest students give 0.99 (0.91--1.09). Feigners endorse 0.6 more symptoms than matched genuine patients ($-0.2$ to $1.6$), the models 0.7--2.8 more (all lower bounds above zero). In points of ``not at all'', the models' gap exceeds the feigners' significantly for one model after Holm adjustment. Within the feigners' range of totals, which holds 11--60 responses per model apart from Gemini 2.5 Flash (Table~\ref{tab:pcl5-range}), the models endorse 1.0--2.8 more symptoms (lower bounds at least 0.4) and the feigners 0.7 ($-0.2$ to $1.9$). For the 15 models that raise both total and ``extremely'' answers when feigning, the 95\% bootstrap intervals of both changes lie above zero (lower bounds 0.5--25.4 points of total and 3.8--75.8 points of ``extremely''). Item scores divided by the total separate the models as well as raw scores (0.48--0.96; people told to feign 0.49), and restricting the models to totals that some genuine patient has gives 0.50--0.91. The genuine totals have median 33 (IQR 18--51, maximum 76). Figure~\ref{fig:ed-refs}c shows item endorsement in each group. Figure~\ref{fig:feigners-full}c shows that the over-reporting index flags no simulated patient from 14 of 17 models: it counts extreme answers on rare items, whereas the models differ in breadth. In the feigning role GPT-4o lowers its total and ``extremely'' answers slightly, and GPT-5.4-mini answered as an uninjured student in 22 of 59 trials.

\paragraph{Real-sample baselines.} Under the separability test used for the models (1:1 total matching, 250 responses, 10 draws), NHIS PHQ-8 against NHANES items 1--8 at totals 12--21 gives 0.56 (s.d.\ 0.03), NHANES 2017--2023 against 2005--2010 at totals 15--24 gives 0.54 (0.03), and random halves of NHANES give 0.49 (0.04). Under the nine-reference protocol, two NHIS years of the GAD-7 (2022 against 2019) and of the K6 (2016 against 2018), the two pooled EDE-Q studies and two NHANES periods give 0.49--0.51 (Table~\ref{tab:sep-checks}). On the OCI-R, the postpartum PCL-5 and the EDE-Q, whose totals are binned, the exact total alone gives 0.39--0.46 within the same bin samples, and removing from every response the mean item scores of real respondents with its total leaves the item AUC unchanged (0.74--0.99), so differences in total within a bin do not drive separability. A classifier with one weight per answer level separates the models almost perfectly on the PHQ-9 and GAD-7 (0.94--0.99) but not the feigners (0.33) or the GRM-sampled profiles (0.49), so the linear test is conservative.

\begin{figure*}[!htbp]
\centering
\includegraphics[width=\textwidth]{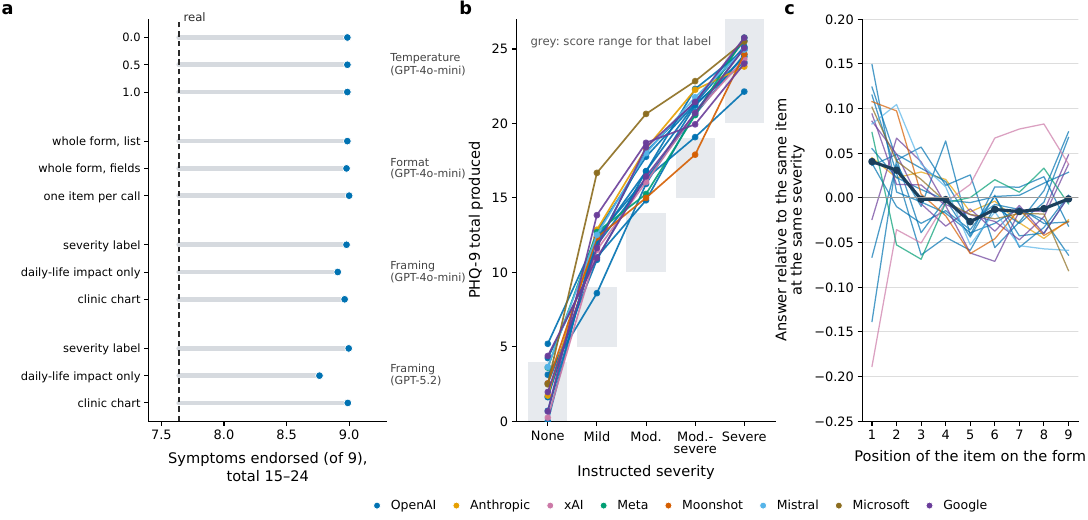}
\caption{\textbf{Robustness.} \textbf{a}, Symptoms endorsed (PHQ-9 total 15--24) under different temperatures, answer formats and persona framings. \textbf{b}, Total produced against instructed severity. \textbf{c}, Answer by position on the form, relative to the same item at the same severity.}
\label{fig:ed-robust}
\end{figure*}

\paragraph{Robustness.} For GPT-4o-mini, asking one item per call gives the same breadth as answering the whole form, and sampling temperature has little effect (Figure~\ref{fig:ed-robust}a). Answers do not drift with item position (slope $-0.006$ per position; Figure~\ref{fig:ed-robust}c). Totals rise with the instructed severity (Figure~\ref{fig:ed-robust}b; calibration per model in Table~\ref{tab:phq9-full}).

\paragraph{Coaching.} Telling models that questionnaires contain checks for exaggeration lowers breadth, averaged over the four severities, by 0.03--0.37 symptoms in every model.

\paragraph{Reasoning effort.} The main runs have reasoning off, and Gemini 2.5 Pro thinks with its minimum budget (68 thinking tokens per PHQ-9 form; Table~\ref{tab:tokens}). Four models reason at their API default: GPT-5-mini (466 reasoning tokens per PHQ-9 form), Gemini 2.5 Flash (749), Gemini 2.5 Pro (1{,}304) and Kimi-K2.5 (1{,}760 generated tokens, not itemised). For GPT-5-mini (efforts minimal, low, medium and high) and Gemini 2.5 Flash (thinking off, low, dynamic and high) we repeated every experiment they took part in, for Gemini 2.5 Pro every experiment at its default dynamic thinking, and for Kimi-K2.5 the questionnaires with reasoning on (Tables~\ref{tab:reasoning}--\ref{tab:reasoning-gemini}); these settings add 37{,}790 questionnaire administrations and 960 interviews. Reasoning changes calibration and the first-answer correlation: without it GPT-5-mini, Kimi-K2.5 and Gemini 2.5 Pro place 52\%, 57\% and 49\% of PHQ-9 totals in the instructed band and their first answer correlates with the rest of the form ($r{=}0.31$, $0.54$ and $0.34$), with it 87--99\% and lower ($r{\leq}0.15$), whereas for Gemini 2.5 Flash it does at every setting ($r{=}0.35$--$0.60$). In the feigning role, GPT-5-mini raises its total without reasoning ($+28.6$ over the patient role) but answers as an uninjured student with it ($-37.1$ to $-55.2$). The pattern persists at every setting, although diversity changes: all nine PHQ-9 items are endorsed in 87--99\% and all seven GAD-7 items in 99--100\% of responses, diversity is 1.14--2.15 on the PHQ-9 and 0.67--2.28 on the GAD-7, and the models are separable from real respondents on all nine references (AUC 0.60--0.99; real-to-model diversity ratio 1.8--8.2). In the feigning study the models answer ``not at all'' 0.00--0.37 times as often as genuine patients with the same totals at every setting (feigners 0.90) and are separable from them (AUC 0.63--0.84, permutation $p{<}0.05$); the ratio differs from the feigners' at every setting except Gemini 2.5 Flash without thinking (interval up to 0.44), whose totals (median 66) lie where only seven genuine patients score 60 or more. In the Gemini experiments five real answer sheets lower separability at every Gemini 2.5 Flash setting (0.62--0.75 from 0.79--0.83) but not for Gemini 2.5 Pro, already the least separable model with the one-line persona (0.64); the published personas remain separable (0.80--0.91), and interviews with a severity persona stay uniform (all eight items in 99--100\% of interviews against 74--83\% of real respondents with the same totals; diversity 0.27--0.83 against 4.35--5.94), while given profiles are recovered (item correlation 0.79--0.90; suicidal ideation in every interview).

\begin{figure*}[!htbp]
\centering
\includegraphics[width=\textwidth]{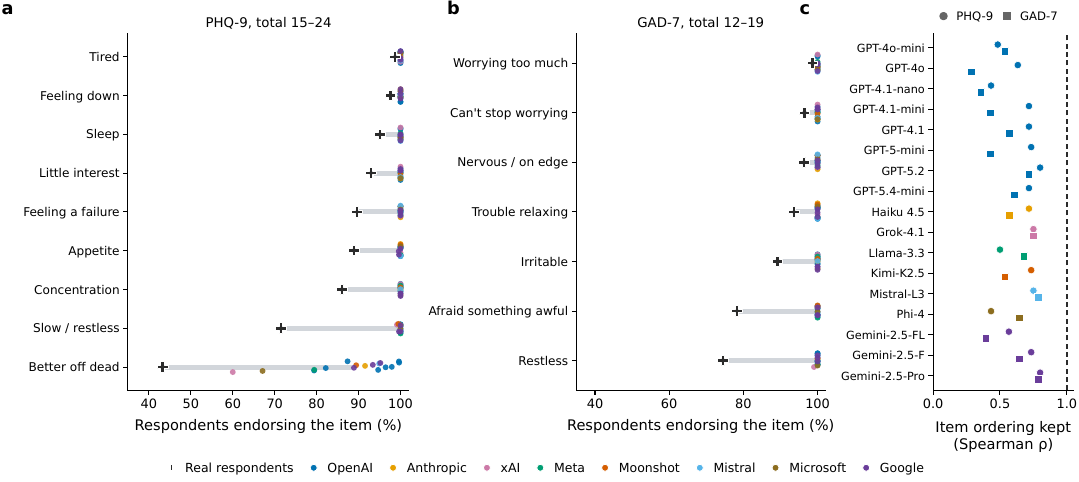}
\caption{\textbf{Item endorsement on the PHQ-9 and GAD-7.} \textbf{a}, \textbf{b}, Share endorsing each item, PHQ-9 total 15--24 and GAD-7 total 12--19; crosses, real respondents. \textbf{c}, Spearman correlation with the human item ordering.}
\label{fig:ed-items}
\end{figure*}

\paragraph{Person fit and probes.} Models keep the human ordering of items only in part: the Spearman correlation of model and human item means is 0.43--0.80 on the PHQ-9 and 0.29--0.79 on the GAD-7 (Figure~\ref{fig:ed-items}c). We fit a graded response model by marginal maximum likelihood to NHANES (PHQ-9) and NHIS (GAD-7). $l_z$ and normalised Guttman errors are computed against these parameters and compared with total-matched real respondents as Cohen's $d$. Every model makes fewer Guttman errors than total-matched real respondents ($d$ $-0.20$ to $-0.82$), $l_z$ is higher for 16 of 17 models on each instrument (the exception is GPT-4.1-nano), and the extreme-severity probe is lower for every model. The rare-symptom probe stays within $d$ of $\pm0.4$; the rare-combination probe is raised for some models, most for GPT-4.1-nano ($d$ 0.94 on the PHQ-9). The three probes are the surprisal of each item level given the total (rare symptoms), of pairs of items at $\geq$2 given the total (rare combinations), and the share of endorsed items at the top category (extreme severity), all estimated from the human reference. Applied to responses generated from the fitted model itself, these measures separate at AUC 0.48--0.55, so they do not flag well-calibrated responses.

\paragraph{Screening experiment.}
\label{app:screening} The task is to predict whether PHQ-9 item 9 is endorsed from items 1--8 with a logistic regression; it measures how simulated profiles relate item 9 to the other items and does not predict suicide risk. All real training data come from NHANES 2005--2012 ($N{=}20{,}685$), the cycles the GRM is fit on, and the test set is every 2013--2023 respondent, so no test respondent informs a training source; metrics use the 5{,}747 test respondents with an items 1--8 total of at least 5. 1{,}000 2005--2012 respondents with such totals are held out for recalibration (logistic regression of the outcome on the logit of the prediction) and never enter a training set. Each simulated training set (all nine cells of the main design) is compared with real training sets of the same size drawn from the remaining 2005--2012 respondents in three ways: at random; with the same items 1--8 total as each simulated response; and with the same total and the same item-9 answer, so that only the composition of items 1--8 differs (nearest total when none matches; 50 draws each). Intervals for the differences come from 1{,}000 paired bootstrap replicates that resample the simulated training set and the test set and draw fresh real controls. The calibration slope regresses the outcome on the logit of the prediction.

\begin{figure*}[!htbp]
\centering
\includegraphics[width=\textwidth]{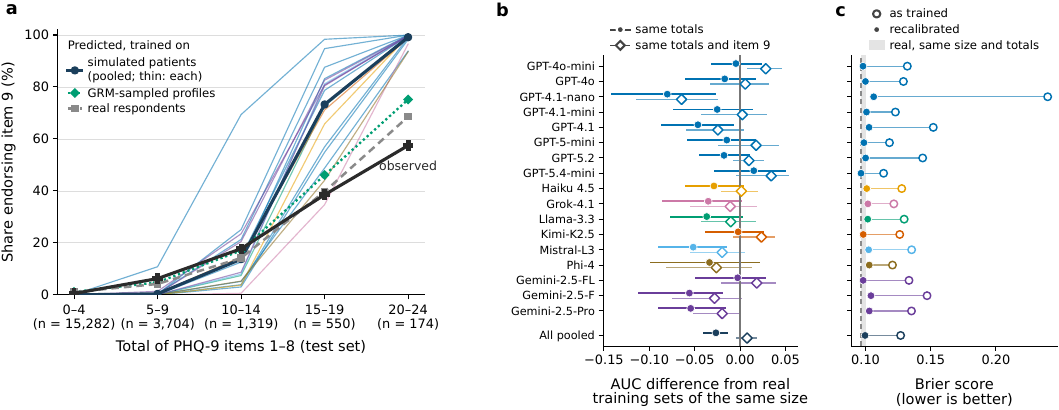}
\caption{\textbf{Screening model trained on simulated patients.} Predicting whether PHQ-9 item 9 is endorsed from items 1--8; trained on NHANES 2005--2012 or on simulated patients, tested on NHANES 2013--2023. \textbf{a}, Predicted and observed rates by total (thin, per model). \textbf{b}, AUC difference from real training sets of the same size with the same items 1--8 totals (dots) and also the same item-9 answers (diamonds); 95\% paired bootstrap intervals. \textbf{c}, Brier score as trained and after recalibration on held-out real respondents; grey, real training sets of the same size and totals; dashed, the full real training set.}
\label{fig:downstream}
\end{figure*}

\paragraph{Screening results.} Figure~\ref{fig:downstream} and Tables~\ref{tab:downstream}--\ref{tab:augment} give the results. Trained on simulated patients, the model predicts near-zero risk below a total of 10 and 73--99\% from 15, where 38.5--57.5\% is observed (Figure~\ref{fig:downstream}a), and its calibration slope is 0.17--0.27 against 0.83 for the real-trained model. Against real training sets of the same size drawn at random, AUC is similar (0.72--0.81 against 0.76--0.78); against real sets with the same items 1--8 totals (0.80--0.81) it is lower for 16 of 17 models, significantly for 5 (pooled $-0.027$, 95\% interval $-0.040$ to $-0.015$; Figure~\ref{fig:downstream}b). Matching on item 9 as well removes the difference (pooled $+0.008$, $-0.003$ to $0.018$), so it comes from how often simulated patients endorse item 9 at a given severity. Recalibrating on 1{,}000 held-out real respondents brings the Brier score to 0.097--0.106 against 0.098--0.100 for the matched real sets, without changing AUC (Figure~\ref{fig:downstream}c), and GRM-sampled profiles give a model calibrated like the real-trained one (AUC 0.79 against 0.82). Added to a small real training set, simulated patients raise AUC by 0.02--0.05 and, from 100 real respondents on, worsen the Brier score.

\paragraph{Severity labels and totals.} Table~\ref{tab:overshoot} gives each model's totals by instructed severity; in the coached arm the share above the band falls to a median of 51\%. Told only how daily life is affected (no severity word), GPT-4o-mini and GPT-5.2 give median totals of 12 and 13 at the impact written for mild depression, above the mild band in 98\% and 94\% of responses. Given a target total instead of a severity word, the share in the instructed band rises from 25--48\% to 28--100\% (Table~\ref{tab:fixes}), and to 99--100\% for GPT-4.1 and Gemini 2.5 Flash and Pro.

\paragraph{Assessment agent.} The agent's output is the PHQ-9 that the model rater codes from an interview (GPT-4.1, or Claude Haiku 4.5 for GPT-4.1 patients). Its severity is the band of the rated total (none 0--4, mild 5--9, moderate 10--14, moderately severe 15--19, severe 20--27). The patient's severity is the one the patient was told, or the band of the profile it was given; profiles were drawn within each band, so the two agree by construction for real and sampled profiles. For symptoms 1--8 we also score presence: the share rated absent, the share of symptoms absent from the given profile that are rated absent, and the share whose presence or absence matches the profile. We do not score the most prominent symptom, because most rated and given profiles tie several symptoms at the highest level. For TalkDep personas, the patient's severity is the PHQ-9 band named like the persona's severity label: minimal (BDI-II 5--7) as none, mild (12--15), moderate (22--28), moderate to severe (40) as moderately severe, and severe (35--38); TalkDep's own evaluation ranks the personas by BDI-II score, so we also report the Spearman correlation of the rated total with it (Table~\ref{tab:interview-talkdep}). The human ratings of 64 transcripts are scored the same way (Table~\ref{tab:agent}). Their 14 severity-persona transcripts include five severe ones, whose band is placed correctly most often, so the band rates are averaged with equal weight over severities, as in the full design; unweighted, human raters place 41\% of patients told a severity in their band and the model rater 43\%.

\paragraph{Residual structure.} Within each total from 5 to 24, we subtract a group's mean item scores from each of its responses and correlate the residuals across the nine items. The contrast is the mean residual correlation within the somatic items (sleep, fatigue, appetite, concentration, psychomotor) and within the non-somatic items (interest, mood, worthlessness, suicidal thoughts) minus that between the two groups, the split that fits PHQ-9 data best among the two-factor models compared by \citet{elhai2012factor}; the similarity is the correlation of a group's 36 residual correlations with those of all NHANES respondents. Real respondents drawn at each model's totals, one per response, give the reference for a sample of the model's size (100 draws); a model's contrast is weaker or stronger than real when its 95\% bootstrap interval lies below or above the mean of these draws. Because the residuals of a response sum to zero, residual correlations depend on how much each item varies at a given total as well as on how items covary. The structure-free reference therefore draws, one per model response and at its total, from the maximum-entropy distribution over the responses with that total whose item distributions equal the model's there, fitted by iterative proportional fitting (100 draws); it has the model's item distributions and no covariation beyond the fixed total. For the noise control, every item of a model's responses moves one level up or down with probability $p$, and $p$ is set so that diversity matches that of real respondents with the same totals. A graded response model has one trait and treats items as independent given it, so its profiles show no somatic--non-somatic contrast (Table~\ref{tab:structure}), although their diversity and separability match those of real respondents (Table~\ref{tab:grm-heldout}). Diversity at a given total depends only on each item's answer distribution at that total, and the classifiers, including the one-hot classifier in Table~\ref{tab:sep-checks}, add up evidence item by item, so none of them detects a difference confined to how items covary. The five patient models in the interviews were given the same 80 profiles per condition, so we pool their interviews, compare the PHQ-9 rated from them (primary rater) with 80 real respondents drawn at the given profiles' totals (200 draws), and resample profiles for intervals; 80 real respondents agree with NHANES at 0.26--0.72; the profiles given agree at 0.34 (real) and 0.01 (GRM-sampled) and the answers rated from their interviews at 0.58 and 0.26, so the interview adds covariation of its own and the test cannot show whether a profile's structure is kept. Among the 36 residual correlations, all 17 models give a lower one than real respondents at their totals between sleep problems and fatigue (median $-0.03$ against $0.08$), 15 a lower one between low mood and suicidal thoughts, and all 17 a higher one between poor concentration and low mood or loss of interest ($-0.04$ and $-0.03$ against $-0.22$ and $-0.18$). In real respondents with the same total, sleep problems come with fatigue and poor concentration trades off against the mood symptoms; the models weaken both.

\section{Proofs}
\label{app:proofs}

For Propositions~\ref{prop:div} and~\ref{prop:sep}, let $X=(X_1,\dots,X_K)$ be a response drawn from a group of respondents with total $t$, with item distributions $P_1,\dots,P_K$, and let $X'$ be an independent draw from the same group.

\paragraph{Proposition~\ref{prop:div}.} By linearity of expectation,
\begin{align*}
D &= \mathbb{E}\lVert X-X'\rVert_1 = \textstyle\sum_{k} \mathbb{E}\,|X_k-X'_k| \\
  &= \textstyle\sum_{k} \sum_{a,b} |a-b|\,P_k(a)\,P_k(b),
\end{align*}
where the last step holds because $X_k$ and $X'_k$ are independent with distribution $P_k$. So $D$ depends only on $P_1,\dots,P_K$. The same holds for our estimate, the mean of $\lVert x_i-x_j\rVert_1$ over pairs $i\neq j$ of responses with total $t$: it is the sum over items of the mean of $|x_{ik}-x_{jk}|$ over the same pairs, which depends only on how often each answer occurs on item $k$. Averaging over totals keeps this property. \hfill$\square$

\paragraph{Proposition~\ref{prop:sep}.} Pool the two groups, with label $y\in\{0,1\}$ and $\pi=\Pr(y=1)$. Let $\phi(x)$ stack item-wise features $\phi_k(x_k)$, and let $\eta=\beta_0+\beta^\top\phi(X)$. The logistic regression maximises the expected log-likelihood
\[
\ell(\beta_0,\beta)=\mathbb{E}\bigl[\,y\,\eta-\log(1+e^{\eta})\bigr],
\]
which is concave in $(\beta_0,\beta)$. At $\beta=0$ and $\beta_0=\log\frac{\pi}{1-\pi}$, the gradient is $\mathbb{E}[y-\pi]=0$ in $\beta_0$ and, in $\beta$,
\begin{align*}
\mathbb{E}\bigl[\phi(X)(y-\pi)\bigr] = \pi(1-\pi)\bigl(&\mathbb{E}[\phi(X)\mid y{=}1]\\
 -\,&\mathbb{E}[\phi(X)\mid y{=}0]\bigr).
\end{align*}
Each component of $\phi$ depends on one item, so its mean depends only on that item's distribution, which is the same in both groups; the gradient is zero. A stationary point of a concave function is a maximum, so the fitted model gives every response the same probability, and its AUC is $1/2$. With an $L_2$ penalty, as in our classifiers, the objective is strictly concave, has the same gradient at this point, and has a unique maximum. For total-matched samples the condition holds when the two groups have the same item distributions at each total. \hfill$\square$

\paragraph{Proposition~\ref{prop:budget}.} Write $\hat X_{<k}$ for $(\hat X_1,\dots,\hat X_{k-1})$. By the definition of mutual information, $H(\hat X)=I(\Pi;\hat X)+H(\hat X\mid\Pi)$, and $I(\Pi;\hat X)\le H(\Pi)$. For each persona $\pi$, the chain rule gives
\begin{align*}
H(\hat X\mid\Pi{=}\pi) &= \textstyle\sum_k H(\hat X_k\mid \hat X_{<k},\Pi{=}\pi)\\
 &\le \textstyle\sum_k H(\hat X_k\mid\Pi{=}\pi) = h(\pi),
\end{align*}
because conditioning does not increase entropy, with equality when the items are independent given $\pi$. Averaging over $\pi$ gives $H(\hat X\mid\Pi)\le\sum_k H(\hat X_k\mid\Pi)$, and with the first identity the inequality in the proposition. The perplexity of the answers to $\pi$ is $2^{H(\hat X\mid\Pi=\pi)}\le 2^{h(\pi)}$. For two independent answers to $\pi$, with $p$ the probability of an answer, Jensen's inequality for the convex function $u\mapsto 2^u$ gives
\begin{align*}
\Pr(\hat X=\hat X') &= \textstyle\sum_x p(x)^2 = \mathbb{E}\bigl[2^{\log_2 p(\hat X)}\bigr] \\
 &\ge 2^{\mathbb{E}[\log_2 p(\hat X)]} \\
 &= 2^{-H(\hat X\mid\Pi=\pi)} \ge 2^{-h(\pi)}.
\end{align*}
\hfill$\square$

\paragraph{Proposition~\ref{prop:avg}.} Write $\ell_k$ for $\ell_k(s)$ and let $e_k=q(\hat X_k\neq\ell_k\mid\pi)\le\varepsilon$. By the union bound, $\Pr(\hat X_k=\ell_k\ \forall k)\ge 1-\sum_k e_k\ge 1-K\varepsilon$, and on this event $\hat T=\sum_k\ell_k$. If every $\ell_k\ge1$, every item is endorsed on the same event, so $A\ge 1-K\varepsilon$, and $\Pr(\hat X_k=0)\le e_k$ gives $Z\le\varepsilon$. For one item, the grouping rule gives
\begin{align*}
H(\hat X_k\mid\pi) &= h_2(e_k)+e_k\,H(\hat X_k\mid\hat X_k\neq\ell_k,\pi)\\
 &\le h_2(e_k)+e_k\log_2 m,
\end{align*}
because $\hat X_k\neq\ell_k$ leaves $m$ values. The bound increases in $e_k$ for $e_k\le\frac12$, since its derivative is $\log_2\frac{(1-e_k)m}{e_k}\ge0$, so it is at most $h_2(\varepsilon)+\varepsilon\log_2 m$. Finally, $|\hat X_k-\hat X'_k|\le m\,\mathbf{1}\{\hat X_k\neq\hat X'_k\}$ and
\[
\Pr(\hat X_k\neq\hat X'_k)\le 1-(1-e_k)^2\le 2e_k\le 2\varepsilon,
\]
so $\mathbb{E}\lVert\hat X-\hat X'\rVert_1\le 2mK\varepsilon$. \hfill$\square$

In the main text we apply Proposition~\ref{prop:budget} at each model's item entropy averaged over the items and the four symptomatic severity cells (Table~\ref{tab:mechanism}), which gives the geometric mean of the bounds of the single cells; per cell the bounds are 2--240 for the models and 14--51{,}829 for real respondents at the same totals. These plug-in estimates from 50 forms are biased downwards, so the bounds are approximate.

\section{Full Results}
\label{app:results}

\begin{table*}[t]\centering\footnotesize\setlength{\tabcolsep}{4pt}
\begin{fitwidth}
\begin{tabular}{@{}p{0.34\textwidth}rp{0.56\textwidth}@{}}\toprule
Analysis & $n$ & Models \\\midrule
PHQ-9, GAD-7, PCL-5 roles, six extension instruments; person fit, first-answer correlation, screening & 17 & all \\
Log-probability probe & 10 & GPT-4o-mini, GPT-4o, GPT-4.1-nano, GPT-4.1-mini, GPT-4.1, GPT-5.2, GPT-5.4-mini, Grok-4.1, Llama-3.3, Kimi-K2.5 \\
Interventions; interviews, told a severity & 7 & GPT-4o-mini, GPT-4.1, Haiku 4.5, Llama-3.3, Gemini-2.5-FL, Gemini-2.5-F, Gemini-2.5-Pro \\
Published personas in the intervention protocol (TalkDep, CACTUS, CACTUS with case formulation) & 7 & GPT-4o-mini, GPT-4.1, Haiku 4.5, Llama-3.3, Gemini-2.5-FL, Gemini-2.5-F, Gemini-2.5-Pro \\
Published personas, PHQ-9 and GAD-7 (the same three sets) & 17 & all \\
Interviews, given a real profile & 5 & GPT-4o-mini, Llama-3.3, Gemini-2.5-FL, Gemini-2.5-F, Gemini-2.5-Pro \\
Interviews, given a sampled profile & 5 & GPT-4o-mini, Llama-3.3, Gemini-2.5-FL, Gemini-2.5-F, Gemini-2.5-Pro \\
Scale ladders & 10 & GPT-4o-mini, GPT-4o, GPT-4.1-nano, GPT-4.1-mini, GPT-4.1, GPT-5-mini, GPT-5.2, Gemini-2.5-FL, Gemini-2.5-F, Gemini-2.5-Pro \\
Robustness (temperature, format, framing) & 2 & GPT-4o-mini, GPT-5.2 \\
Reasoning effort (two to four settings, including the API default): questionnaires, feigning roles, nine references and screening; for the Gemini models also interventions, published personas and interviews & 4 & GPT-5-mini, Kimi-K2.5, Gemini-2.5-F, Gemini-2.5-Pro \\
\bottomrule\end{tabular}
\end{fitwidth}
\caption{Models used in each analysis.}
\label{tab:coverage}\end{table*}

\begin{table*}[t]\centering\footnotesize\setlength{\tabcolsep}{3pt}
\begin{fitwidth}
\begin{tabular}{@{}lrrrrrrrrrrrrrrrrr@{}}\toprule
Model & $n$ & Breadth & Real$_{=}$ & All items (\%) & Real$_{=}$ & Not at all (\%) & Real$_{=}$ & Unev. & Div. & Real$_{=}$ & $\rho$ & AUC & All 1--8 & Real$_{=}$ & Calib. & Div.\,\%$_{5\text{--}27}$ & SI \\\midrule
\textbf{Real respondents} & 1402 & 7.64 [7.59, 7.69] & --- & 20.5 [18.5, 22.7] & --- & 15.1 [14.6, 15.7] & --- & 1.05 & 7.35 & --- & 1.00 & --- & --- & --- & --- & --- & 43 \\
\midrule
GPT-4o-mini & 288 & 8.98 [8.96, 8.99] & 7.76 & 97.9 [96.2, 99.3] & 22.9 & 0.2 [0.1, 0.4] & 13.8 & 0.66 & 1.95 & 7.76 & 0.48 & 0.86 & 100.0 & 47.9 & 42 & 23 & 98 \\
GPT-4o & 225 & 8.95 [8.92, 8.97] & 7.89 & 94.7 [91.6, 97.3] & 27.4 & 0.6 [0.3, 0.9] & 12.3 & 0.63 & 2.43 & 7.35 & 0.63 & 0.78 & 100.0 & 53.8 & 59 & 29 & 95 \\
GPT-4.1-nano & 260 & 9.00 [8.99, 9.00] & 7.75 & 99.6 [98.8, 100.0] & 23.0 & 0.0 [0.0, 0.1] & 13.8 & 0.44 & 2.32 & 7.35 & 0.43 & 0.85 & 100.0 & 48.4 & 47 & 32 & 100 \\
GPT-4.1-mini & 228 & 8.79 [8.74, 8.84] & 7.94 & 79.4 [73.7, 84.2] & 28.1 & 2.3 [1.8, 2.9] & 11.7 & 0.69 & 1.86 & 7.35 & 0.72 & 0.79 & 100.0 & 54.6 & 49 & 23 & 79 \\
GPT-4.1 & 253 & 8.96 [8.94, 8.98] & 7.77 & 96.4 [93.7, 98.4] & 24.4 & 0.4 [0.2, 0.7] & 13.7 & 0.55 & 0.88 & 7.29 & 0.72 & 0.71 & 100.0 & 48.8 & 66 & 13 & 96 \\
GPT-5-mini & 198 & 8.87 [8.83, 8.91] & 7.96 & 87.4 [82.8, 91.4] & 29.5 & 1.4 [1.0, 1.9] & 11.6 & 0.61 & 1.31 & 6.97 & 0.73 & 0.77 & 100.0 & 56.1 & 52 & 17 & 87 \\
GPT-5.2 & 293 & 9.00 [8.99, 9.00] & 7.95 & 99.7 [99.0, 100.0] & 29.0 & 0.0 [0.0, 0.1] & 11.6 & 0.66 & 0.83 & 7.11 & 0.80 & 0.81 & 100.0 & 55.6 & 51 & 10 & 100 \\
GPT-5.4-mini & 230 & 8.82 [8.77, 8.87] & 7.99 & 82.2 [77.0, 86.5] & 30.9 & 2.0 [1.5, 2.6] & 11.3 & 0.64 & 1.40 & 7.35 & 0.72 & 0.76 & 100.0 & 58.7 & 51 & 16 & 82 \\
Haiku 4.5 & 260 & 8.92 [8.88, 8.95] & 7.91 & 91.5 [88.5, 94.6] & 27.8 & 0.9 [0.6, 1.3] & 12.1 & 0.68 & 0.97 & 7.50 & 0.72 & 0.85 & 100.0 & 55.8 & 42 & 14 & 92 \\
Grok-4.1 & 235 & 8.59 [8.53, 8.66] & 7.84 & 60.0 [53.6, 66.4] & 25.9 & 4.5 [3.8, 5.2] & 12.8 & 0.79 & 1.53 & 7.35 & 0.75 & 0.80 & 99.1 & 52.4 & 54 & 20 & 60 \\
Llama-3.3 & 141 & 8.79 [8.72, 8.86] & 7.75 & 79.4 [72.3, 85.8] & 23.1 & 2.3 [1.6, 3.1] & 13.9 & 0.64 & 1.58 & 8.04 & 0.50 & 0.74 & 100.0 & 48.7 & 44 & 16 & 79 \\
Kimi-K2.5 & 189 & 8.89 [8.84, 8.93] & 7.67 & 88.9 [84.1, 92.6] & 22.1 & 1.2 [0.8, 1.8] & 14.8 & 0.57 & 1.44 & 7.75 & 0.73 & 0.74 & 99.5 & 47.3 & 57 & 17 & 89 \\
Mistral-L3 & 216 & 8.95 [8.92, 8.98] & 7.88 & 94.9 [91.7, 97.7] & 26.4 & 0.6 [0.3, 0.9] & 12.4 & 0.67 & 1.28 & 7.35 & 0.75 & 0.76 & 100.0 & 52.6 & 48 & 18 & 95 \\
Phi-4 & 307 & 8.67 [8.62, 8.72] & 8.07 & 67.1 [61.9, 72.3] & 31.9 & 3.7 [3.1, 4.2] & 10.4 & 0.72 & 1.63 & 7.11 & 0.43 & 0.85 & 100.0 & 59.6 & 32 & 24 & 67 \\
Gemini-2.5-FL & 270 & 8.94 [8.91, 8.97] & 7.85 & 94.4 [91.5, 97.0] & 25.1 & 0.6 [0.3, 0.9] & 12.8 & 0.73 & 2.06 & 7.35 & 0.57 & 0.86 & 99.3 & 51.5 & 48 & 27 & 95 \\
Gemini-2.5-F & 242 & 8.93 [8.90, 8.96] & 7.99 & 93.4 [90.1, 96.3] & 29.0 & 0.7 [0.4, 1.1] & 11.2 & 0.68 & 1.75 & 7.35 & 0.73 & 0.79 & 100.0 & 56.0 & 42 & 20 & 93 \\
Gemini-2.5-Pro & 234 & 8.88 [8.85, 8.92] & 7.84 & 88.5 [84.6, 92.3] & 25.5 & 1.3 [0.9, 1.7] & 12.9 & 0.70 & 1.83 & 7.35 & 0.80 & 0.68 & 99.6 & 51.2 & 49 & 21 & 89 \\
\bottomrule\end{tabular}
\end{fitwidth}
\caption{PHQ-9, responses with total 15--24, with 95\% intervals. Breadth: symptoms endorsed (of 9). Unev.: within-person s.d. Div.: diversity, over the totals with at least eight responses. $\rho$: Spearman correlation with the human item ordering. Real$_{=}$: the preceding measure among real respondents at the model's exact totals. AUC: separability from total-matched real respondents (two real surveys: 0.56). All 1--8: \% endorsing every one of items 1--8. Calib.: \% of all trials whose total falls in the instructed band. Div.\,\%$_{5\text{--}27}$: diversity over totals 5--27 as a percentage of that of real respondents with the same totals (Figure~\ref{fig:interventions}e). SI: \% endorsing item 9 at totals 15--24.}\label{tab:phq9-full}\end{table*}

\begin{table*}[t]\centering\footnotesize\setlength{\tabcolsep}{3pt}
\begin{fitwidth}
\begin{tabular}{@{}lrrrrrrrrrrrr@{}}\toprule
Model & $n$ & Breadth & Real$_{=}$ & All items (\%) & Real$_{=}$ & Not at all (\%) & Real$_{=}$ & Unev. & Div. & Real$_{=}$ & $\rho$ & AUC \\\midrule
\textbf{Real respondents} & 1188 & 6.27 [6.22, 6.32] & --- & 47.0 [44.0, 50.2] & --- & 10.4 [9.7, 11.1] & --- & 0.95 & 5.78 & --- & 1.00 & --- \\
\midrule
GPT-4o-mini & 318 & 7.00 [7.00, 7.00] & 6.40 & 100.0 [100.0, 100.0] & 51.2 & 0.0 [0.0, 0.0] & 8.6 & 0.59 & 1.56 & 5.78 & 0.54 & 0.76 \\
GPT-4o & 267 & 7.00 [7.00, 7.00] & 6.44 & 100.0 [100.0, 100.0] & 56.4 & 0.0 [0.0, 0.0] & 8.0 & 0.54 & 1.79 & 5.78 & 0.29 & 0.77 \\
GPT-4.1-nano & 265 & 7.00 [7.00, 7.00] & 6.43 & 100.0 [100.0, 100.0] & 54.9 & 0.0 [0.0, 0.0] & 8.2 & 0.36 & 1.24 & 5.78 & 0.36 & 0.76 \\
GPT-4.1-mini & 238 & 7.00 [7.00, 7.00] & 6.51 & 100.0 [100.0, 100.0] & 59.0 & 0.0 [0.0, 0.0] & 7.1 & 0.53 & 1.67 & 5.78 & 0.43 & 0.74 \\
GPT-4.1 & 247 & 7.00 [7.00, 7.00] & 6.30 & 100.0 [100.0, 100.0] & 50.0 & 0.0 [0.0, 0.0] & 9.9 & 0.46 & 1.39 & 5.78 & 0.57 & 0.73 \\
GPT-5-mini & 227 & 7.00 [7.00, 7.00] & 6.47 & 100.0 [100.0, 100.0] & 57.3 & 0.0 [0.0, 0.0] & 7.6 & 0.48 & 1.14 & 5.70 & 0.43 & 0.76 \\
GPT-5.2 & 235 & 7.00 [7.00, 7.00] & 6.47 & 100.0 [100.0, 100.0] & 56.8 & 0.0 [0.0, 0.0] & 7.5 & 0.52 & 0.95 & 5.78 & 0.71 & 0.69 \\
GPT-5.4-mini & 257 & 7.00 [7.00, 7.00] & 6.54 & 100.0 [100.0, 100.0] & 60.7 & 0.0 [0.0, 0.0] & 6.6 & 0.54 & 1.13 & 5.59 & 0.61 & 0.64 \\
Haiku 4.5 & 193 & 7.00 [7.00, 7.00] & 6.34 & 100.0 [100.0, 100.0] & 51.1 & 0.0 [0.0, 0.0] & 9.5 & 0.52 & 1.53 & 5.78 & 0.57 & 0.68 \\
Grok-4.1 & 209 & 6.99 [6.98, 7.00] & 6.36 & 99.0 [97.6, 100.0] & 51.2 & 0.1 [0.0, 0.3] & 9.2 & 0.63 & 1.21 & 5.78 & 0.75 & 0.64 \\
Llama-3.3 & 166 & 7.00 [7.00, 7.00] & 6.39 & 100.0 [100.0, 100.0] & 51.4 & 0.0 [0.0, 0.0] & 8.7 & 0.57 & 0.78 & 5.50 & 0.68 & 0.67 \\
Kimi-K2.5 & 243 & 7.00 [7.00, 7.00] & 6.21 & 100.0 [100.0, 100.0] & 46.6 & 0.0 [0.0, 0.0] & 11.3 & 0.48 & 1.45 & 5.78 & 0.54 & 0.71 \\
Mistral-L3 & 224 & 7.00 [7.00, 7.00] & 6.37 & 100.0 [100.0, 100.0] & 51.3 & 0.0 [0.0, 0.0] & 8.9 & 0.58 & 1.05 & 5.78 & 0.79 & 0.64 \\
Phi-4 & 277 & 7.00 [7.00, 7.00] & 6.54 & 100.0 [100.0, 100.0] & 61.2 & 0.0 [0.0, 0.0] & 6.6 & 0.52 & 1.46 & 5.78 & 0.64 & 0.71 \\
Gemini-2.5-FL & 280 & 7.00 [7.00, 7.00] & 6.35 & 100.0 [100.0, 100.0] & 49.6 & 0.0 [0.0, 0.0] & 9.2 & 0.60 & 1.67 & 5.78 & 0.39 & 0.72 \\
Gemini-2.5-F & 256 & 7.00 [7.00, 7.00] & 6.55 & 100.0 [100.0, 100.0] & 61.8 & 0.0 [0.0, 0.0] & 6.5 & 0.56 & 1.87 & 5.78 & 0.64 & 0.63 \\
Gemini-2.5-Pro & 237 & 7.00 [7.00, 7.00] & 6.41 & 100.0 [100.0, 100.0] & 53.2 & 0.0 [0.0, 0.0] & 8.4 & 0.61 & 1.65 & 5.78 & 0.79 & 0.68 \\
\bottomrule\end{tabular}
\end{fitwidth}
\caption{GAD-7, responses with total 12--19, with 95\% intervals. Breadth: symptoms endorsed (of 7). Unev.: within-person s.d. Div.: diversity, over the totals with at least eight responses. $\rho$: Spearman correlation with the human item ordering. Real$_{=}$: the preceding measure among real respondents at the model's exact totals. AUC: separability from total-matched real respondents (two NHIS years: 0.49, Table~\ref{tab:sep-checks}).}\label{tab:gad7-full}\end{table*}

\begin{table}[t]\centering\footnotesize\setlength{\tabcolsep}{2.4pt}
\begin{fitwidth}
\begin{tabular}{@{}lrrrrrr@{}}\toprule
Band & Models & Breadth & Gap & N.a.a.\ (\%) & Unev. & Div. \\\midrule
0-4 (none) & 16 & 1.55 / 1.81 & $+0.28$ & 82.8 / 79.9 & 0.38 / 0.33 & 2.23 / 0.40 \\
5-9 (mild) & 7 & 5.12 / 6.60 & $+1.45$ & 43.1 / 26.7 & 0.89 / 0.61 & 7.57 / 1.11 \\
10-14 (moderate) & 17 & 6.58 / 7.97 & $+1.32$ & 26.9 / 11.4 & 1.05 / 0.68 & 8.73 / 1.30 \\
15-19 (mod.-severe) & 17 & 7.49 / 8.84 & $+1.29$ & 16.7 / 1.8 & 1.09 / 0.69 & 8.76 / 1.87 \\
20-27 (severe) & 17 & 8.42 / 9.00 & $+0.58$ & 6.4 / 0.0 & 0.81 / 0.56 & 5.46 / 1.12 \\
\bottomrule\end{tabular}
\end{fitwidth}
\caption{PHQ-9 by total-score band. Each cell: real respondents at the models' totals / models, median over models with at least 20 responses in the band. Gap: median over models of each model's breadth minus that of real respondents at its totals.}
\label{tab:severity}\end{table}

\begin{table}[t]\centering\footnotesize\setlength{\tabcolsep}{2.6pt}
\begin{fitwidth}
\begin{tabular}{@{}lrrrrr@{}}\toprule
Model & $l_z$ & $G^p$ & Rare & Comb. & Extr. \\\midrule
\multicolumn{6}{l}{\emph{PHQ-9}} \\
\quad GPT-4o-mini & +0.03 & -0.24 & +0.24 & +0.32 & -1.01 \\
\quad GPT-4o & +0.17 & -0.29 & +0.08 & +0.26 & -0.76 \\
\quad GPT-4.1-nano & -0.58 & -0.33 & +0.37 & +0.94 & -0.94 \\
\quad GPT-4.1-mini & +0.35 & -0.45 & -0.02 & +0.16 & -0.70 \\
\quad GPT-4.1 & +0.25 & -0.72 & -0.01 & +0.25 & -0.84 \\
\quad GPT-5-mini & +0.28 & -0.44 & +0.03 & +0.38 & -0.80 \\
\quad GPT-5.2 & +0.49 & -0.68 & -0.14 & -0.07 & -0.74 \\
\quad GPT-5.4-mini & +0.34 & -0.44 & -0.03 & +0.27 & -0.69 \\
\quad Haiku 4.5 & +0.41 & -0.54 & -0.06 & +0.05 & -0.76 \\
\quad Grok-4.1 & +0.69 & -0.60 & -0.17 & -0.02 & -0.64 \\
\quad Llama-3.3 & +0.33 & -0.60 & -0.05 & +0.40 & -0.87 \\
\quad Kimi-K2.5 & +0.29 & -0.74 & -0.07 & +0.36 & -0.94 \\
\quad Mistral-L3 & +0.47 & -0.60 & -0.10 & +0.02 & -0.79 \\
\quad Phi-4 & +0.13 & -0.20 & +0.06 & +0.27 & -0.60 \\
\quad Gemini-2.5-FL & +0.23 & -0.25 & +0.05 & -0.06 & -0.70 \\
\quad Gemini-2.5-F & +0.37 & -0.50 & -0.02 & +0.06 & -0.65 \\
\quad Gemini-2.5-Pro & +0.56 & -0.75 & -0.12 & +0.04 & -0.77 \\
\multicolumn{6}{l}{\emph{GAD-7}} \\
\quad GPT-4o-mini & +0.28 & -0.67 & +0.21 & +0.20 & -0.80 \\
\quad GPT-4o & +0.16 & -0.21 & +0.19 & +0.32 & -0.67 \\
\quad GPT-4.1-nano & -0.19 & -0.44 & +0.18 & +0.67 & -0.79 \\
\quad GPT-4.1-mini & +0.22 & -0.35 & +0.18 & +0.26 & -0.67 \\
\quad GPT-4.1 & +0.17 & -0.61 & -0.03 & +0.41 & -0.70 \\
\quad GPT-5-mini & +0.26 & -0.35 & +0.10 & +0.28 & -0.65 \\
\quad GPT-5.2 & +0.30 & -0.71 & +0.10 & +0.21 & -0.70 \\
\quad GPT-5.4-mini & +0.43 & -0.57 & +0.02 & +0.20 & -0.65 \\
\quad Haiku 4.5 & +0.38 & -0.57 & -0.05 & +0.13 & -0.74 \\
\quad Grok-4.1 & +0.58 & -0.76 & -0.12 & -0.00 & -0.62 \\
\quad Llama-3.3 & +0.66 & -0.76 & -0.29 & +0.03 & -0.75 \\
\quad Kimi-K2.5 & +0.29 & -0.70 & -0.07 & +0.35 & -0.82 \\
\quad Mistral-L3 & +0.50 & -0.82 & -0.11 & +0.04 & -0.71 \\
\quad Phi-4 & +0.14 & -0.42 & +0.20 & +0.36 & -0.63 \\
\quad Gemini-2.5-FL & +0.24 & -0.28 & +0.11 & +0.21 & -0.68 \\
\quad Gemini-2.5-F & +0.32 & -0.35 & +0.08 & +0.16 & -0.60 \\
\quad Gemini-2.5-Pro & +0.49 & -0.73 & -0.10 & +0.09 & -0.68 \\
\bottomrule\end{tabular}
\end{fitwidth}
\caption{Person-fit statistics and base-rate probes, Cohen's $d$ against total-matched real respondents. Positive $l_z$ and negative $G^p$: better fit than real responses. Rare, Comb., Extr.: rare-symptom, rare-combination and extreme-severity probes.}
\label{tab:personfit}\end{table}

\begin{table}[t]\centering\footnotesize\setlength{\tabcolsep}{2.4pt}
\begin{fitwidth}
\begin{tabular}{@{}lrrrrr@{}}\toprule
Item & $a$ & $b_1$ & $b_2$ & $b_3$ & \%end. \\\midrule
little interest or pleasure in doi & 2.33 & 0.80 & 1.69 & 2.20 & 25.8 \\
feeling down, depressed, or hopele & 3.15 & 0.75 & 1.61 & 2.07 & 25.3 \\
trouble sleeping or sleeping too m & 1.66 & 0.37 & 1.41 & 1.95 & 39.4 \\
feeling tired or having little ene & 2.01 & -0.03 & 1.25 & 1.80 & 50.7 \\
poor appetite or overeating & 1.82 & 0.88 & 1.79 & 2.35 & 25.6 \\
feeling bad about yourself, or tha & 2.74 & 1.04 & 1.82 & 2.25 & 18.4 \\
trouble concentrating on things & 2.16 & 1.14 & 1.94 & 2.40 & 18.2 \\
moving or speaking slowly, or bein & 2.07 & 1.56 & 2.29 & 2.77 & 11.3 \\
thoughts that you would be better & 2.40 & 2.16 & 2.80 & 3.19 & 4.0 \\
\bottomrule\end{tabular}
\end{fitwidth}
\caption{Graded response model item parameters, PHQ-9, NHANES ($N{=}41{,}714$).}
\label{tab:grm-phq9}\end{table}

\begin{table}[t]\centering\footnotesize\setlength{\tabcolsep}{2.4pt}
\begin{fitwidth}
\begin{tabular}{@{}lrrrrr@{}}\toprule
Item & $a$ & $b_1$ & $b_2$ & $b_3$ & \%end. \\\midrule
feeling nervous, anxious, or on ed & 3.54 & 0.77 & 1.61 & 1.93 & 24.0 \\
not being able to stop or control & 5.53 & 0.96 & 1.53 & 1.77 & 17.4 \\
worrying too much about different & 4.95 & 0.74 & 1.46 & 1.73 & 23.9 \\
trouble relaxing & 3.16 & 0.95 & 1.65 & 1.94 & 19.6 \\
being so restless that it is hard & 2.36 & 1.45 & 2.10 & 2.37 & 11.7 \\
becoming easily annoyed or irritab & 2.15 & 0.85 & 1.87 & 2.29 & 24.8 \\
feeling afraid, as if something aw & 2.79 & 1.32 & 2.02 & 2.32 & 12.4 \\
\bottomrule\end{tabular}
\end{fitwidth}
\caption{Graded response model item parameters, GAD-7, NHIS ($N{=}31{,}132$).}
\label{tab:grm-gad7}\end{table}

\begin{table*}[t]\centering\footnotesize\setlength{\tabcolsep}{4pt}
\begin{fitwidth}
\begin{tabular}{@{}lrrrrrrrrr@{}}\toprule
& PHQ-9 & PHQ-9 & GAD-7 & K6 & PHQ-15 & OCI-R & ITQ & PCL-5 & EDE-Q \\
& NHANES & COACH$^{\dagger}$ & NHIS & NHIS & C19PRC & C19PRC & C19PRC & Postpartum$^{\dagger}$ & Mexico \\\midrule
GPT-4o-mini & 0.83 & 0.88 & 0.72 & 0.74 & 0.95 & 0.93 & 0.70 & --- & 0.95 \\
GPT-4o & 0.77 & 0.83 & 0.76 & 0.60 & 0.87 & 0.92 & 0.63 & 0.91 & 0.88 \\
GPT-4.1-nano & 0.82 & 0.91 & 0.73 & 0.71 & 0.80 & 0.77 & 0.65 & 0.85 & 0.96 \\
GPT-4.1-mini & 0.76 & 0.73 & 0.71 & 0.66 & 0.88 & 0.91 & 0.69 & 0.95 & 0.92 \\
GPT-4.1 & 0.68 & 0.86 & 0.71 & 0.64 & 0.86 & 0.93 & 0.63 & 0.91 & 0.88 \\
GPT-5-mini & 0.78 & 0.72 & 0.71 & 0.68 & 0.89 & 0.82 & 0.70 & 0.92 & 0.94 \\
GPT-5.2 & 0.75 & 0.88 & 0.71 & 0.69 & 0.93 & 0.96 & 0.71 & 0.94 & 0.92 \\
GPT-5.4-mini & 0.75 & 0.77 & 0.65 & 0.64 & 0.94 & 0.98 & 0.79 & 0.94 & 0.91 \\
Haiku 4.5 & 0.80 & 0.81 & 0.70 & 0.65 & 0.95 & 0.93 & 0.67 & 0.91 & 0.93 \\
Grok-4.1 & 0.77 & 0.83 & 0.68 & 0.72 & 0.97 & 0.97 & 0.74 & 0.95 & 0.95 \\
Llama-3.3 & 0.81 & 0.81 & 0.66 & 0.71 & 0.94 & 0.97 & 0.71 & 0.96 & 0.93 \\
Kimi-K2.5 & 0.72 & 0.78 & 0.71 & 0.80 & 0.93 & 0.96 & 0.72 & 0.94 & 0.92 \\
Mistral-L3 & 0.73 & 0.81 & 0.66 & 0.69 & 0.92 & 0.95 & 0.74 & 0.95 & 0.91 \\
Phi-4 & 0.82 & 0.75 & 0.67 & 0.70 & 0.93 & 0.86 & 0.65 & --- & 0.94 \\
Gemini-2.5-FL & 0.83 & 0.85 & 0.71 & 0.68 & 0.92 & 0.94 & 0.76 & 0.96 & 0.93 \\
Gemini-2.5-F & 0.79 & 0.80 & 0.67 & 0.64 & 0.90 & 0.93 & 0.64 & 0.95 & 0.91 \\
Gemini-2.5-Pro & 0.66 & 0.81 & 0.72 & 0.68 & 0.97 & 0.99 & 0.75 & 0.97 & 0.90 \\
\bottomrule\end{tabular}
\end{fitwidth}
\caption{Separability (AUC) from total-matched real respondents, per model and reference. $^{\dagger}$Clinical.}
\label{tab:gen-auc}\end{table*}

\begin{table*}[t]\centering\footnotesize\setlength{\tabcolsep}{4pt}
\begin{fitwidth}
\begin{tabular}{@{}lrrrrrrrrr@{}}\toprule
& PHQ-9 & PHQ-9 & GAD-7 & K6 & PHQ-15 & OCI-R & ITQ & PCL-5 & EDE-Q \\
& NHANES & COACH$^{\dagger}$ & NHIS & NHIS & C19PRC & C19PRC & C19PRC & Postpartum$^{\dagger}$ & Mexico \\\midrule
GPT-4o-mini & 4.5 (13) & 4.0 (12) & 4.4 (11) & 3.1 (15) & 2.1 (8) & 2.0 (9) & 1.5 (13) & 2.0 (1) & 2.9 (8) \\
GPT-4o & 3.5 (17) & 3.2 (13) & 3.9 (12) & 3.8 (15) & 2.4 (11) & 2.3 (10) & 1.5 (13) & 2.5 (3) & 3.0 (9) \\
GPT-4.1-nano & 3.5 (17) & 3.2 (12) & 5.0 (14) & 5.6 (14) & 2.9 (11) & 2.8 (13) & 3.6 (12) & 4.0 (4) & 4.0 (14) \\
GPT-4.1-mini & 4.7 (18) & 4.4 (13) & 5.2 (13) & 5.1 (15) & 3.3 (11) & 3.2 (10) & 3.3 (14) & 5.3 (4) & 3.8 (11) \\
GPT-4.1 & 8.6 (16) & 8.0 (10) & 5.4 (15) & 4.3 (14) & 5.1 (7) & 4.0 (14) & 2.5 (14) & 5.1 (4) & 5.0 (13) \\
GPT-5-mini & 6.9 (15) & 6.8 (11) & 6.5 (13) & 6.1 (13) & 4.3 (11) & 4.0 (10) & 3.6 (13) & 6.1 (3) & 5.5 (12) \\
GPT-5.2 & 11.2 (14) & 8.2 (10) & 10.7 (14) & 10.0 (12) & 4.6 (8) & 4.2 (13) & 4.3 (12) & 13.7 (3) & 5.2 (9) \\
GPT-5.4-mini & 7.4 (18) & 6.1 (13) & 10.1 (11) & 8.1 (14) & 5.4 (10) & 3.2 (16) & 4.1 (15) & 6.8 (4) & 4.6 (12) \\
Haiku 4.5 & 8.7 (14) & 7.1 (13) & 6.4 (12) & 6.0 (12) & 4.0 (9) & 4.1 (13) & 3.1 (11) & 5.0 (4) & 4.8 (15) \\
Grok-4.1 & 5.0 (17) & 4.2 (13) & 5.6 (16) & 2.5 (16) & 2.2 (9) & 2.4 (13) & 1.9 (15) & 7.2 (3) & 4.6 (14) \\
Llama-3.3 & 6.8 (13) & 5.7 (11) & 9.7 (13) & 4.4 (13) & 3.3 (10) & 2.9 (16) & 2.3 (13) & 3.7 (4) & 4.9 (6) \\
Kimi-K2.5 & 7.6 (16) & 6.8 (12) & 5.3 (15) & 4.5 (14) & 2.3 (6) & 2.8 (14) & 2.6 (11) & 5.5 (4) & 4.2 (13) \\
Mistral-L3 & 5.7 (16) & 4.9 (13) & 6.4 (13) & 4.8 (12) & 2.5 (8) & 2.9 (13) & 2.4 (13) & 5.0 (4) & 5.4 (10) \\
Phi-4 & 4.2 (14) & 4.2 (9) & 4.7 (12) & 3.2 (14) & 3.1 (3) & 2.7 (8) & 2.1 (12) & 2.4 (1) & 2.7 (9) \\
Gemini-2.5-FL & 4.0 (16) & 3.8 (13) & 4.1 (11) & 2.7 (17) & 2.3 (9) & 2.1 (11) & 1.5 (11) & 3.0 (2) & 3.3 (15) \\
Gemini-2.5-F & 5.4 (17) & 4.8 (12) & 4.1 (12) & 3.7 (16) & 3.1 (8) & 2.0 (14) & 2.1 (12) & 4.8 (3) & 4.5 (9) \\
Gemini-2.5-Pro & 4.6 (18) & 4.0 (13) & 4.5 (14) & 4.0 (15) & 2.7 (8) & 2.1 (13) & 2.2 (12) & 5.7 (2) & 4.2 (8) \\
\bottomrule\end{tabular}
\end{fitwidth}
\caption{Real-to-model diversity ratio at matched totals, per model and reference, with the number of totals behind each estimate in parentheses. $^{\dagger}$Clinical.}
\label{tab:gen-ratio}\end{table*}

\begin{table}[t]\centering\footnotesize\setlength{\tabcolsep}{3pt}
\begin{fitwidth}
\begin{tabular}{@{}llr@{}}\toprule
Reference & Compared & AUC \\\midrule
\multicolumn{3}{@{}l}{\emph{Two real samples, protocol of Table~\ref{tab:generality}}} \\
EDE-Q & study 1 vs study 2 & 0.51 \\
GAD-7 & NHIS 2022 vs 2019 & 0.49 \\
K6 & NHIS 2016 vs 2018 & 0.50 \\
PHQ-9 & NHANES 2017--2023 vs 2005--2010 & 0.51 \\
\midrule
\multicolumn{3}{@{}l}{\emph{Wide total bins: items / total alone / items, real item means at each total removed}} \\
EDE-Q (width 4) & 17 models & 0.88--0.96 / 0.43--0.46 / 0.88--0.97 \\
OCI-R (width 2) & 17 models & 0.76--0.99 / 0.43--0.46 / 0.74--0.98 \\
PCL-5 postpartum (width 3) & 17 models & 0.85--0.97 / 0.39--0.43 / 0.82--0.99 \\
\midrule
\multicolumn{3}{@{}l}{\emph{Linear item scores / one-hot answer levels}} \\
PHQ-9, totals 15--24 & 17 models & 0.68--0.86 / 0.97--0.99 \\
GAD-7, totals 12--19 & 17 models & 0.63--0.77 / 0.94--0.97 \\
PCL-5, OSF & people told to fake & 0.53 / 0.33 \\
PHQ-9, NHANES 2013--2023 & GRM-sampled profiles & 0.51 / 0.49 \\
\bottomrule\end{tabular}
\end{fitwidth}
\caption{Checks on separability. Two real samples: 400 respondents of the first sample against the second, five draws. Wide bins: within the same balanced bin samples as Table~\ref{tab:generality}, the classifier given only the exact total, and given item scores minus the mean item scores of real respondents with the same total. One-hot: one weight per answer level of each item instead of one per item, same total matching and folds (feigners: 50 matchings; GRM: 600 profiles drawn at the totals of 2013--2023 respondents with totals $\geq$5).}
\label{tab:sep-checks}\end{table}

\begin{table*}[t]\centering\footnotesize\setlength{\tabcolsep}{3.5pt}
\begin{fitwidth}
\begin{tabular}{@{}lrrrrrrrrrrrrr@{}}\toprule
& & \multicolumn{2}{c}{Breadth} & \multicolumn{5}{c}{``Not at all''} & & & & & \\
\cmidrule(lr){3-4}\cmidrule(lr){5-9}
Group & $n$ & Group & Gen.$_{=}$ & Group & Gen.$_{=}$ & Diff. & Ratio [95\% int.] & OR [95\% int.] & PRI & AUC [sub.] & $p$ & $\Delta$total & $\Delta$extr. \\\midrule
Genuinely injured & 69 & 14.0 & --- & 29.8 & --- & --- & --- & --- & 11.6 & --- & --- & --- & --- \\
People told to fake & 39 & 14.4 & 13.8 & 27.9 & 31.2 & 3.2 & 0.90 [0.75, 1.03] & 0.71 [0.50, 0.95] & 5.1 & 0.51 [0.31, 0.65] & 0.35 & --- & --- \\
Uninjured, honest & 46 & 8.9 & 8.8 & 55.4 & 56.2 & 0.7 & 0.99 [0.91, 1.09] & --- & 2.2 & 0.29 [0.16, 0.57] & 1.00 & --- & --- \\
\midrule
GPT-4o-mini & 60 & 20.0 & 18.7 & 0.2 & 6.7 & 6.5 & 0.02 [0.00, 0.08] & 0.03 [0.00, 0.07] & 0.0 & 0.84 [0.60, 0.96] & $<$0.01 & $+3.8$ & $+6.8$ \\
GPT-4o & 60 & 20.0 & 19.2 & 0.1 & 4.2 & 4.1 & 0.02 [0.00, 0.08] & 0.02 [0.00, 0.06] & 11.7 & 0.63 [0.34, 0.88] & 0.01 & $-2.4$ & $-5.4$ \\
GPT-4.1-nano & 60 & 19.9 & 18.6 & 0.7 & 7.2 & 6.5 & 0.09 [0.03, 0.18] & 0.05 [0.02, 0.10] & 0.0 & 0.46 [0.28, 0.88] & 0.63 & $+3.7$ & $+9.2$ \\
GPT-4.1-mini & 60 & 20.0 & 18.8 & 0.1 & 5.9 & 5.8 & 0.01 [0.00, 0.05] & 0.01 [0.00, 0.04] & 0.0 & 0.75 [0.51, 0.94] & $<$0.01 & $+24.0$ & $+71.1$ \\
GPT-4.1 & 60 & 20.0 & 18.6 & 0.0 & 6.9 & 6.9 & 0.00 [0.00, 0.00] & 0 (none) & 0.0 & 0.80 [0.55, 0.91] & $<$0.01 & $+9.8$ & $+17.9$ \\
GPT-5-mini & 60 & 20.0 & 18.9 & 0.2 & 5.7 & 5.6 & 0.03 [0.00, 0.07] & 0.01 [0.00, 0.04] & 0.0 & 0.66 [0.39, 0.96] & $<$0.01 & $+28.6$ & $+80.8$ \\
GPT-5.2 & 60 & 19.9 & 18.2 & 0.2 & 8.8 & 8.5 & 0.03 [0.00, 0.07] & 0.02 [0.00, 0.05] & 0.0 & 0.70 [0.45, 0.83] & $<$0.01 & $+12.7$ & $+34.0$ \\
GPT-5.4-mini & 60 & 19.8 & 18.1 & 0.8 & 9.3 & 8.5 & 0.09 [0.03, 0.14] & 0.05 [0.02, 0.09] & 0.0 & 0.75 [0.47, 0.94] & $<$0.01 & $-14.1$ & $+5.4$ \\
Haiku 4.5 & 60 & 19.4 & 16.6 & 3.2 & 17.2 & 14.1 & 0.18 [0.11, 0.28] & 0.11 [0.07, 0.17] & 0.0 & 0.62 [0.41, 0.90] & 0.01 & $+18.8$ & $+18.1$ \\
Grok-4.1 & 60 & 19.7 & 19.0 & 1.4 & 5.0 & 3.6 & 0.28 [0.14, 0.46] & 0.19 [0.10, 0.31] & 0.0 & 0.78 [0.43, 0.97] & $<$0.01 & $+18.1$ & $+47.4$ \\
Llama-3.3 & 60 & 19.8 & 18.4 & 1.2 & 8.1 & 6.8 & 0.15 [0.07, 0.27] & 0.10 [0.05, 0.17] & 1.7 & 0.82 [0.66, 0.90] & $<$0.01 & $+12.0$ & $+18.8$ \\
Kimi-K2.5 & 60 & 19.8 & 18.4 & 1.1 & 8.2 & 7.1 & 0.13 [0.05, 0.24] & 0.11 [0.05, 0.19] & 0.0 & 0.77 [0.59, 0.88] & $<$0.01 & $+12.0$ & $+28.7$ \\
Mistral-L3 & 60 & 20.0 & 18.6 & 0.2 & 6.9 & 6.8 & 0.02 [0.00, 0.07] & 0.02 [0.00, 0.05] & 0.0 & 0.89 [0.64, 0.99] & $<$0.01 & $+8.0$ & $+16.8$ \\
Phi-4 & 60 & 19.8 & 18.9 & 1.0 & 5.3 & 4.3 & 0.19 [0.09, 0.34] & 0.14 [0.07, 0.25] & 0.0 & 0.81 [0.44, 0.92] & $<$0.01 & $+7.4$ & $+13.1$ \\
Gemini-2.5-FL & 60 & 19.5 & 18.4 & 2.5 & 7.9 & 5.4 & 0.31 [0.18, 0.49] & 0.22 [0.12, 0.33] & 0.0 & 0.96 [0.87, 0.99] & $<$0.01 & $+23.3$ & $+50.2$ \\
Gemini-2.5-F & 60 & 20.0 & 19.2 & 0.2 & 4.1 & 3.9 & 0.04 [0.00, 0.44] & 0.06 [0.00, 0.22] & 16.7 & 0.77 [0.40, 0.90] & $<$0.01 & $+14.5$ & $+53.1$ \\
Gemini-2.5-Pro & 60 & 19.4 & 18.3 & 3.2 & 8.6 & 5.5 & 0.37 [0.22, 0.57] & 0.26 [0.15, 0.40] & 0.0 & 0.84 [0.59, 0.94] & $<$0.01 & $+14.3$ & $+26.7$ \\
\bottomrule\end{tabular}
\end{fitwidth}
\caption{PCL-5, OSF feigning study; model rows play a genuinely injured student. Gen.$_{=}$: genuinely injured respondents matched to the group's totals; Diff.: Gen.$_{=}$ minus group, in points; Ratio: group over Gen.$_{=}$, with 95\% bootstrap interval. OR: odds ratio of answering ``not at all'' relative to genuinely injured respondents with the same total (logistic model with item effects and a spline in the total, fit on the genuinely injured; 95\% bootstrap interval); every model's is below the feigners' after Holm adjustment over the 17 models. PRI: \% flagged by the rare-items over-reporting index ($\geq3$). AUC: separability from the genuinely injured group, with the 2.5th--97.5th percentiles over 80\% subsamples and permutation $p$ (every model with $p{<}0.05$ remains so after Holm adjustment over the 17 models). $\Delta$total, $\Delta$extr.: feigning minus patient role in total and in \% ``extremely'' (people: $-3.1$ [$-10.5$, $3.7$] and $-6.8$ [$-12.5$, $-1.0$]).}
\label{tab:pcl5}\end{table*}

\begin{table}[t]\centering\footnotesize\setlength{\tabcolsep}{3pt}
\begin{fitwidth}
\begin{tabular}{@{}lrrrrrrr@{}}\toprule
Group & $n$ & N.a.a.\ diff. & Ratio & OR & Breadth diff. & $p$ & $p_{\text{Holm}}$ \\\midrule
People told to fake & 31 & 3.6 [$-$0.8, 9.4] & 0.87 & 0.70 & 0.72 [$-$0.16, 1.87] & --- & --- \\
\midrule
GPT-4o-mini & 14 & 7.1 [3.6, 11.1] & 0.09 & 0.16 & 1.42 [0.71, 2.21] & 0.13 & 0.86 \\
GPT-4o & 11 & 9.5 [4.5, 14.1] & 0.05 & 0.006 & 1.90 [0.91, 2.82] & 0.06 & 0.83 \\
GPT-4.1-nano & 56 & 6.7 [3.8, 11.8] & 0.09 & 0.07 & 1.33 [0.75, 2.36] & 0.10 & 0.86 \\
GPT-4.1-mini & 33 & 6.6 [3.2, 11.7] & 0.02 & 0.007 & 1.31 [0.64, 2.33] & 0.16 & 0.86 \\
GPT-4.1 & 40 & 7.9 [4.6, 12.1] & 0.00 & 0.000 & 1.57 [0.93, 2.43] & 0.06 & 0.83 \\
GPT-5-mini & 56 & 5.6 [2.9, 10.8] & 0.03 & 0.02 & 1.12 [0.57, 2.16] & 0.17 & 0.86 \\
GPT-5.2 & 31 & 10.9 [6.0, 15.0] & 0.04 & 0.05 & 2.17 [1.19, 3.00] & 0.02 & 0.22 \\
GPT-5.4-mini & 51 & 9.1 [6.4, 16.1] & 0.10 & 0.06 & 1.82 [1.27, 3.22] & 0.01 & 0.17 \\
Haiku 4.5 & 60 & 14.2 [8.8, 19.5] & 0.18 & 0.19 & 2.83 [1.77, 3.90] & $<$0.01 & $<$0.01 \\
Grok-4.1 & 16 & 8.3 [2.8, 15.0] & 0.39 & 0.28 & 1.66 [0.56, 3.00] & 0.09 & 0.86 \\
Llama-3.3 & 38 & 6.8 [3.3, 12.9] & 0.22 & 0.22 & 1.37 [0.66, 2.58] & 0.10 & 0.86 \\
Kimi-K2.5 & 42 & 8.3 [4.3, 12.1] & 0.15 & 0.24 & 1.67 [0.86, 2.43] & 0.07 & 0.83 \\
Mistral-L3 & 24 & 7.5 [3.8, 11.9] & 0.05 & 0.08 & 1.50 [0.75, 2.38] & 0.09 & 0.86 \\
Phi-4 & 21 & 5.7 [2.1, 10.7] & 0.27 & 0.23 & 1.15 [0.43, 2.14] & 0.22 & 0.86 \\
Gemini-2.5-FL & 38 & 5.4 [2.1, 10.1] & 0.43 & 0.32 & 1.07 [0.42, 2.03] & 0.23 & 0.86 \\
Gemini-2.5-F & 4 & --- & --- & --- & --- & --- & --- \\
Gemini-2.5-Pro & 52 & 5.2 [2.0, 9.2] & 0.41 & 0.41 & 1.03 [0.40, 1.85] & 0.25 & 0.86 \\
\bottomrule\end{tabular}
\end{fitwidth}
\caption{PCL-5 responses with totals in the feigners' 10th--90th percentile (6--51), against genuinely injured respondents matched to these totals, with 95\% bootstrap intervals. N.a.a.\ diff.: matched genuine minus group, points of ``not at all''. Ratio: as in Table~\ref{tab:pcl5}. OR: odds ratio as in Table~\ref{tab:pcl5}, with the responses reweighted to the feigners' distribution of totals (bins of five points); every model's is below the feigners' after Holm adjustment. Breadth diff.: group minus matched genuine. $p$: one-sided bootstrap test that the model's ``not at all'' difference exceeds the feigners'; $p_{\text{Holm}}$: Holm-adjusted over the models shown; ---, fewer than 10 responses in the range.}
\label{tab:pcl5-range}\end{table}

\begin{table*}[t]\centering\footnotesize\setlength{\tabcolsep}{3pt}
\begin{fitwidth}
\begin{tabular}{@{}lrrrrrrrrrrrrrr@{}}\toprule
& \multicolumn{9}{c}{Whole forms (17 models)} & \multicolumn{5}{c}{Log-probabilities (10 models)} \\
\cmidrule(lr){2-10}\cmidrule(l){11-15}
Model & $H$ & Real$_{=}$ & $P_0$ & Real$_{=}$ & Lock & Real$_{=}$ & Dist. & Real$_{=}$ & First $r$ & $H$ & $P_0$ & Lock$_L$ & Lock$_I$ & Dist. \\\midrule
Real (bands) & 1.47 & & 0.285 & & & & 72--100 & & --- & 1.51 & 0.285 & 44 & --- & 72--100 \\
\midrule
GPT-4o-mini & 0.59 & 1.51 & 0.007 & 0.183 & 44 & 48 & 19.6 & 91.8 & $0.27$ & 0.19 & 0.000 & 74 & 81 & 3.2 \\
GPT-4o & 0.73 & 1.37 & 0.024 & 0.181 & 67 & 52 & 32.1 & 81.5 & $0.44$ & 0.11 & 0.000 & 67 & 67 & 1.3 \\
GPT-4.1-nano & 0.70 & 1.30 & 0.000 & 0.149 & 85 & 52 & 28.8 & 76.2 & $0.57$ & 0.35 & 0.000 & 70 & 70 & 8.2 \\
GPT-4.1-mini & 0.54 & 1.25 & 0.044 & 0.146 & 59 & 52 & 15.3 & 73.8 & $0.45$ & 0.30 & 0.000 & 74 & 78 & 5.7 \\
GPT-4.1 & 0.37 & 1.31 & 0.002 & 0.189 & 74 & 52 & 7.2 & 78.4 & $0.51$ & 0.09 & 0.000 & 78 & 70 & 1.2 \\
GPT-5-mini & 0.45 & 1.33 & 0.091 & 0.213 & 70 & 56 & 11.2 & 78.0 & $0.31$ & --- & --- & --- & --- & --- \\
GPT-5.2 & 0.29 & 1.32 & 0.000 & 0.165 & 56 & 48 & 4.4 & 78.2 & $0.36$ & 0.10 & 0.000 & 67 & 56 & 0.9 \\
GPT-5.4-mini & 0.42 & 1.28 & 0.067 & 0.169 & 63 & 52 & 9.6 & 73.7 & $0.36$ & 0.27 & 0.000 & 67 & 59 & 4.5 \\
Haiku 4.5 & 0.41 & 1.30 & 0.004 & 0.136 & 52 & 48 & 9.0 & 77.7 & $0.62$ & --- & --- & --- & --- & --- \\
Grok-4.1 & 0.51 & 1.37 & 0.107 & 0.189 & 52 & 52 & 15.2 & 80.4 & $0.53$ & 0.21 & 0.069 & 85 & 81 & 3.3 \\
Llama-3.3 & 0.43 & 1.30 & 0.096 & 0.199 & 63 & 52 & 11.8 & 75.8 & $0.58$ & 0.03 & 0.000 & 63 & 70 & 0.5 \\
Kimi-K2.5 & 0.49 & 1.44 & 0.058 & 0.203 & 59 & 56 & 13.4 & 82.8 & $0.54$ & 0.44 & 0.016 & 85 & 59 & 12.9 \\
Mistral-L3 & 0.46 & 1.28 & 0.000 & 0.144 & 59 & 52 & 10.6 & 74.4 & $0.58$ & --- & --- & --- & --- & --- \\
Phi-4 & 0.56 & 1.15 & 0.024 & 0.101 & 52 & 33 & 17.9 & 69.3 & $0.41$ & --- & --- & --- & --- & --- \\
Gemini-2.5-FL & 0.67 & 1.38 & 0.000 & 0.134 & 44 & 48 & 24.7 & 83.2 & $0.47$ & --- & --- & --- & --- & --- \\
Gemini-2.5-F & 0.53 & 1.24 & 0.002 & 0.139 & 63 & 52 & 15.3 & 74.1 & $0.35$ & --- & --- & --- & --- & --- \\
Gemini-2.5-Pro & 0.53 & 1.32 & 0.040 & 0.178 & 63 & 52 & 16.7 & 77.8 & $0.34$ & --- & --- & --- & --- & --- \\
\bottomrule\end{tabular}
\end{fitwidth}
\caption{Per-item answer distributions, estimated from the answers across the 50 whole-form responses of each severity cell (all models) and from log-probabilities. Real$_{=}$: real respondents with the same totals as the model's forms in each cell (one per form, 200 draws). Real (bands): 50 real respondents per score band (whole forms) or the whole band (log-probabilities). $H$: entropy in bits, mean over items and the four symptomatic severities. $P_0$: share or probability of ``not at all'' at moderate severity. Dist.: \% distinct among 400 forms sampled item by item, mean over severities; for real respondents, sampled the same way from their answers in each band (range over bands). First $r$: correlation of the first answer with the rest of the form, after removing severity and item means. Lock: \% of items whose most common (whole forms) or most likely (log-probabilities) answer equals the severity level (mild 1, moderate 2, severe 3; moderately severe has no single level and is left out), label (L) or impairment (I) framing.}
\label{tab:mechanism}\end{table*}

\begin{table}[t]\centering\footnotesize\setlength{\tabcolsep}{4pt}
\begin{fitwidth}
\begin{tabular}{@{}lrrrr@{}}\toprule
& \multicolumn{2}{c}{Main runs} & \multicolumn{2}{c}{API default} \\
\cmidrule(lr){2-3}\cmidrule(l){4-5}
Model & Generated & Reasoning & Generated & Reasoning \\\midrule
GPT-4o-mini & 46 & 0 & --- & --- \\
GPT-4o & 46 & 0 & --- & --- \\
GPT-4.1-nano & 48 & 0 & --- & --- \\
GPT-4.1-mini & 46 & 0 & --- & --- \\
GPT-4.1 & 46 & 0 & --- & --- \\
GPT-5-mini & 59 & 0 & 525 & 466 \\
GPT-5.2 & 53 & 0 & --- & --- \\
GPT-5.4-mini & 53 & 0 & --- & --- \\
Haiku 4.5 & 66 & --- & --- & --- \\
Grok-4.1 & 46 & 0 & --- & --- \\
Llama-3.3 & 64 & 0 & --- & --- \\
Kimi-K2.5 & 64 & --- & 1,760 & --- \\
Mistral-L3 & 74 & 0 & --- & --- \\
Phi-4 & 71 & 0 & --- & --- \\
Gemini-2.5-FL & 78 & 0 & --- & --- \\
Gemini-2.5-F & 63 & 0 & 810 & 749 \\
Gemini-2.5-Pro & 121 & 68 & 1,368 & 1,304 \\
\bottomrule\end{tabular}
\end{fitwidth}
\caption{Tokens per PHQ-9 form, mean. Main runs: reasoning off, or at the minimum thinking budget (128 tokens) for Gemini 2.5 Pro. API default: for the four models whose default reasons. Generated: all tokens produced, including reasoning. Reasoning: as itemised by the API (---, not itemised or not run).}
\label{tab:tokens}\end{table}

\begin{table*}[t]\centering\footnotesize\setlength{\tabcolsep}{3pt}
\begin{fitwidth}
\begin{tabular}{@{}llrrrrrrrrrrrrrr@{}}\toprule
& & & \multicolumn{5}{c}{PHQ-9} & \multicolumn{3}{c}{GAD-7} & \multicolumn{3}{c}{PCL-5 roles} & \multicolumn{2}{c}{Screening} \\
\cmidrule(lr){4-8}\cmidrule(lr){9-11}\cmidrule(lr){12-14}\cmidrule(l){15-16}
Model & Effort & Tokens & Calib. & First $r$ & All 9 & Div. & AUC & All 7 & Div. & AUC & Ratio & AUC & $\Delta$total & AUC & FPR \\\midrule
GPT-5-mini & minimal$^*$ & 0 & 52 & 0.31 & 87 & 1.31 & 0.77 & 100 & 1.14 & 0.76 & 0.03 & 0.66 & 28.6 & 0.79 & 5.0 \\
 & low & 190 & 98 & $-0.08$ & 88 & 2.15 & 0.83 & 99 & 2.20 & 0.67 & 0.01 & 0.74 & $-37.1$ & 0.75 & 10.9 \\
 & medium$^{\dagger}$ & 466 & 98 & 0.03 & 93 & 1.49 & 0.84 & 99 & 2.28 & 0.62 & 0.00 & 0.65 & $-46.4$ & 0.80 & 8.0 \\
 & high & 1,149 & 99 & 0.01 & 91 & 1.77 & 0.85 & 100 & 2.12 & 0.66 & 0.08 & 0.63 & $-55.2$ & 0.77 & 7.7 \\
\midrule
Gemini-2.5-F & off$^*$ & 0 & 42 & 0.35 & 93 & 1.75 & 0.79 & 100 & 1.87 & 0.63 & 0.04 & 0.77 & 14.5 & 0.75 & 10.8 \\
 & low & 735 & 38 & 0.46 & 97 & 1.56 & 0.80 & 100 & 0.90 & 0.62 & 0.10 & 0.69 & 16.1 & 0.78 & 10.6 \\
 & dynamic$^{\dagger}$ & 749 & 34 & 0.47 & 94 & 1.14 & 0.84 & 100 & 1.18 & 0.64 & 0.08 & 0.74 & 17.0 & 0.78 & 9.3 \\
 & high & 1,139 & 69 & 0.60 & 94 & 1.26 & 0.86 & 100 & 0.67 & 0.64 & 0.07 & 0.64 & 14.5 & 0.74 & 11.9 \\
\midrule
Kimi-K2.5 & off$^*$ & 64$^\ddagger$ & 57 & 0.54 & 89 & 1.44 & 0.74 & 100 & 1.45 & 0.71 & 0.13 & 0.77 & 12.0 & 0.80 & 8.0 \\
 & on$^{\dagger}$ & 1,760$^\ddagger$ & 95 & 0.15 & 90 & 1.77 & 0.77 & 100 & 1.51 & 0.70 & 0.20 & 0.77 & 8.9 & 0.79 & 5.5 \\
\midrule
Gemini-2.5-Pro & min$^*$ & 68 & 49 & 0.34 & 88 & 1.83 & 0.68 & 100 & 1.65 & 0.68 & 0.37 & 0.84 & 14.3 & 0.75 & 8.0 \\
 & dynamic$^{\dagger}$ & 1,304 & 87 & $-0.01$ & 99 & 1.35 & 0.75 & 100 & 1.68 & 0.59 & 0.15 & 0.73 & 12.4 & 0.75 & 13.3 \\
\bottomrule\end{tabular}
\end{fitwidth}
\caption{Reasoning effort; $^*$main run (reasoning off, or at the minimum for Gemini 2.5 Pro), $^{\dagger}$API default; off: no thinking; min: a thinking budget of 128 tokens. Tokens: mean reasoning tokens per PHQ-9 form ($^\ddagger$generated tokens; Kimi-K2.5 does not itemise reasoning). First $r$: as in Table~\ref{tab:mechanism}. PHQ-9 (totals 15--24) and GAD-7 (12--19) measures as in Tables~\ref{tab:phq9-full} and~\ref{tab:gad7-full}. PCL-5 roles: ``not at all'' ratio and separability of the genuine-patient role against the genuinely injured students, and the change in total from the patient to the feigning role, as in Table~\ref{tab:pcl5}. Screening: model trained on the PHQ-9 responses, as in Table~\ref{tab:downstream}.}
\label{tab:reasoning}\end{table*}

\begin{table*}[t]\centering\footnotesize\setlength{\tabcolsep}{3pt}
\begin{fitwidth}
\begin{tabular}{@{}llrrrrrrrrr@{}}\toprule
Model & Effort & PHQ-9 & COACH & GAD-7 & K6 & PHQ-15 & OCI-R & ITQ & PCL-5 pp. & EDE-Q \\\midrule
GPT-5-mini & minimal$^*$ & 0.78 / 6.9 & 0.72 / 6.8 & 0.71 / 6.5 & 0.68 / 6.1 & 0.89 / 4.3 & 0.82 / 4.0 & 0.70 / 3.6 & 0.92 / --- & 0.94 / 5.5 \\
 & low & 0.79 / 2.9 & 0.75 / 2.7 & 0.66 / 3.2 & 0.69 / 4.9 & 0.89 / 4.7 & 0.84 / 6.3 & 0.69 / 3.2 & 0.93 / --- & 0.91 / 5.3 \\
 & medium$^{\dagger}$ & 0.80 / 3.6 & 0.77 / 3.3 & 0.62 / 2.8 & 0.67 / 4.7 & 0.90 / 4.2 & 0.86 / 6.0 & 0.67 / 3.5 & 0.90 / --- & 0.92 / 5.4 \\
 & high & 0.81 / 3.6 & 0.77 / 3.2 & 0.62 / 3.1 & 0.67 / 4.0 & 0.90 / 2.1 & 0.89 / 3.3 & 0.63 / 3.9 & 0.88 / --- & 0.91 / 5.5 \\
\midrule
Gemini-2.5-F & off$^*$ & 0.79 / 5.4 & 0.80 / 4.8 & 0.67 / 4.1 & 0.64 / 3.7 & 0.90 / 3.1 & 0.93 / 2.0 & 0.64 / 2.1 & 0.95 / --- & 0.91 / 4.5 \\
 & low & 0.80 / 4.9 & 0.81 / 4.4 & 0.64 / 8.0 & 0.66 / 6.8 & 0.89 / 4.1 & 0.95 / 3.2 & 0.69 / 3.8 & 0.92 / --- & 0.90 / 3.5 \\
 & dynamic$^{\dagger}$ & 0.81 / 5.6 & 0.80 / 5.3 & 0.65 / 7.2 & 0.66 / 5.9 & 0.90 / 3.4 & 0.95 / 3.9 & 0.69 / 4.0 & 0.93 / --- & 0.88 / 2.9 \\
 & high & 0.83 / 3.8 & 0.82 / 3.4 & 0.65 / 8.2 & 0.66 / 7.1 & 0.90 / 2.9 & 0.96 / 3.4 & 0.70 / 4.3 & 0.92 / --- & 0.90 / 3.3 \\
\midrule
Kimi-K2.5 & off$^*$ & 0.72 / 7.6 & 0.78 / 6.8 & 0.71 / 5.3 & 0.80 / 4.5 & 0.93 / 2.3 & 0.96 / 2.8 & 0.72 / 2.6 & 0.94 / --- & 0.92 / 4.2 \\
 & on$^{\dagger}$ & 0.75 / 4.4 & 0.74 / 4.4 & 0.68 / 4.1 & 0.73 / 3.6 & 0.93 / 1.8 & 0.98 / 3.2 & 0.84 / 3.2 & 0.91 / --- & 0.91 / 5.2 \\
\midrule
Gemini-2.5-Pro & min$^*$ & 0.66 / 4.6 & 0.81 / 4.0 & 0.72 / 4.5 & 0.68 / 4.0 & 0.97 / 2.7 & 0.99 / 2.1 & 0.75 / 2.2 & 0.97 / --- & 0.90 / 4.2 \\
 & dynamic$^{\dagger}$ & 0.76 / 4.6 & 0.82 / 4.2 & 0.60 / 3.8 & 0.70 / 5.2 & 0.94 / 2.0 & 0.98 / 1.8 & 0.70 / 2.5 & 0.96 / --- & 0.90 / 3.6 \\
\bottomrule\end{tabular}
\end{fitwidth}
\caption{Reasoning effort on the nine references, protocol of Table~\ref{tab:generality}; efforts marked as in Table~\ref{tab:reasoning}. Each cell: separability (AUC) / real-to-model diversity ratio, shown where at least five totals carry it. Clinical samples: COACH PHQ-9 and postpartum PCL-5.}
\label{tab:reasoning-refs}\end{table*}

\begin{table*}[t]\centering\footnotesize\setlength{\tabcolsep}{3pt}
\begin{fitwidth}
\begin{tabular}{@{}llrrrrrrrrrr@{}}\toprule
& & \multicolumn{2}{c}{Interventions (AUC)} & \multicolumn{3}{c}{Published personas} & \multicolumn{5}{c}{Interviews (rater GPT-4.1)} \\
\cmidrule(lr){3-4}\cmidrule(lr){5-7}\cmidrule(l){8-12}
Model & Effort & One-line & +5 sheets & TalkDep & CACTUS & SI & All 8 & Div. & $r$ real & $r$ sampled & SI agr. \\\midrule
Gemini-2.5-F & off$^*$ & 0.81 & 0.62 & 0.85 & 0.88 & 16 & 100 / 83 & 0.69 / 5.40 & 0.83 & 0.79 & 100 \\
 & low & 0.80 & 0.75 & 0.85 & 0.87 & 9 & 100 / 74 & 0.83 / 5.41 & 0.89 & 0.89 & 100 \\
 & dynamic$^{\dagger}$ & 0.79 & 0.72 & 0.89 & 0.90 & 7 & 100 / 77 & 0.77 / 5.34 & 0.90 & 0.89 & 100 \\
 & high & 0.83 & 0.74 & 0.88 & 0.91 & 6 & 99 / 77 & 0.48 / 5.94 & 0.89 & 0.88 & 100 \\
\midrule
Gemini-2.5-Pro & min$^*$ & 0.64 & 0.69 & 0.86 & 0.88 & 2 & 100 / 83 & 0.71 / 5.50 & 0.83 & 0.83 & 100 \\
 & dynamic$^{\dagger}$ & 0.64 & 0.71 & 0.80 & 0.88 & 5 & 100 / 83 & 0.27 / 4.35 & 0.85 & 0.82 & 100 \\
\bottomrule\end{tabular}
\end{fitwidth}
\caption{Reasoning effort in the Gemini 2.5 Flash and Pro experiments; efforts marked as in Table~\ref{tab:reasoning}. Interventions and published personas as in Tables~\ref{tab:fixes} and~\ref{tab:personas} (AUC at totals 15--24 against held-out real respondents; SI: \% endorsing item 9 with CACTUS personas). Interviews as in Table~\ref{tab:interview}: severity-persona interviews, all eight endorsed and diversity, each against real respondents at the same totals; $r$: correlation of given and rated answers with a real or a sampled profile; SI agr.: \% of real-profile interviews whose rated suicidal ideation matches the profile.}
\label{tab:reasoning-gemini}\end{table*}

\begin{table}[t]\centering\footnotesize\setlength{\tabcolsep}{2.6pt}
\begin{fitwidth}
\begin{tabular}{@{}llrrrrrr@{}}\toprule
Model & Condition & Calib. & Div. & Real$_{=}$ & N.a.a. & AUC & SI$_{\text{ms/s}}$ \\\midrule
GPT-4o-mini & One-line & 46 & 2.23 & 8.09 & 0.0 & 0.88 & 100 \\
 & + ``people differ'' & 42 & 2.49 & 7.96 & 0.1 & 0.86 & 100 \\
 & + biography & 49 & 1.95 & 8.25 & 0.0 & 0.86 & 99 \\
 & Total, no label & 68 & 2.28 & 8.44 & 0.2 & 0.82 & 78 \\
 & + 5 real sheets & 63 & 4.51 & 8.10 & 0.5 & 0.69 & 86 \\
\midrule
GPT-4.1 & One-line & 48 & 0.66 & 6.69 & 0.0 & 0.74 & 100 \\
 & + ``people differ'' & 73 & 1.89 & 7.34 & 0.0 & 0.76 & 98 \\
 & + biography & 52 & 1.17 & 7.04 & 0.0 & 0.71 & 100 \\
 & Total, no label & 100 & 3.01 & 7.72 & 0.2 & 0.69 & 88 \\
 & + 5 real sheets & 83 & 3.16 & 7.85 & 0.0 & 0.57 & 92 \\
\midrule
Haiku 4.5 & One-line & 25 & 1.05 & 7.32 & 0.0 & 0.85 & 100 \\
 & + ``people differ'' & 48 & 2.28 & 8.36 & 0.1 & 0.84 & 99 \\
 & + biography & 41 & 1.34 & 7.59 & 0.0 & 0.81 & 100 \\
 & Total, no label & 60 & 2.43 & 8.13 & 0.0 & 0.79 & 34 \\
 & + 5 real sheets & 88 & 2.27 & 7.41 & 0.0 & 0.64 & 81 \\
\midrule
Llama-3.3 & One-line & 42 & 0.63 & 7.28 & 0.0 & 0.73 & 100 \\
 & + ``people differ'' & 46 & 2.45 & 8.01 & 0.1 & 0.86 & 82 \\
 & + biography & 49 & 1.26 & 7.40 & 0.0 & 0.80 & 86 \\
 & Total, no label & 47 & 2.76 & 7.84 & --- & --- & 6 \\
 & + 5 real sheets & 76 & 3.43 & 7.10 & 0.0 & 0.68 & 75 \\
\midrule
Gemini-2.5-FL & One-line & 33 & 2.03 & 6.95 & 0.1 & 0.87 & 100 \\
 & + ``people differ'' & 45 & 2.28 & 7.96 & 0.1 & 0.87 & 99 \\
 & + biography & 40 & 2.24 & 7.83 & 0.2 & 0.85 & 94 \\
 & Total, no label & 28 & 2.47 & 7.37 & 0.0 & --- & 21 \\
 & + 5 real sheets & 82 & 4.08 & 7.95 & 0.8 & 0.68 & 70 \\
\midrule
Gemini-2.5-F & One-line & 29 & 0.98 & 6.07 & 0.0 & 0.81 & 100 \\
 & + ``people differ'' & 51 & 1.95 & 7.57 & 0.5 & 0.79 & 98 \\
 & + biography & 30 & 1.87 & 7.48 & 0.1 & 0.76 & 100 \\
 & Total, no label & 99 & 4.00 & 7.79 & 1.3 & 0.65 & 70 \\
 & + 5 real sheets & 84 & 3.64 & 7.82 & 1.3 & 0.62 & 76 \\
\midrule
Gemini-2.5-Pro & One-line & 39 & 1.75 & 7.12 & 0.0 & 0.64 & 100 \\
 & + ``people differ'' & 64 & 3.57 & 8.18 & 6.5 & 0.77 & 89 \\
 & + biography & 47 & 1.76 & 7.73 & 0.1 & 0.72 & 98 \\
 & Total, no label & 100 & 3.11 & 7.72 & 0.4 & 0.59 & 70 \\
 & + 5 real sheets & 88 & 4.52 & 7.98 & 5.5 & 0.69 & 82 \\
\bottomrule\end{tabular}
\end{fitwidth}
\caption{Interventions (PHQ-9). Calib.: \% in the instructed band. Div.: diversity, totals 5--27; Real$_{=}$: held-out real respondents at the same totals. N.a.a.: \% ``not at all'' on items 1--8, totals 15--24 (real: 9.9). AUC: separability from held-out real respondents (chance 0.44--0.57; ---, too few responses at totals 15--24). SI$_{\text{ms/s}}$: \% endorsing item 9 among moderately severe and severe personas, all totals.}
\label{tab:fixes}\end{table}

\begin{table}[t]\centering\footnotesize\setlength{\tabcolsep}{2.6pt}
\begin{fitwidth}
\begin{tabular}{@{}llrrrrrrrr@{}}\toprule
Model & Persona & $n$ & Div. & Real$_{=}$ & N.a.a.\ 1--8 & SI & All 9 & AUC & AUC$_{1\text{--}8}$ \\\midrule
GPT-4o-mini & One-line & 160 & 2.23 & 8.09 & 0.0 & 99.1 & 99.1 & 0.88 & 0.85 \\
 & TalkDep & 360 & 3.95 & 7.94 & 3.6 & 43.1 & 29.2 & 0.89 & 0.88 \\
 & CACTUS & 400 & 3.07 & 8.79 & 1.1 & 1.0 & 1.0 & 0.92 & 0.88 \\
 & CACTUS + form. & 400 & 2.97 & 8.73 & 1.3 & 3.5 & 2.5 & 0.92 & 0.91 \\
\midrule
GPT-4.1 & One-line & 160 & 0.66 & 6.69 & 0.0 & 100.0 & 100.0 & 0.74 & 0.71 \\
 & TalkDep & 360 & 2.39 & 7.39 & 0.0 & 72.0 & 72.0 & 0.75 & 0.74 \\
 & CACTUS & 400 & 2.74 & 8.22 & 0.0 & 17.2 & 17.2 & 0.84 & 0.82 \\
 & CACTUS + form. & 400 & 2.66 & 8.27 & 0.0 & 19.9 & 19.9 & 0.86 & 0.84 \\
\midrule
Haiku 4.5 & One-line & 160 & 1.05 & 7.32 & 0.0 & 94.3 & 94.3 & 0.85 & 0.83 \\
 & TalkDep & 360 & 2.63 & 7.27 & 0.0 & 51.8 & 51.8 & 0.88 & 0.89 \\
 & CACTUS & 400 & 3.21 & 7.96 & 0.0 & 8.0 & 8.0 & 0.86 & 0.81 \\
 & CACTUS + form. & 400 & 3.18 & 8.22 & 0.0 & 8.0 & 8.0 & 0.89 & 0.83 \\
\midrule
Llama-3.3 & One-line & 160 & 0.63 & 7.28 & 0.0 & 83.6 & 83.6 & 0.73 & 0.80 \\
 & TalkDep & 360 & 2.31 & 7.27 & 0.0 & 32.0 & 32.0 & 0.86 & 0.84 \\
 & CACTUS & 400 & 3.08 & 8.19 & 0.2 & 0.0 & 0.0 & 0.90 & 0.82 \\
 & CACTUS + form. & 400 & 3.00 & 8.19 & 0.3 & 0.0 & 0.0 & 0.92 & 0.86 \\
\midrule
Gemini-2.5-FL & One-line & 160 & 2.03 & 6.95 & 0.1 & 88.2 & 87.4 & 0.87 & 0.85 \\
 & TalkDep & 360 & 3.49 & 7.64 & 0.8 & 48.3 & 45.5 & 0.88 & 0.91 \\
 & CACTUS & 400 & 3.36 & 8.37 & 1.5 & 2.8 & 1.9 & 0.92 & 0.88 \\
 & CACTUS + form. & 400 & 3.35 & 8.31 & 2.0 & 0.7 & 0.7 & 0.93 & 0.87 \\
\midrule
Gemini-2.5-F & One-line & 160 & 0.98 & 6.07 & 0.0 & 93.1 & 93.1 & 0.81 & 0.77 \\
 & TalkDep & 360 & 3.41 & 7.26 & 0.3 & 44.5 & 44.5 & 0.85 & 0.86 \\
 & CACTUS & 400 & 3.43 & 8.19 & 2.7 & 15.8 & 12.9 & 0.88 & 0.83 \\
 & CACTUS + form. & 400 & 3.25 & 8.18 & 2.1 & 18.3 & 17.3 & 0.91 & 0.87 \\
\midrule
Gemini-2.5-Pro & One-line & 160 & 1.75 & 7.12 & 0.0 & 90.5 & 90.5 & 0.64 & 0.62 \\
 & TalkDep & 360 & 3.02 & 7.29 & 0.0 & 49.3 & 49.3 & 0.86 & 0.87 \\
 & CACTUS & 400 & 3.18 & 8.19 & 1.7 & 1.8 & 0.9 & 0.88 & 0.81 \\
 & CACTUS + form. & 400 & 3.04 & 8.32 & 1.2 & 0.7 & 0.7 & 0.90 & 0.84 \\
\midrule
\textbf{Real} (held-out) & & 707 & --- & --- & 9.9 & 42.3 & 18.5 & --- & --- \\
\bottomrule\end{tabular}
\end{fitwidth}
\caption{Published simulator personas (PHQ-9), intervention protocol. One-line: the severity persona, from the intervention runs. CACTUS + form.: the CACTUS intake form with the automatic thought and cognitive distortions of the client's case formulation. $n$: responses; for Real, held-out respondents at totals 15--24. Div.: diversity, totals 5--27; Real$_{=}$: held-out real respondents at the same totals. At totals 15--24: N.a.a.\ 1--8, \% ``not at all'' on items 1--8; SI, \% endorsing item 9; All 9, \% endorsing every item; AUC, separability from held-out real respondents (chance 0.44--0.55); AUC$_{1\text{--}8}$, the same without item 9. The reference is the held-out half of NHANES, so one-line values differ slightly from Table~\ref{tab:phq9-full}. TalkDep personas are each answered repeatedly, so their diversity uses pairs of different personas and a persona's responses stay in one cross-validation fold.}
\label{tab:personas}\end{table}

\begin{table*}[t]\centering\footnotesize\setlength{\tabcolsep}{4pt}
\begin{fitwidth}
\begin{tabular}{@{}lrrrrrrrrrrr@{}}\toprule
& & & & & \multicolumn{2}{c}{Breadth} & \multicolumn{2}{c}{Several days} & Key & \multicolumn{2}{c}{Other} \\
\cmidrule(lr){6-7}\cmidrule(lr){8-9}\cmidrule(lr){11-12}
Persona & BDI-II & PHQ-9 & s.d. & Distinct & Model & Real & Model & Real & Model & Model & Real \\\midrule
Noah & 5 & 4.7 & 0.7 & 23 & 4.7 & 3.6 & 100 & 78 & 91 & 33 & 36 \\
Maya & 6 & 3.3 & 0.7 & 20 & 3.3 & 2.6 & 100 & 83 & 97 & 7 & 27 \\
Priya & 7 & 4.7 & 0.8 & 23 & 4.7 & 3.6 & 100 & 78 & 100 & 30 & 36 \\
Ethan & 12 & 8.3 & 0.9 & 27 & 6.3 & 5.4 & 67 & 61 & 100 & 54 & 55 \\
Gabriel & 13 & 10.7 & 1.1 & 37 & 7.8 & 6.2 & 62 & 50 & 100 & 82 & 68 \\
Alex & 15 & 9.4 & 0.9 & 27 & 7.6 & 5.7 & 77 & 56 & 100 & 77 & 60 \\
James & 22 & 14.6 & 0.7 & 20 & 8.0 & 7.0 & 29 & 30 & 100 & 83 & 74 \\
Laura & 23 & 16.4 & 1.1 & 37 & 8.1 & 7.4 & 26 & 25 & 100 & 82 & 76 \\
Linda & 28 & 16.4 & 1.2 & 40 & 8.0 & 7.5 & 24 & 25 & 100 & 81 & 76 \\
Elena & 35 & 22.4 & 1.3 & 43 & 9.0 & 8.4 & 8 & 8 & 100 & 100 & 91 \\
Marco & 38 & 23.5 & 0.9 & 30 & 9.0 & 8.5 & 8 & 6 & 100 & 100 & 93 \\
Maria & 40 & 16.8 & 1.4 & 47 & 8.5 & 7.5 & 30 & 23 & 100 & 91 & 74 \\
\bottomrule\end{tabular}
\end{fitwidth}
\caption{TalkDep personas, median over 17 models. BDI-II: published score. PHQ-9: mean total over 30 runs, with its s.d. Distinct: \% distinct answer patterns. Breadth: symptoms endorsed. Several days: \% of endorsed answers at ``several days''. Key, Other: \% endorsing the items matching the persona's key symptoms, and the rest. Real: real respondents at the same totals.}
\label{tab:talkdep}\end{table*}

\begin{table*}[t]\centering\footnotesize\setlength{\tabcolsep}{3pt}
\begin{fitwidth}
\begin{tabular}{@{}lrrrrrrrr@{}}\toprule
& \multicolumn{4}{c}{PHQ-9} & \multicolumn{4}{c}{GAD-7} \\
\cmidrule(lr){2-5}\cmidrule(l){6-9}
Model & One-line & TalkDep & CACTUS & + form. & One-line & TalkDep & CACTUS & + form. \\\midrule
GPT-4o-mini & 0.83 / 4.3 & 0.81 / 2.0 & 0.93 / 2.9 & 0.94 / 3.0 & 0.70 / 4.8 & 0.63 / 2.3 & 0.66 / 3.1 & 0.70 / 3.1 \\
GPT-4o & 0.77 / 3.4 & 0.72 / 2.3 & 0.89 / 2.4 & 0.89 / 2.6 & 0.76 / 3.8 & 0.54 / 2.1 & 0.72 / 2.8 & 0.74 / 2.9 \\
GPT-4.1-nano & 0.83 / 3.2 & 0.76 / 1.6 & 0.88 / 2.3 & 0.90 / 2.3 & 0.72 / 4.6 & 0.58 / 1.9 & 0.62 / 2.5 & 0.58 / 2.2 \\
GPT-4.1-mini & 0.77 / 4.3 & 0.73 / 2.1 & 0.88 / 2.7 & 0.90 / 3.2 & 0.72 / 4.4 & 0.62 / 2.3 & 0.61 / 3.4 & 0.64 / 3.2 \\
GPT-4.1 & 0.67 / 7.7 & 0.71 / 3.1 & 0.89 / 3.0 & 0.88 / 3.1 & 0.71 / 5.0 & 0.49 / 2.6 & 0.64 / 3.4 & 0.68 / 4.0 \\
GPT-5-mini & 0.75 / 6.0 & 0.76 / 2.5 & 0.89 / 2.9 & 0.90 / 3.3 & 0.69 / 5.7 & 0.55 / 3.2 & 0.63 / 4.2 & 0.67 / 4.1 \\
GPT-5.2 & 0.75 / 9.7 & 0.71 / 3.0 & 0.89 / 2.8 & 0.89 / 2.8 & 0.71 / 9.1 & 0.57 / 3.6 & 0.66 / 4.2 & 0.67 / 4.5 \\
GPT-5.4-mini & 0.75 / 6.3 & 0.71 / 2.7 & 0.92 / 2.6 & 0.92 / 2.8 & 0.67 / 5.9 & 0.52 / 3.7 & 0.69 / 4.8 & 0.71 / 4.0 \\
Haiku 4.5 & 0.80 / 7.3 & 0.73 / 2.8 & 0.91 / 2.5 & 0.92 / 2.6 & 0.70 / 4.8 & 0.64 / 3.4 & 0.58 / 3.3 & 0.60 / 3.3 \\
Grok-4.1 & 0.76 / 4.9 & 0.62 / 2.0 & 0.91 / 2.1 & 0.92 / 2.3 & 0.69 / 5.2 & 0.70 / 2.4 & 0.75 / 2.6 & 0.76 / 2.3 \\
Llama-3.3 & 0.80 / 6.2 & 0.78 / 3.2 & 0.92 / 2.7 & 0.93 / 2.7 & 0.66 / 6.3 & 0.66 / 3.5 & 0.67 / 3.3 & 0.64 / 3.3 \\
Kimi-K2.5 & 0.72 / 5.8 & 0.76 / 2.7 & 0.89 / 2.5 & 0.90 / 2.5 & 0.70 / 4.6 & 0.56 / 3.0 & 0.70 / 3.0 & 0.72 / 3.1 \\
Mistral-L3 & 0.72 / 5.7 & 0.72 / 3.1 & 0.89 / 2.5 & 0.89 / 2.6 & 0.66 / 5.7 & 0.71 / 3.5 & 0.64 / 3.7 & 0.65 / 3.8 \\
Phi-4 & 0.81 / 4.2 & 0.75 / 2.3 & 0.93 / 2.3 & 0.94 / 2.5 & 0.65 / 4.9 & 0.72 / 2.5 & 0.73 / 2.8 & 0.68 / 2.8 \\
Gemini-2.5-FL & 0.84 / 3.7 & 0.81 / 2.2 & 0.93 / 2.5 & 0.95 / 2.5 & 0.70 / 4.0 & 0.69 / 2.9 & 0.65 / 3.4 & 0.68 / 3.2 \\
Gemini-2.5-F & 0.79 / 5.0 & 0.74 / 2.1 & 0.92 / 2.4 & 0.92 / 2.5 & 0.66 / 3.5 & 0.65 / 2.8 & 0.65 / 2.4 & 0.66 / 2.3 \\
Gemini-2.5-Pro & 0.67 / 4.7 & 0.78 / 2.4 & 0.93 / 2.6 & 0.94 / 2.8 & 0.72 / 4.4 & 0.66 / 2.5 & 0.68 / 3.1 & 0.67 / 3.1 \\
\bottomrule\end{tabular}
\end{fitwidth}
\caption{Published simulator personas on every model and on both questionnaires. Each cell: separability (AUC) from real respondents with the same totals (NHANES for the PHQ-9, NHIS for the GAD-7; totals $\geq$5) / real-to-model diversity ratio. One-line: the severity persona of the main runs. + form.: CACTUS with the automatic thought and cognitive distortions of the client's case formulation. All totals of at least 5 are used, so one-line values differ slightly from Tables~\ref{tab:phq9-full}, \ref{tab:gad7-full} and~\ref{tab:personas} (totals 15--24). TalkDep personas are each answered repeatedly, so their diversity uses pairs of different personas and a persona's responses stay in one cross-validation fold.}
\label{tab:personas-all}\end{table*}

\begin{table}[t]\centering\footnotesize\setlength{\tabcolsep}{3pt}
\begin{fitwidth}
\begin{tabular}{@{}llrr@{}}\toprule
Family & Model & Diversity [2.5th, 97.5th] & Calib.\ (\%) \\\midrule
GPT-4.1 & GPT-4.1-nano & 2.32 [2.23, 2.44] & 46.8 \\
GPT-4.1 & GPT-4.1-mini & 1.86 [1.76, 1.94] & 49.1 \\
GPT-4.1 & GPT-4.1 & 0.88 [0.77, 0.94] & 65.9 \\
GPT-4o & GPT-4o-mini & 1.95 [1.87, 2.06] & 41.8 \\
GPT-4o & GPT-4o & 2.43 [2.34, 2.51] & 58.6 \\
GPT-5 & GPT-5-mini & 1.31 [1.11, 1.40] & 52.3 \\
GPT-5 & GPT-5.2 & 0.83 [0.76, 0.91] & 50.7 \\
Gemini-2.5 & Gemini-2.5-FL & 2.06 [1.97, 2.15] & 48.2 \\
Gemini-2.5 & Gemini-2.5-F & 1.75 [1.56, 1.82] & 41.6 \\
Gemini-2.5 & Gemini-2.5-Pro & 1.83 [1.76, 2.05] & 48.9 \\
\midrule Real & & 7.35 & --- \\
\bottomrule\end{tabular}
\end{fitwidth}
\caption{Diversity by model tier within families (PHQ-9, totals 15--24), with the 2.5th--97.5th percentiles over 80\% subsamples drawn without replacement.}
\label{tab:scale}\end{table}

\begin{table*}[t]\centering\footnotesize\setlength{\tabcolsep}{3.5pt}
\begin{fitwidth}
\begin{tabular}{@{}llrrrrrrrrrrr@{}}\toprule
Patient & Rater & $n$ & Breadth (/8) & All 8 (\%) & Real$_{=}$ & Not at all (\%) & Div. & Real$_{=}$ & SI (\%) & SI agr. & $r_{\text{item}}$ & $r_{\text{total}}$ \\\midrule
NHANES, total $\geq$5 & --- & 10760 & 5.25 & 10.7 & --- & 34.3 & 6.21 & --- & 14.3 & --- & --- & --- \\
severity, Haiku 4.5 & GPT-4.1 & 80 & 7.99 & 98.8 & 81.8 & 0.2 & 1.20 & 5.79 & 82.5 & --- & --- & --- \\
severity, Gemini-2.5-F & GPT-4.1 & 80 & 8.00 & 100.0 & 82.8 & 0.0 & 0.69 & 5.40 & 97.5 & --- & --- & --- \\
severity, Gemini-2.5-F & Haiku 4.5 & 80 & 8.00 & 100.0 & 67.4 & 0.0 & 0.60 & 6.11 & 97.5 & --- & --- & --- \\
severity, Gemini-2.5-FL & GPT-4.1 & 80 & 8.00 & 100.0 & 72.2 & 0.0 & 0.79 & 5.65 & 75.0 & --- & --- & --- \\
severity, Gemini-2.5-FL & Haiku 4.5 & 80 & 8.00 & 100.0 & 58.0 & 0.0 & 1.01 & 6.25 & 75.0 & --- & --- & --- \\
severity, Gemini-2.5-Pro & GPT-4.1 & 80 & 8.00 & 100.0 & 82.9 & 0.0 & 0.71 & 5.50 & 77.5 & --- & --- & --- \\
severity, Gemini-2.5-Pro & Haiku 4.5 & 80 & 8.00 & 100.0 & 62.1 & 0.0 & 0.36 & 5.33 & 77.5 & --- & --- & --- \\
severity, GPT-4.1 & Haiku 4.5 & 80 & 7.97 & 97.5 & 53.4 & 0.3 & 0.57 & 5.47 & 88.8 & --- & --- & --- \\
severity, GPT-4o-mini & GPT-4.1 & 80 & 8.00 & 100.0 & 67.2 & 0.0 & 0.97 & 5.90 & 97.5 & --- & --- & --- \\
severity, GPT-4o-mini & Haiku 4.5 & 80 & 8.00 & 100.0 & 48.8 & 0.0 & 0.85 & 6.61 & 96.2 & --- & --- & --- \\
severity, Llama-3.3 & GPT-4.1 & 71 & 8.00 & 100.0 & 84.8 & 0.0 & 0.48 & 4.65 & --- & --- & --- & --- \\
severity, Llama-3.3 & Haiku 4.5 & 71 & 8.00 & 100.0 & 56.9 & 0.0 & 0.82 & 6.39 & --- & --- & --- & --- \\
real profile, Gemini-2.5-F & GPT-4.1 & 80 & 6.67 & 35.0 & 38.8 & 16.6 & 4.91 & 6.83 & 31.2 & 100 & 0.83 & 0.90 \\
real profile, Gemini-2.5-F & Haiku 4.5 & 80 & 6.70 & 35.0 & 33.7 & 16.2 & 4.81 & 7.09 & 31.2 & 100 & 0.82 & 0.90 \\
real profile, Gemini-2.5-FL & GPT-4.1 & 80 & 6.75 & 37.5 & 49.7 & 15.6 & 4.03 & 6.63 & 31.2 & 100 & 0.75 & 0.83 \\
real profile, Gemini-2.5-FL & Haiku 4.5 & 80 & 6.79 & 40.0 & 39.0 & 15.2 & 4.25 & 7.22 & 31.2 & 100 & 0.73 & 0.81 \\
real profile, Gemini-2.5-Pro & GPT-4.1 & 80 & 6.64 & 35.0 & 44.8 & 17.0 & 4.64 & 6.41 & 31.2 & 100 & 0.83 & 0.89 \\
real profile, Gemini-2.5-Pro & Haiku 4.5 & 80 & 6.61 & 35.0 & 37.7 & 17.3 & 4.53 & 6.81 & 31.2 & 100 & 0.82 & 0.88 \\
real profile, GPT-4o-mini & GPT-4.1 & 80 & 6.81 & 40.0 & 41.7 & 14.8 & 3.59 & 6.55 & 31.2 & 100 & 0.78 & 0.89 \\
real profile, GPT-4o-mini & Haiku 4.5 & 80 & 6.79 & 40.0 & 31.5 & 15.2 & 3.30 & 7.25 & 31.2 & 100 & 0.74 & 0.86 \\
real profile, Llama-3.3 & GPT-4.1 & 74 & 6.77 & 36.5 & 50.2 & 15.4 & 4.57 & 6.77 & --- & --- & 0.72 & 0.83 \\
real profile, Llama-3.3 & Haiku 4.5 & 74 & 6.74 & 35.1 & 37.7 & 15.7 & 4.46 & 7.34 & --- & --- & 0.70 & 0.83 \\
sampled profile, Gemini-2.5-F & GPT-4.1 & 80 & 6.81 & 42.5 & 43.7 & 14.8 & 4.86 & 7.01 & 36.2 & 100 & 0.79 & 0.90 \\
sampled profile, Gemini-2.5-F & Haiku 4.5 & 80 & 6.83 & 42.5 & 37.3 & 14.7 & 4.30 & 7.04 & 36.2 & 100 & 0.79 & 0.90 \\
sampled profile, Gemini-2.5-FL & GPT-4.1 & 80 & 6.89 & 46.2 & 50.7 & 13.9 & 3.47 & 6.95 & 36.2 & 100 & 0.73 & 0.85 \\
sampled profile, Gemini-2.5-FL & Haiku 4.5 & 80 & 6.88 & 45.0 & 38.9 & 14.1 & 3.41 & 6.98 & 36.2 & 100 & 0.71 & 0.81 \\
sampled profile, Gemini-2.5-Pro & GPT-4.1 & 80 & 6.74 & 42.5 & 45.8 & 15.8 & 4.13 & 6.09 & 36.2 & 100 & 0.83 & 0.89 \\
sampled profile, Gemini-2.5-Pro & Haiku 4.5 & 80 & 6.76 & 42.5 & 38.6 & 15.5 & 4.42 & 6.74 & 36.2 & 100 & 0.82 & 0.90 \\
sampled profile, GPT-4o-mini & GPT-4.1 & 80 & 6.86 & 43.8 & 46.1 & 14.2 & 3.93 & 7.16 & 36.2 & 100 & 0.75 & 0.89 \\
sampled profile, GPT-4o-mini & Haiku 4.5 & 80 & 6.88 & 46.2 & 35.7 & 14.1 & 3.26 & 7.12 & 36.2 & 100 & 0.73 & 0.87 \\
sampled profile, Llama-3.3 & GPT-4.1 & 72 & 6.78 & 41.7 & 53.1 & 15.3 & 3.81 & 6.58 & --- & --- & 0.71 & 0.85 \\
sampled profile, Llama-3.3 & Haiku 4.5 & 72 & 6.76 & 41.7 & 40.3 & 15.5 & 3.51 & 6.90 & --- & --- & 0.70 & 0.85 \\
\bottomrule\end{tabular}
\end{fitwidth}
\caption{Interviews, PHQ-9 items 1--8 rated from the transcript. Severity: the patient is told a severity; real profile: given a real NHANES profile; sampled profile: given a GRM-sampled profile. Real$_{=}$: real NHANES respondents at the same item 1--8 totals. SI agr.: \% of interviews whose rated suicidal ideation (present or absent) matches the given profile. $r$: correlation of given and rated profile. SI is not reported for Llama-3.3 (Appendix~\ref{app:filter}). The two raters agree within one point on 95--100\% of items.}
\label{tab:interview}\end{table*}

\begin{table*}[t]\centering\footnotesize\setlength{\tabcolsep}{4pt}
\begin{fitwidth}
\begin{tabular}{@{}llrrrrrrrrr@{}}\toprule
Given & Rater & Items & Exact & Within 1 & $\kappa_w$ & Present$\to$present & Absent$\to$absent & Exact if present & 1 rated higher & Shift \\\midrule
Real profile & GPT-4.1 & 3{,}152 & 56.6 & 92.4 & 0.71 & 100.0 & 84.8 & 50.1 & 80.7 & +0.31 \\
Real profile & Haiku 4.5 & 3{,}152 & 54.2 & 93.5 & 0.71 & 100.0 & 84.9 & 47.1 & 76.6 & +0.20 \\
Sampled profile & GPT-4.1 & 3{,}136 & 53.4 & 90.4 & 0.68 & 100.0 & 83.2 & 46.9 & 84.6 & +0.36 \\
Sampled profile & Haiku 4.5 & 3{,}136 & 52.2 & 92.5 & 0.68 & 100.0 & 82.8 & 45.6 & 80.5 & +0.25 \\
\bottomrule\end{tabular}
\end{fitwidth}
\caption{Given against rated answers, PHQ-9 items 1--8, pooled over the patient models (a model never rates its own transcripts). Items: given answers compared. Exact, Within 1: \% of items rated at the given level, or within one level. $\kappa_w$: quadratically weighted kappa. Present$\to$present: \% of symptoms given as present (above 0) that are rated present; Absent$\to$absent: \% of symptoms given at 0 that are rated 0. Exact if present: \% of present symptoms rated at the given level. 1 rated higher: \% of symptoms given at ``several days'' rated higher. Shift: mean rated minus given level.}
\label{tab:interview-agreement}\end{table*}

\begin{table*}[t]\centering\footnotesize\setlength{\tabcolsep}{4pt}
\begin{fitwidth}
\begin{tabular}{@{}lrrrrrrr@{}}\toprule
& \multicolumn{2}{c}{Real-to-model diversity} & \multicolumn{2}{c}{All 8 endorsed (\%)} & \multicolumn{3}{c}{AUC against real respondents} \\
\cmidrule(lr){2-3}\cmidrule(lr){4-5}\cmidrule(l){6-8}
Patient model & Questionnaire & Interview & Severity & Real profile & Severity & Real profile & Sampled profile \\\midrule
Haiku 4.5 & 7.7 & --- & --- & --- & 0.66 & --- & --- \\
Gemini-2.5-F & 4.2 & 3.7 & 100 & 57 & 0.72 & 0.66 & 0.72 \\
Gemini-2.5-FL & 3.6 & 3.0 & 100 & 46 & 0.73 & 0.70 & 0.79 \\
Gemini-2.5-Pro & 4.0 & 4.6 & 100 & 50 & 0.68 & 0.65 & 0.62 \\
GPT-4.1 & 8.3 & --- & --- & --- & 0.78 & --- & --- \\
GPT-4o-mini & 4.0 & 2.9 & 100 & 59 & 0.73 & 0.69 & 0.72 \\
Llama-3.3 & 5.1 & 3.6 & 100 & 47 & 0.69 & 0.73 & 0.76 \\
\bottomrule\end{tabular}
\end{fitwidth}
\caption{Interviews against real profiles passed through the same interviewer and rater (PHQ-9 items 1--8). Questionnaire: real-to-model diversity ratio at totals 15--24, as in Table~\ref{tab:phq9-full}. Interview: diversity of real-profile interviews over severity-persona interviews, both computed within the same bins of three total points (bins with at least three interviews of each kind). All 8: \% of interviews rated as endorsing all eight items, mean over the same bins. AUC: separability of the rated profiles from real NHANES respondents with the same item 1--8 totals. ---: no real-profile interviews for this model.}
\label{tab:interview-pipeline}\end{table*}

\begin{table*}[t]\centering\footnotesize\setlength{\tabcolsep}{4pt}
\begin{fitwidth}
\begin{tabular}{@{}lrrrrrrrrrrrrr@{}}\toprule
& & & \multicolumn{2}{c}{All 8 (\%)} & \multicolumn{2}{c}{Diversity} & & & \multicolumn{3}{c}{Agent against the label} & \multicolumn{2}{c}{$\rho$ with BDI-II} \\
\cmidrule(lr){4-5}\cmidrule(lr){6-7}\cmidrule(lr){10-12}\cmidrule(l){13-14}
Patient model & $n$ & Breadth (/8) & TalkDep & Real$_{=}$ & TalkDep & Real$_{=}$ & Ratio & AUC & Exact & Within 1 & Above & Interviews & Personas \\\midrule
GPT-4o-mini & 84 & 6.82 & 83 & 58 & 2.14 & 6.72 & 1.9 & 0.67 & 37 & 87 & 62 & 0.90 & 0.93 \\
GPT-4.1 & 84 & 6.33 & 95 & 45 & 2.09 & 6.72 & --- & 0.71 & 62 & 94 & 38 & 0.94 & 0.96 \\
Haiku 4.5 & 84 & 6.63 & 95 & 80 & 2.55 & 5.98 & --- & 0.73 & 40 & 82 & 60 & 0.95 & 0.96 \\
Llama-3.3 & 72 & 7.25 & 100 & 72 & 1.59 & 6.49 & 2.2 & 0.63 & 22 & 72 & 78 & 0.90 & 0.94 \\
Gemini-2.5-FL & 84 & 6.54 & 86 & 50 & 2.40 & 6.67 & 1.6 & 0.66 & 51 & 98 & 46 & 0.87 & 0.89 \\
Gemini-2.5-F & 84 & 6.32 & 81 & 52 & 2.31 & 6.28 & 1.7 & 0.70 & 60 & 92 & 35 & 0.87 & 0.91 \\
Gemini-2.5-Pro & 84 & 6.43 & 81 & 60 & 2.14 & 6.08 & 2.1 & 0.64 & 57 & 89 & 42 & 0.91 & 0.91 \\
\bottomrule\end{tabular}
\end{fitwidth}
\caption{Interviews with the 12 TalkDep personas, seven per persona, PHQ-9 items 1--8 rated by the rater used in the paper. All 8: \% rated as endorsing all eight items, personas labelled moderate or above. Real$_{=}$: real NHANES respondents at the same item 1--8 totals. Diversity and Ratio use pairs of interviews with different personas; Ratio: diversity of real-profile interviews over TalkDep interviews within the same bins of three total points, as in Table~\ref{tab:interview-pipeline} (---: no real-profile interviews). AUC: separability from real respondents with the same item 1--8 totals, the interviews of one persona in one cross-validation fold. Agent: rated PHQ-9 band against the band named like the persona's severity label (minimal: none 0--4; mild 5--9; moderate 10--14; moderate to severe: 15--19; severe 20--27), \% exact, within one band and above. $\rho$: Spearman correlation of the rated total with the persona's BDI-II score, over interviews and over the persona means.}
\label{tab:interview-talkdep}\end{table*}

\begin{table*}[t]\centering\footnotesize\setlength{\tabcolsep}{4pt}
\begin{fitwidth}
\begin{tabular}{@{}lrrrrrrrr@{}}\toprule
& \multicolumn{7}{c}{Against the given profile (50 interviews)} & Severity persona \\
\cmidrule(lr){2-8}\cmidrule(l){9-9}
Rater & Exact & Within 1 & $\kappa_w$ & Absent$\to$absent & Exact if present & 1 rated higher & Shift & All 8 (\%) \\\midrule
R1 & 55.2 & 92.8 & 0.71 & 89.9 & 46.7 & 68 & +0.05 & 100 \\
R2 & 57.0 & 92.0 & 0.70 & 89.9 & 48.9 & 70 & +0.13 & 100 \\
R3 & 58.2 & 89.0 & 0.68 & 89.9 & 50.5 & 75 & +0.22 & 100 \\
R4 & 55.5 & 92.8 & 0.70 & 87.3 & 47.7 & 72 & +0.15 & 100 \\
R5 & 54.0 & 87.5 & 0.65 & 87.3 & 45.8 & 89 & +0.34 & 100 \\
\midrule
All five & 56.0 & 90.8 & 0.69 & 88.9 & 47.9 & 75 [67, 81] & +0.18 & 100 \\
Median of five & 57.8 & 92.0 & 0.70 & 89.9 & 49.8 & 73 & +0.17 & --- \\
Model rater & 53.8 & 89.5 & 0.67 & 87.3 & 45.5 & 91 [85, 97] & +0.39 & 100 \\
\bottomrule\end{tabular}
\end{fitwidth}
\caption{Human ratings of 64 interview transcripts, PHQ-9 items 1--8 (Appendix~\ref{app:prompts}). Columns 2--8: the 40 real-profile and 10 sampled-profile interviews against the profile given to the patient model, as in Table~\ref{tab:interview-agreement}; 1 rated higher: \% of symptoms given as ``several days'' rated higher, with 95\% bootstrap intervals over transcripts. All 8: \% of the 14 severity-persona interviews rated as endorsing all eight items. Model rater: the rater used in the paper, on the same transcripts. Agreement between the human raters, all pairs: exact 62--76\%, $\kappa_w$ 0.79--0.87; each human rater against the model rater: $\kappa_w$ 0.78--0.89, the model rater higher by 0.07--0.38 levels on average.}
\label{tab:human-rating}\end{table*}

\begin{table}[t]\centering\footnotesize\setlength{\tabcolsep}{2.6pt}
\begin{fitwidth}
\begin{tabular}{@{}lrrrrrrr@{}}\toprule
Source & $n$ & Breadth & All 9 & N.a.a. & SI & Div. & AUC \\\midrule
Real, 2013--2023 (held out) & 5{,}779 & 5.43 & 4.1 & 39.6 & 14.0 & 7.30 & --- \\
GRM fit on 2005--2012 & 5{,}779 & 5.49 & 4.9 & 39.1 & 15.6 & 7.23 & 0.52 \\
Real, 2005--2012 & 4{,}981 & 5.38 & 4.2 & 40.2 & 14.9 & 7.01 & 0.53 \\
\bottomrule\end{tabular}
\end{fitwidth}
\caption{Profiles sampled from a GRM fit to NHANES 2005--2012 (Appendix~\ref{app:prompts}), selected to match the total distribution of the held-out 2013--2023 respondents (totals $\geq$5); 2005--2012 respondents are reweighted to it. Div.: diversity. AUC: separability from the held-out respondents.}
\label{tab:grm-heldout}\end{table}

\begin{table*}[t]\centering\footnotesize\setlength{\tabcolsep}{4pt}
\begin{fitwidth}
\begin{tabular}{@{}lrrrrrrrr@{}}\toprule
Trained on & $n$ & SI & AUC & Brier & Slope & Sens. & FPR & Recal.\ Brier \\\midrule
\textbf{Real (NHANES 2005--2012)} & 19{,}685 & 3.2 & 0.82 & 0.096 & 0.83 & 25.6 & 3.1 & 0.096 \\
GRM-sampled profiles & 20{,}000 & 4.3 & 0.79 & 0.100 & 0.81 & 25.6 & 3.7 & 0.099 \\
\midrule
GPT-4o-mini & 450 & 64.7 & 0.80 & 0.132 & 0.18 & 44.5 & 8.2 & 0.098 \\
GPT-4o & 437 & 59.0 & 0.79 & 0.129 & 0.23 & 42.5 & 8.4 & 0.100 \\
GPT-4.1-nano & 436 & 76.8 & 0.72 & 0.240 & 0.21 & 63.4 & 28.1 & 0.106 \\
GPT-4.1-mini & 436 & 50.9 & 0.78 & 0.123 & 0.22 & 30.3 & 5.4 & 0.101 \\
GPT-4.1 & 437 & 62.2 & 0.75 & 0.152 & 0.19 & 48.1 & 12.3 & 0.103 \\
GPT-5-mini & 436 & 48.6 & 0.79 & 0.118 & 0.23 & 31.4 & 5.0 & 0.099 \\
GPT-5.2 & 450 & 66.4 & 0.78 & 0.144 & 0.18 & 48.3 & 10.8 & 0.100 \\
GPT-5.4-mini & 450 & 50.0 & 0.81 & 0.114 & 0.27 & 31.7 & 4.6 & 0.097 \\
Haiku 4.5 & 450 & 54.0 & 0.77 & 0.128 & 0.17 & 34.2 & 5.9 & 0.101 \\
Grok-4.1 & 450 & 35.6 & 0.77 & 0.122 & 0.19 & 22.1 & 3.4 & 0.102 \\
Llama-3.3 & 450 & 35.1 & 0.76 & 0.130 & 0.18 & 36.9 & 7.1 & 0.102 \\
Kimi-K2.5 & 450 & 43.6 & 0.80 & 0.126 & 0.23 & 42.7 & 8.0 & 0.098 \\
Mistral-L3 & 450 & 54.9 & 0.76 & 0.135 & 0.17 & 38.0 & 8.2 & 0.102 \\
Phi-4 & 450 & 58.7 & 0.76 & 0.121 & 0.26 & 26.7 & 4.6 & 0.103 \\
Gemini-2.5-FL & 450 & 65.6 & 0.80 & 0.133 & 0.26 & 49.9 & 10.5 & 0.098 \\
Gemini-2.5-F & 450 & 62.2 & 0.75 & 0.147 & 0.19 & 42.7 & 10.8 & 0.104 \\
Gemini-2.5-Pro & 450 & 55.8 & 0.75 & 0.135 & 0.18 & 37.9 & 8.0 & 0.103 \\
All models pooled & 7{,}582 & 55.5 & 0.79 & 0.127 & 0.25 & 40.4 & 8.1 & 0.100 \\
\bottomrule\end{tabular}
\end{fitwidth}
\caption{Predicting whether PHQ-9 item 9 is endorsed from items 1--8. Test set: NHANES 2013--2023 respondents with items 1--8 total $\geq$5 ($n{=}5{,}747$). $n$, SI: size of the training set and \% endorsing item 9 in it. GRM-sampled: profiles from the GRM fit to NHANES 2005--2012. Slope: calibration slope (1 when calibrated). Sens., FPR: sensitivity and false-positive rate at $p\geq0.5$ (\%). Recal.\ Brier: after logistic recalibration on 1{,}000 held-out 2005--2012 respondents with items 1--8 total $\geq$5.}
\label{tab:downstream}\end{table*}

\begin{table*}[t]\centering\footnotesize\setlength{\tabcolsep}{3.4pt}
\begin{fitwidth}
\begin{tabular}{@{}lrrrrrrrrrr@{}}\toprule
& \multicolumn{2}{c}{SI in training (\%)} & \multicolumn{4}{c}{AUC} & \multicolumn{2}{c}{AUC difference from real} & \multicolumn{2}{c}{Brier} \\
\cmidrule(lr){2-3}\cmidrule(lr){4-7}\cmidrule(lr){8-9}\cmidrule(lr){10-11}
Simulated patients & Sim. & Real$_{t}$ & Sim. & Real$_{r}$ & Real$_{t}$ & Real$_{t,9}$ & Real$_{t}$ & Real$_{t,9}$ & Sim.\ recal. & Real$_{t}$ \\\midrule
GPT-4o-mini & 65 & 28 & 0.803 & 0.764 & 0.808 & 0.775 & $-$0.005 [$-$0.030, +0.023] & +0.028 [+0.008, +0.045] & 0.098 & 0.098 \\
GPT-4o & 59 & 30 & 0.786 & 0.769 & 0.803 & 0.780 & $-$0.017 [$-$0.059, +0.017] & +0.006 [$-$0.032, +0.031] & 0.100 & 0.099 \\
GPT-4.1-nano & 77 & 29 & 0.721 & 0.767 & 0.801 & 0.786 & $-$0.080 [$-$0.141, $-$0.027] & $-$0.064 [$-$0.114, $-$0.025] & 0.106 & 0.099 \\
GPT-4.1-mini & 51 & 33 & 0.780 & 0.772 & 0.805 & 0.777 & $-$0.025 [$-$0.073, +0.013] & +0.002 [$-$0.043, +0.029] & 0.101 & 0.098 \\
GPT-4.1 & 62 & 29 & 0.755 & 0.777 & 0.801 & 0.779 & $-$0.046 [$-$0.086, $-$0.008] & $-$0.024 [$-$0.059, +0.004] & 0.103 & 0.099 \\
GPT-5-mini & 49 & 31 & 0.793 & 0.769 & 0.808 & 0.776 & $-$0.015 [$-$0.057, +0.017] & +0.018 [$-$0.024, +0.042] & 0.099 & 0.098 \\
GPT-5.2 & 66 & 33 & 0.785 & 0.775 & 0.803 & 0.775 & $-$0.018 [$-$0.044, +0.010] & +0.010 [$-$0.007, +0.026] & 0.100 & 0.099 \\
GPT-5.4-mini & 50 & 34 & 0.812 & 0.773 & 0.797 & 0.778 & +0.015 [$-$0.028, +0.049] & +0.034 [$-$0.004, +0.054] & 0.097 & 0.100 \\
Haiku 4.5 & 54 & 32 & 0.775 & 0.778 & 0.804 & 0.774 & $-$0.029 [$-$0.060, +0.003] & +0.001 [$-$0.020, +0.019] & 0.101 & 0.099 \\
Grok-4.1 & 36 & 30 & 0.767 & 0.761 & 0.802 & 0.778 & $-$0.035 [$-$0.085, +0.002] & $-$0.011 [$-$0.054, +0.018] & 0.102 & 0.099 \\
Llama-3.3 & 35 & 28 & 0.764 & 0.772 & 0.801 & 0.775 & $-$0.037 [$-$0.076, +0.002] & $-$0.010 [$-$0.042, +0.017] & 0.102 & 0.098 \\
Kimi-K2.5 & 44 & 27 & 0.801 & 0.768 & 0.803 & 0.778 & $-$0.002 [$-$0.037, +0.025] & +0.024 [$-$0.007, +0.038] & 0.098 & 0.098 \\
Mistral-L3 & 55 & 31 & 0.757 & 0.767 & 0.809 & 0.777 & $-$0.052 [$-$0.090, $-$0.015] & $-$0.020 [$-$0.055, +0.005] & 0.102 & 0.098 \\
Phi-4 & 59 & 41 & 0.764 & 0.764 & 0.798 & 0.790 & $-$0.034 [$-$0.099, +0.021] & $-$0.026 [$-$0.081, +0.012] & 0.103 & 0.100 \\
Gemini-2.5-FL & 66 & 30 & 0.801 & 0.770 & 0.804 & 0.783 & $-$0.003 [$-$0.049, +0.028] & +0.018 [$-$0.020, +0.039] & 0.098 & 0.098 \\
Gemini-2.5-F & 62 & 34 & 0.748 & 0.770 & 0.803 & 0.776 & $-$0.056 [$-$0.111, $-$0.019] & $-$0.028 [$-$0.075, +0.001] & 0.104 & 0.099 \\
Gemini-2.5-Pro & 56 & 32 & 0.755 & 0.772 & 0.809 & 0.774 & $-$0.054 [$-$0.089, $-$0.017] & $-$0.020 [$-$0.051, $-$0.000] & 0.103 & 0.098 \\
\midrule
All models pooled & 55 & 31 & 0.792 & 0.818 & 0.818 & 0.784 & $-$0.027 [$-$0.040, $-$0.015] & +0.008 [$-$0.003, +0.018] & 0.100 & 0.097 \\
\bottomrule\end{tabular}
\end{fitwidth}
\caption{Simulated training sets against real training sets of the same size drawn from NHANES 2005--2012: at random (Real$_{r}$), with the same items 1--8 totals (Real$_{t}$), and with the same totals and the same item-9 answers (Real$_{t,9}$); control values are means over 50 draws. Differences: simulated minus real, with 95\% paired bootstrap intervals (1{,}000 resamples of the simulated training set and the test set, fresh real draws each time). Sim.\ recal.: simulated-trained model after recalibration on held-out real respondents. Test set as in Table~\ref{tab:downstream}.}
\label{tab:downstream-controls}\end{table*}

\begin{table}[t]\centering\footnotesize\setlength{\tabcolsep}{3pt}
\begin{fitwidth}
\begin{tabular}{@{}lrrrrr@{}}\toprule
Items 1--8 & $n$ & Observed & Real & GRM & Sim. \\\midrule
0--4 & 15{,}282 & 0.5 & 1.1 & 1.1 & 0.0 \\
5--9 & 3{,}704 & 6.3 & 4.2 & 4.8 & 0.4 \\
10--14 & 1{,}319 & 17.7 & 14.2 & 17.0 & 13.7 \\
15--19 & 550 & 38.5 & 39.7 & 46.1 & 73.3 \\
20--24 & 174 & 57.5 & 68.6 & 75.2 & 99.1 \\
\bottomrule\end{tabular}
\end{fitwidth}
\caption{Observed and predicted rate (\%) of endorsing item 9 by total on items 1--8, all NHANES 2013--2023 respondents, for models trained on real respondents, GRM-sampled profiles and simulated patients (pooled over models).}
\label{tab:downstream-cal}\end{table}

\begin{table}[t]\centering\footnotesize\setlength{\tabcolsep}{2.4pt}
\begin{fitwidth}
\begin{tabular}{@{}rrrrrrr@{}}\toprule
Real $n$ & Added & Draws & Base AUC & $\Delta$AUC & $\Delta$Brier ($\times10^{-3}$) & $\Delta$Pred. \\\midrule
50 & +1$\times$ & 82 & 0.735 & +0.040 [+0.028, +0.053] & +0.1 [$-$4.0, +4.3] & +2.2 \\
50 & +2$\times$ & 82 & 0.735 & +0.040 [+0.027, +0.053] & +2.9 [$-$1.3, +7.1] & +1.7 \\
50 & +4$\times$ & 82 & 0.735 & +0.046 [+0.033, +0.059] & +2.0 [$-$2.8, +6.8] & +0.5 \\
100 & +1$\times$ & 97 & 0.743 & +0.039 [+0.029, +0.049] & +4.9 [+1.9, +8.0] & +3.3 \\
100 & +2$\times$ & 97 & 0.743 & +0.038 [+0.029, +0.048] & +6.5 [+3.5, +9.4] & +2.2 \\
100 & +4$\times$ & 97 & 0.743 & +0.043 [+0.032, +0.053] & +7.8 [+5.0, +10.7] & +1.8 \\
200 & +1$\times$ & 99 & 0.747 & +0.040 [+0.034, +0.047] & +4.0 [+1.8, +6.2] & +3.2 \\
200 & +2$\times$ & 99 & 0.747 & +0.040 [+0.033, +0.048] & +6.1 [+3.7, +8.6] & +2.5 \\
200 & +4$\times$ & 99 & 0.747 & +0.042 [+0.035, +0.050] & +8.0 [+5.5, +10.4] & +1.9 \\
500 & +1$\times$ & 100 & 0.777 & +0.020 [+0.016, +0.025] & +8.5 [+7.1, +10.0] & +2.1 \\
500 & +2$\times$ & 100 & 0.777 & +0.019 [+0.014, +0.024] & +12.3 [+10.8, +13.9] & +1.6 \\
500 & +4$\times$ & 100 & 0.777 & +0.017 [+0.012, +0.022] & +15.4 [+13.8, +16.9] & +1.2 \\
\bottomrule\end{tabular}
\end{fitwidth}
\caption{Adding simulated patients (pooled over models) to a small random real training set from NHANES 2005--2012. Mean change over draws with its 95\% interval; draws without both classes are skipped. $\Delta$Pred.: change in mean predicted risk minus the observed rate (points). Test set as in Table~\ref{tab:downstream}.}
\label{tab:augment}\end{table}

\begin{table*}[t]\centering\footnotesize\setlength{\tabcolsep}{4pt}
\begin{fitwidth}
\begin{tabular}{@{}lrrrrrrrrrrrr@{}}\toprule
& \multicolumn{4}{c}{Told a severity} & \multicolumn{6}{c}{Given a real profile} & \multicolumn{2}{c}{Sampled profile} \\
\cmidrule(lr){2-5}\cmidrule(lr){6-11}\cmidrule(l){12-13}
Patient model & Exact & Within 1 & Above & Absent & Exact & Within 1 & Above & Absent & Abs.\ det. & Agree & Exact & Within 1 \\\midrule
GPT-4o-mini & 25 & 56 & 75 & 0.0 & 55 & 96 & 40 & 14.8 & 79 & 96 & 56 & 98 \\
GPT-4.1 & 26 & 72 & 74 & 0.3 & --- & --- & --- & --- & --- & --- & --- & --- \\
Haiku 4.5 & 25 & 55 & 75 & 0.2 & --- & --- & --- & --- & --- & --- & --- & --- \\
Llama-3.3 & 20 & 44 & 80 & 0.0 & 45 & 91 & 51 & 15.4 & 82 & 97 & 49 & 88 \\
Gemini-2.5-FL & 25 & 56 & 75 & 0.0 & 42 & 92 & 51 & 15.6 & 83 & 97 & 42 & 91 \\
Gemini-2.5-F & 26 & 51 & 74 & 0.0 & 61 & 98 & 34 & 16.6 & 88 & 98 & 56 & 96 \\
Gemini-2.5-Pro & 25 & 59 & 75 & 0.0 & 59 & 94 & 35 & 17.0 & 91 & 98 & 64 & 94 \\
\midrule
64 transcripts, model rater & 33 & 50 & 67 & 0 & 48 & 98 & 48 & 16 & 85 & 97 & 54 & 92 \\
64 transcripts, human raters & 35 (22--56) & 69 (62--88) & 60 (44--67) & 0 (0--0) & 56 (52--68) & 97 (92--100) & 26 (10--40) & 17 (16--17) & 87 (85--89) & 98 (97--98) & 62 (54--71) & 98 (92--100) \\
\bottomrule\end{tabular}
\end{fitwidth}
\caption{The interviewer and model rater as an assessment agent. Exact, Within 1: \% of interviews whose rated PHQ-9 total falls in the patient's severity band (none 0--4, mild 5--9, moderate 10--14, moderately severe 15--19, severe 20--27), or within one band; the patient's severity is the instructed one for patients told a severity and the band of the given profile otherwise. Above: \% rated in a higher band. Absent: \% of symptoms 1--8 rated absent. Abs.\ det.: \% of symptoms absent from the given profile that are rated absent. Agree: \% of symptoms 1--8 whose presence or absence matches the given profile. Bottom rows: the 64 transcripts rated by people (14, 40 and 10 per condition), scored the same way, with Exact, Within 1 and Above averaged with equal weight over the patients' severities, as in the full design; human raters, mean (range) over the five raters. Table~\ref{tab:interview-talkdep} gives the TalkDep personas.}
\label{tab:agent}\end{table*}

\begin{table}[t]\centering\footnotesize\setlength{\tabcolsep}{4pt}
\begin{fitwidth}
\begin{tabular}{@{}lrr@{}}\toprule
Group & Similarity & Contrast \\\midrule
NHANES, random halves & 0.98 & +0.10 \\
Real, at each model's totals & 0.53--0.77 & +0.09 to +0.12 \\
Models & 0.13--0.61 & $-$0.01 to +0.18 \\
Models' item distributions, no covariation & $-$0.02 to 0.35 & $-$0.02 to +0.07 \\
Models $+$ noise & 0.01--0.12 & +0.00 to +0.01 \\
GRM-sampled profiles & 0.40 & +0.00 \\
\bottomrule\end{tabular}
\end{fitwidth}
\caption{Item covariance at a fixed PHQ-9 total (totals 5--24). Similarity: correlation of a group's 36 residual correlations with those of all NHANES respondents. Contrast: mean residual correlation within the somatic and within the non-somatic items minus that between them. Ranges over the 17 models. Real respondents drawn at each model's totals, and responses with the model's item distributions at each total and no further covariation, match its sample size (100 draws); GRM-sampled profiles: 20{,}000. Table~\ref{tab:structure} gives each model and the interviews.}
\label{tab:structure-summary}\end{table}

\begin{table*}[t]\centering\footnotesize\setlength{\tabcolsep}{4pt}
\begin{fitwidth}
\begin{tabular}{@{}lrrrrrrrrr@{}}\toprule
& & \multicolumn{2}{c}{Group} & \multicolumn{2}{c}{Real$_{=}$} & \multicolumn{2}{c}{No covariation} & \multicolumn{2}{c}{Model $+$ noise} \\
\cmidrule(lr){3-4}\cmidrule(lr){5-6}\cmidrule(lr){7-8}\cmidrule(l){9-10}
Group & $n$ & Contrast & Similarity & Contrast & Similarity & Contrast & Similarity & Contrast & Similarity \\\midrule
NHANES, random halves & 5{,}352 & +0.10 & 0.98 & & & & & & \\
GRM-sampled profiles & 20{,}000 & +0.00 & 0.40 & & & & & & \\
\midrule
GPT-4o-mini & 436 & +0.05 [+0.01, +0.09] & 0.34 & +0.11 & 0.74 [0.63] & +0.01 [+0.05] & 0.21 [0.42] & +0.01 & 0.09 \\
GPT-4o & 360 & +0.11 [+0.06, +0.16] & 0.61 & +0.11 & 0.71 [0.59] & +0.01 [+0.04] & 0.18 [0.41] & +0.01 & 0.12 \\
GPT-4.1-nano & 350 & +0.06 [+0.01, +0.10] & 0.13 & +0.12 & 0.70 [0.59] & +0.00 [+0.04] & $-$0.02 [0.20] & +0.01 & 0.10 \\
GPT-4.1-mini & 348 & +0.02 [$-$0.01, +0.06] & 0.14 & +0.10 & 0.68 [0.55] & +0.03 [+0.07] & 0.20 [0.38] & +0.00 & 0.04 \\
GPT-4.1 & 361 & +0.03 [$-$0.03, +0.07] & 0.44 & +0.11 & 0.74 [0.62] & $-$0.02 [+0.02] & 0.29 [0.39] & +0.01 & 0.10 \\
GPT-5-mini & 374 & +0.09 [+0.04, +0.14] & 0.53 & +0.10 & 0.74 [0.62] & +0.00 [+0.05] & 0.13 [0.41] & +0.01 & 0.10 \\
GPT-5.2 & 394 & +0.18 [+0.14, +0.24] & 0.46 & +0.11 & 0.69 [0.59] & +0.07 [+0.13] & 0.24 [0.41] & +0.01 & 0.02 \\
GPT-5.4-mini & 365 & +0.10 [+0.04, +0.15] & 0.48 & +0.10 & 0.70 [0.57] & +0.03 [+0.08] & 0.24 [0.42] & +0.00 & 0.05 \\
Haiku 4.5 & 396 & +0.06 [+0.02, +0.12] & 0.32 & +0.11 & 0.70 [0.59] & +0.04 [+0.08] & 0.21 [0.39] & +0.01 & 0.04 \\
Grok-4.1 & 381 & $-$0.01 [$-$0.04, +0.03] & 0.20 & +0.11 & 0.74 [0.64] & +0.01 [+0.05] & 0.20 [0.40] & +0.00 & 0.03 \\
Llama-3.3 & 363 & +0.03 [$-$0.01, +0.08] & 0.25 & +0.12 & 0.77 [0.67] & +0.03 [+0.09] & 0.24 [0.39] & +0.00 & 0.12 \\
Kimi-K2.5 & 376 & +0.10 [+0.05, +0.15] & 0.35 & +0.12 & 0.76 [0.65] & +0.03 [+0.07] & 0.10 [0.31] & +0.00 & 0.06 \\
Mistral-L3 & 366 & +0.09 [+0.05, +0.13] & 0.54 & +0.11 & 0.72 [0.58] & +0.03 [+0.07] & 0.35 [0.49] & +0.01 & 0.09 \\
Phi-4 & 351 & +0.04 [$-$0.02, +0.08] & 0.22 & +0.09 & 0.53 [0.40] & +0.01 [+0.05] & $-$0.02 [0.18] & +0.00 & 0.08 \\
Gemini-2.5-FL & 381 & +0.10 [+0.06, +0.16] & 0.40 & +0.11 & 0.70 [0.59] & +0.00 [+0.05] & 0.05 [0.25] & +0.01 & 0.04 \\
Gemini-2.5-F & 349 & +0.05 [+0.01, +0.09] & 0.38 & +0.10 & 0.66 [0.52] & +0.02 [+0.06] & 0.12 [0.33] & +0.00 & 0.01 \\
Gemini-2.5-Pro & 373 & +0.07 [+0.03, +0.11] & 0.53 & +0.11 & 0.74 [0.65] & +0.01 [+0.05] & 0.31 [0.47] & +0.01 & 0.07 \\
\midrule
Real profiles given & 79 & +0.04 & 0.34 & +0.11 & 0.51 [0.26] & & & & \\
Interviews, real profile & 381 & +0.14 [+0.06, +0.20] & 0.58 & +0.11 & 0.51 [0.26] & & & & \\
Sampled profiles given & 76 & $-$0.01 & 0.01 & +0.12 & 0.53 [0.24] & & & & \\
Interviews, sampled profile & 370 & +0.04 [$-$0.07, +0.14] & 0.26 & +0.12 & 0.53 [0.24] & & & & \\
\bottomrule\end{tabular}
\end{fitwidth}
\caption{Item covariance at a fixed PHQ-9 total (totals 5--24). Residuals are item scores minus the group's mean item scores at the same total. Contrast: mean residual correlation within the somatic (sleep, fatigue, appetite, concentration, psychomotor) and the non-somatic items (interest, mood, worthlessness, suicidal thoughts) minus that between them (all NHANES respondents: +0.10), with 95\% bootstrap intervals for models. Similarity: correlation of the group's 36 residual correlations with those of all NHANES respondents. Real$_{=}$: real respondents drawn at the model's totals, one per response, mean over 100 draws (2.5th percentile in brackets). No covariation: responses drawn, one per model response, from the maximum-entropy distribution with the model's item distributions at each total, mean over 100 draws (97.5th percentile in brackets). Model $+$ noise: each item moved one level up or down with a probability chosen so that diversity reaches that of real respondents with the same totals, mean over 20 draws. Bottom rows: the profiles given in interviews and the PHQ-9 rated from those interviews (primary rater), pooled over the five patient models, which were given the same profiles; intervals resample profiles, and Real$_{=}$ draws as many real respondents as there are profiles, at their totals (200 draws).}
\label{tab:structure}\end{table*}

\begin{table}[t]\centering\footnotesize\setlength{\tabcolsep}{2.4pt}
\begin{fitwidth}
\begin{fitwidth}
\begin{tabular}{@{}lrrrrrrrr@{}}\toprule
& \multicolumn{2}{c}{Mild} & \multicolumn{2}{c}{Moderate} & \multicolumn{2}{c}{Mod.\ severe} & \multicolumn{2}{c}{Above, all} \\
\cmidrule(lr){2-3}\cmidrule(lr){4-5}\cmidrule(lr){6-7}\cmidrule(l){8-9}
Model & Med. & Above & Med. & Above & Med. & Above & Naive & Coached \\\midrule
GPT-4o-mini & 12 & 100 & 16 & 96 & 19 & 42 & 79 & 58 \\
GPT-4o & 12 & 92 & 16 & 78 & 21 & 72 & 81 & 39 \\
GPT-4.1-nano & 13 & 90 & 18 & 100 & 21 & 78 & 89 & 51 \\
GPT-4.1-mini & 12 & 96 & 19 & 92 & 21 & 100 & 96 & 56 \\
GPT-4.1 & 9 & 28 & 16 & 96 & 21 & 100 & 75 & 25 \\
GPT-5-mini & 11 & 88 & 15 & 54 & 21 & 90 & 77 & 53 \\
GPT-5.2 & 11 & 72 & 17 & 100 & 21 & 96 & 89 & 59 \\
GPT-5.4-mini & 12 & 96 & 17 & 88 & 22 & 98 & 94 & 53 \\
Haiku 4.5 & 13 & 100 & 19 & 100 & 23 & 96 & 99 & 41 \\
Grok-4.1 & 12 & 88 & 16 & 86 & 20 & 88 & 87 & 35 \\
Llama-3.3 & 13 & 100 & 15 & 52 & 21 & 86 & 79 & 35 \\
Kimi-K2.5 & 12 & 98 & 15 & 54 & 18 & 30 & 61 & 31 \\
Mistral-L3 & 12 & 100 & 18 & 100 & 22 & 96 & 99 & 39 \\
Phi-4 & 17 & 100 & 21 & 100 & 23 & 100 & 100 & 98 \\
Gemini-2.5-FL & 13 & 100 & 18 & 100 & 20 & 58 & 86 & 50 \\
Gemini-2.5-F & 12 & 98 & 19 & 96 & 22 & 92 & 95 & 79 \\
Gemini-2.5-Pro & 11 & 84 & 16 & 90 & 21 & 84 & 86 & 57 \\
\bottomrule\end{tabular}
\end{fitwidth}
\end{fitwidth}
\caption{PHQ-9 totals by instructed severity. Med.: median total; Above: \% of responses above the instructed band (mild 5--9, moderate 10--14, moderately severe 15--19), naive arm. Above, all: mean over the three bands, naive and coached arms.}
\label{tab:overshoot}\end{table}

\clearpage
\onecolumn
\section{Additional Figures}
\label{app:figs}

\begin{figure*}[!htbp]
\centering
\includegraphics[width=\textwidth]{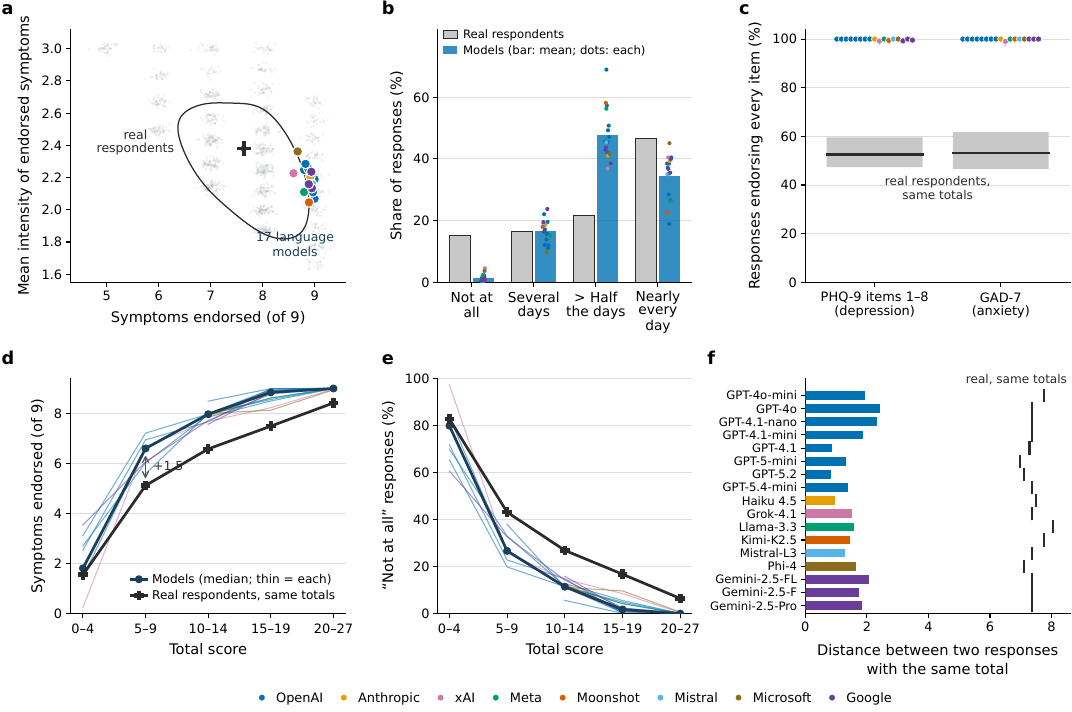}
\caption{\textbf{Response profiles on the PHQ-9 and GAD-7, full version.} Panels a, b and f use PHQ-9 responses with total 15--24. \textbf{a}, Symptoms endorsed against their mean intensity (grey dots, real respondents; contour, their densest half; coloured, model means). \textbf{b}, Use of the response options. \textbf{c}, Responses endorsing every item, PHQ-9 items 1--8 (total 15--24) and GAD-7 (12--19), with 95\% intervals; grey, real respondents with the same totals as each model (range and median over models). \textbf{d}, \textbf{e}, Symptoms endorsed and ``not at all'' answers by PHQ-9 total, against real respondents with the same totals. \textbf{f}, Diversity; ticks, real respondents at each model's totals.}
\label{fig:phenomenon-full}
\end{figure*}

\begin{figure*}[!htbp]
\centering
\includegraphics[width=\textwidth]{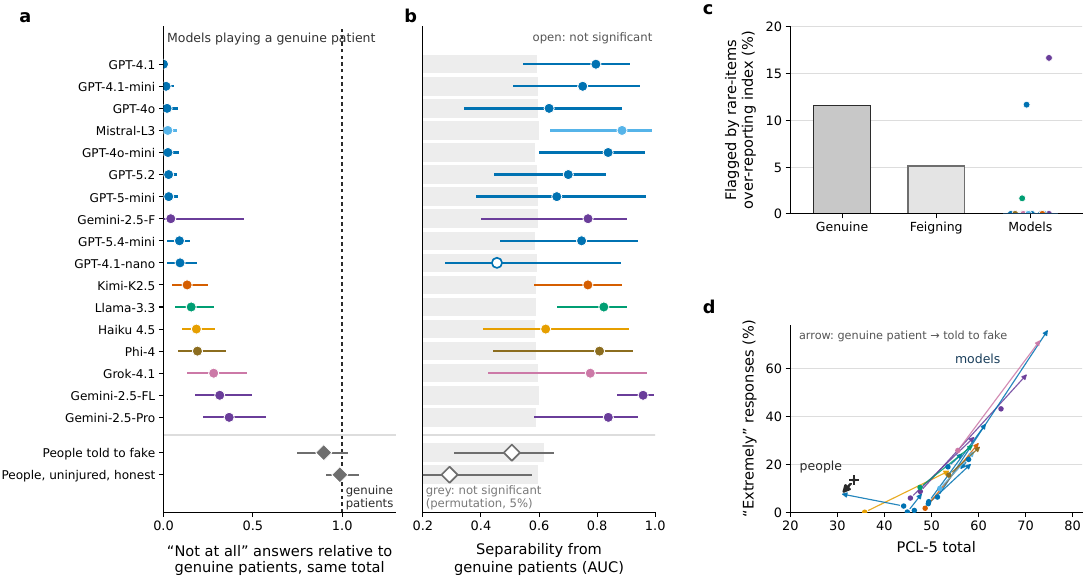}
\caption{\textbf{Comparison with people instructed to feign} (PCL-5), full version. \textbf{a}, ``Not at all'' answers relative to genuinely injured students with the same totals (95\% bootstrap intervals). \textbf{b}, Separability from them, with the 2.5th--97.5th percentiles over 80\% subsamples; grey, below each group's one-sided 5\% critical value of a label-permutation test; open, not significant (Holm-adjusted for models). \textbf{c}, Rate flagged by the rare-items over-reporting index. \textbf{d}, Change in total and in ``extremely'' answers from the patient to the feigning role.}
\label{fig:feigners-full}
\end{figure*}

\begin{figure*}[!htbp]
\centering
\includegraphics[width=\textwidth]{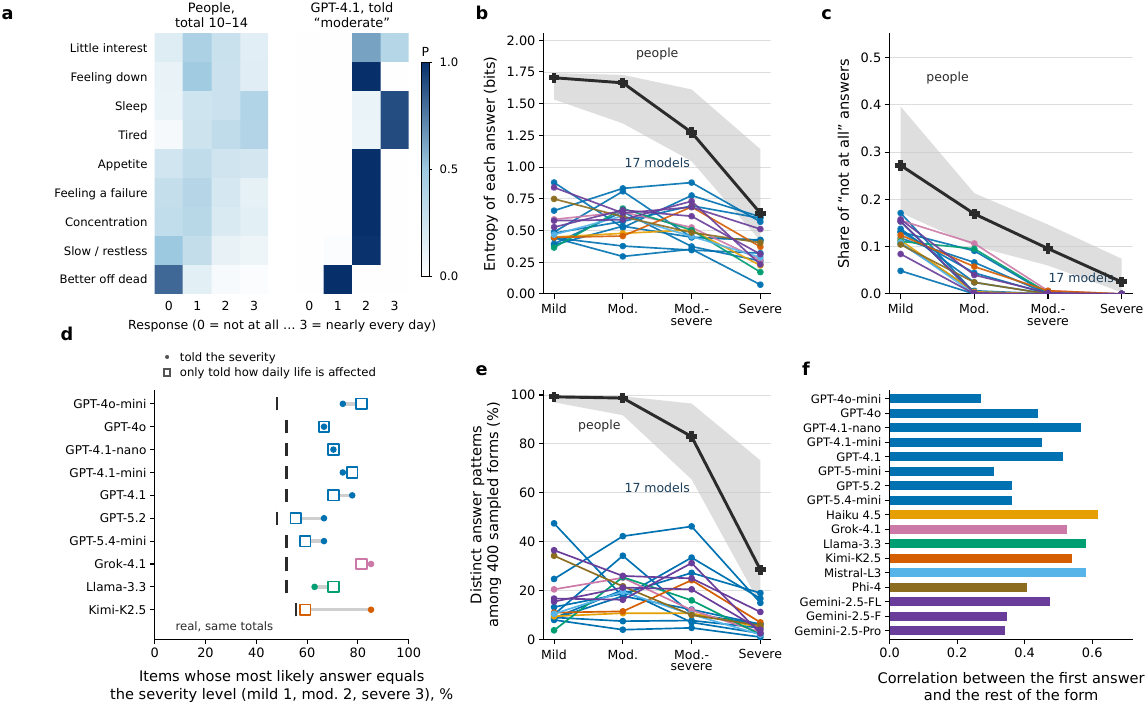}
\caption{\textbf{Per-item answer distributions, full version.} \textbf{a}, Answer probabilities per PHQ-9 item: real respondents with total 10--14 and GPT-4.1 told ``moderate'' (log-probabilities). \textbf{b}, Entropy of each item's answers across the whole-form responses of a severity cell. \textbf{c}, Share of ``not at all'' answers. \textbf{d}, Items whose most likely answer, from log-probabilities, equals the severity level, told the severity (dots) or only how daily life is affected (squares); ticks, real respondents with the same totals as the model's whole forms; moderately severe is left out. \textbf{e}, Distinct profiles among 400 forms sampled item by item from the answers in b. \textbf{f}, Correlation between the first answer and the rest of the form. People in b, c, e: real respondents with the same totals as each model's forms (grey, range over models; line, median).}
\label{fig:mechanism-full}
\end{figure*}

\begin{figure*}[!htbp]
\centering
\includegraphics[width=\textwidth]{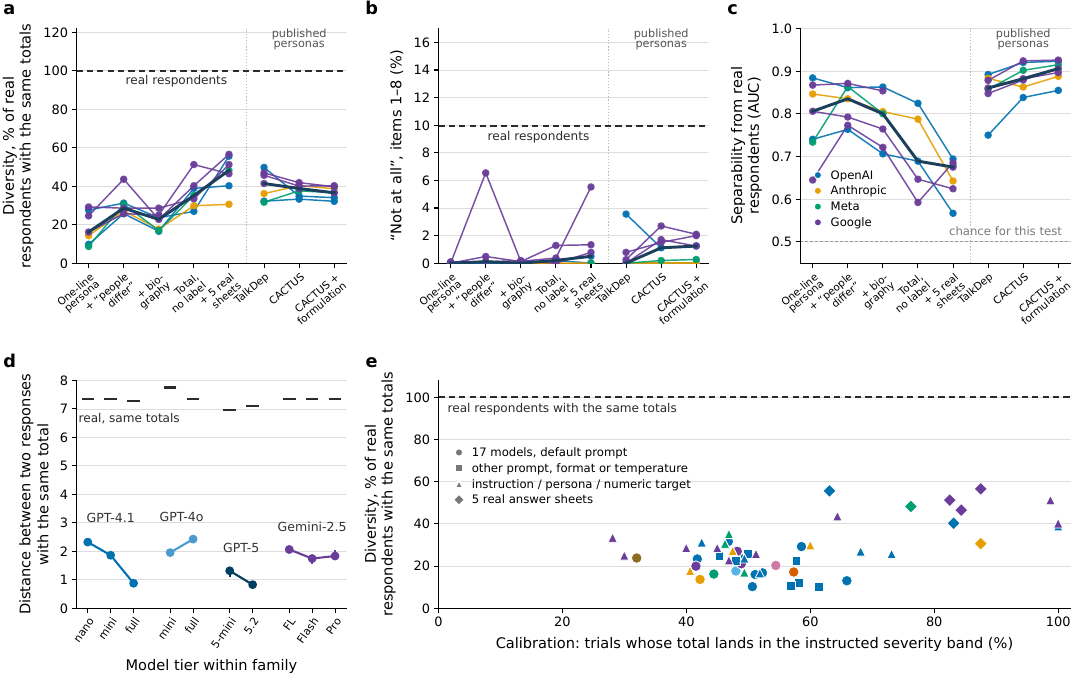}
\caption{\textbf{Interventions, published personas and model scale.} \textbf{a}--\textbf{c}, Diversity as a percentage of held-out real respondents with the same totals (totals 5--27), ``not at all'' answers and separability (totals 15--24) with the one-line persona, four changes to it, and the TalkDep and CACTUS personas, CACTUS also with each client's case formulation (thin, models; thick, median; dashed, held-out real respondents or chance). \textbf{d}, Diversity (totals 15--24) by model tier within four families, with the 2.5th--97.5th percentiles over 80\% subsamples; dashes, real respondents at each model's totals. \textbf{e}, Calibration against diversity (as in a), all configurations.}
\label{fig:interventions}
\end{figure*}

\begin{figure*}[!htbp]
\centering
\includegraphics[width=\textwidth]{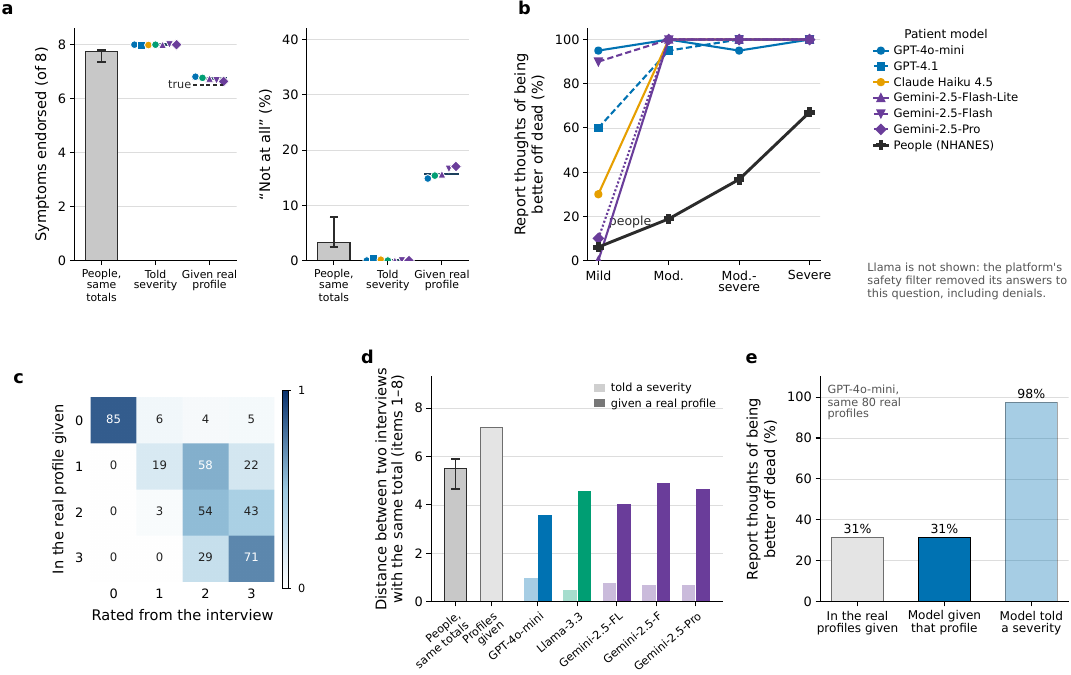}
\caption{\textbf{Simulated clinical interviews, full version.} \textbf{a}, Symptoms endorsed and ``not at all'' answers, PHQ-9 items 1--8 rated from transcripts (people, real respondents with the same totals; dashed, the profiles given). \textbf{b}, Reported thoughts of being better off dead by severity. \textbf{c}, Given against rated answer, row percentages. \textbf{d}, Diversity. \textbf{e}, Suicidal ideation in the real profiles given, in the interviews built on them, and with a severity persona (GPT-4o-mini).}
\label{fig:conversation-full}
\end{figure*}

Figures~\ref{fig:phenomenon-full}, \ref{fig:mechanism-full} and~\ref{fig:conversation-full} are the full versions of Figures~\ref{fig:phenomenon}, \ref{fig:mechanism} and~\ref{fig:conversation}; Figures~\ref{fig:feigners-full} and~\ref{fig:interventions} show the full feigning comparison and the interventions.

\begin{figure*}[!htbp]
\centering
\includegraphics[width=\textwidth]{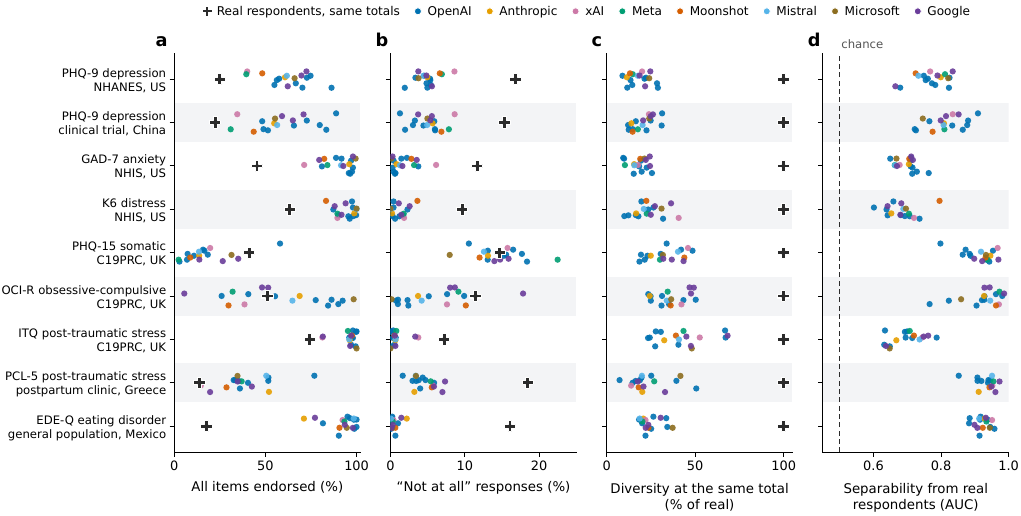}
\caption{\textbf{Profile measures on the nine human references.} \textbf{a}, Responses endorsing every item. \textbf{b}, ``Not at all'' answers. \textbf{c}, Diversity as a percentage of the real value. \textbf{d}, Separability; dashed line, chance. Crosses, real respondents with the same totals.}
\label{fig:ed-gen}
\end{figure*}

\end{document}